\documentclass{article}

\usepackage{iclr2026_conference,times}
\iclrfinalcopy

\usepackage{amsmath,amsfonts,bm}

\def\eqref#1{equation~\ref{#1}}

\def\partref#1{part~\ref{#1}}

\def\1{\bm{1}}

\DeclareMathAlphabet{\mathsfit}{\encodingdefault}{\sfdefault}{m}{sl}
\SetMathAlphabet{\mathsfit}{bold}{\encodingdefault}{\sfdefault}{bx}{n}

\usepackage[utf8]{inputenc} 
\usepackage{lipsum} 

\usepackage[T1]{fontenc}

\usepackage{etoolbox}
\newbool{includeappendix}
\setbool{includeappendix}{true}

\ifdefined\isoverfull
\else
\fi

\usepackage{subcaption}

\usepackage[x11names]{xcolor}

\definecolor{my-full-blue}{HTML}{1F77B4}

\definecolor{my-full-orange}{HTML}{FF7F0E}
\definecolor{my-extra-orange}{HTML}{7C4514}

\definecolor{my-full-green}{HTML}{2CA02C}

\definecolor{my-full-red}{HTML}{d62728}

\definecolor{my-full-purple}{HTML}{9467bd}

\colorlet{my-blue}{my-full-blue!30}
\colorlet{my-orange}{my-full-orange!30}
\colorlet{my-green}{my-full-green!90}
\colorlet{my-red}{my-full-red!90}
\colorlet{my-purple}{my-full-purple!30}

\definecolor{mygreen}{HTML}{B5E3B5}
\colorlet{myred}{LightPink1}
\colorlet{myblue}{SteelBlue2}
\definecolor{mylightblue}{HTML}{B0DBFF}
\colorlet{myorange}{DarkOrange1}

\definecolor{darkblue}{RGB}{34,49,63}
\definecolor{lightblue}{RGB}{238,244,249}
\definecolor{accentblue}{RGB}{88,139,202}
\definecolor{lightgreen}{RGB}{220,245,230}
\definecolor{lightred}{RGB}{250,225,225}
\definecolor{darkgreen}{RGB}{40,167,69}
\definecolor{darkred}{RGB}{220,53,69}
\definecolor{textgray}{RGB}{120,120,120}
\definecolor{lightgray}{RGB}{240,240,240}   
\definecolor{accentorange}{RGB}{127,17,144} 
\definecolor{gray2}{HTML}{FCFCFC}

\usepackage{listings}

\usepackage{textcomp}

\usepackage{xcolor}

\usepackage[scaled=0.8]{beramono}

\definecolor{ckeyword}{HTML}{7F0055}
\definecolor{ccomment}{HTML}{3F7F5F}
\definecolor{cstring}{HTML}{2A0099}

\lstdefinestyle{numbers}{
	numbers=left,
	framexleftmargin=20pt,
	numberstyle=\tiny,
	firstnumber=auto,
	numbersep=1em,
	xleftmargin=2em
}

\lstdefinestyle{layout}{
	frame=none,
	captionpos=b,
}

\lstdefinestyle{comment-style}{
	morecomment=[l]//,
	morecomment=[s]{/*}{*/},
	commentstyle={\color{ccomment}\itshape},
}

\lstdefinestyle{string-style}{
	showstringspaces=false,%
}

\lstdefinestyle{keyword-style}{
	keywordstyle={\ttfamily\bfseries},
	morekeywords={
		function,
		constructor,
		int,
		bool,
		return,
		returns,
		uint
	},
	morekeywords = [2]{},
	keywordstyle = [2]{\text},
	sensitive=true,
}

\lstdefinestyle{input-encoding}{
	inputencoding=utf8,
	extendedchars=true,
	literate=
	{ℝ}{$\reals$}1%
	{→}{$\rightarrow$}1%
	{α}{$\alpha$}1%
	{β}{$\beta$}1%
	{λ}{$\lambda$}1%
	{θ}{$\theta$}1%
	{ϕ}{$\phi$}1%
}

\lstdefinestyle{escaping}{
	moredelim={**[is][\color{blue}]{\%}{\%}},
	moredelim={**[is][\color{pastelgreen}]{??}{??}},
	mathescape=true
}

\lstdefinestyle{default-style}{
	basicstyle=\fontencoding{T1}\ttfamily\footnotesize,
	style=numbers,
	style=layout,
	style=comment-style,
	style=string-style,
	style=keyword-style,
	style=input-encoding,
	style=escaping,
	tabsize=2,
	upquote=true
}

\lstdefinelanguage{BASIC}{
	language=C++,
	style=default-style
}[keywords,comments,strings]%

\lstdefinelanguage{JavaScript}{
morekeywords=[1]{break, continue, delete, else, for, function, if, in,
new, return, this, typeof, var, void, while, with, const, let},
morekeywords=[2]{false, null, true, boolean, number, undefined, string,
Array, Boolean, Date, Math, Number, String, Object},
morekeywords=[3]{eval, parseFloat, escape, unescape},
sensitive,
morecomment=[s]{/*}{*/},
morecomment=[l]//,
morecomment=[s]{/**}{*/}, 
morestring=[b]',
morestring=[b]"
}[keywords, comments, strings]
\definecolor{delim}{RGB}{20,105,176}
\definecolor{numb}{RGB}{106, 109, 32}
\definecolor{string}{rgb}{0.64,0.08,0.08}

\lstdefinelanguage{json}{
    showspaces=false,
    showtabs=false,
    breaklines=true,
    postbreak=\raisebox{0ex}[0ex][0ex]{\ensuremath{\color{gray}\hookrightarrow\space}},
    upquote=true,
    morestring=[b]",
    morecomment=[l]//,
    stringstyle=\color{string},
    literate=
     *{0}{{{\color{numb}0}}}{1}
      {1}{{{\color{numb}1}}}{1}
      {2}{{{\color{numb}2}}}{1}
      {3}{{{\color{numb}3}}}{1}
      {4}{{{\color{numb}4}}}{1}
      {5}{{{\color{numb}5}}}{1}
      {6}{{{\color{numb}6}}}{1}
      {7}{{{\color{numb}7}}}{1}
      {8}{{{\color{numb}8}}}{1}
      {9}{{{\color{numb}9}}}{1}
      {\{}{{{\color{delim}{\{}}}}{1}
      {\}}{{{\color{delim}{\}}}}}{1}
      {[}{{{\color{delim}{[}}}}{1}
      {]}{{{\color{delim}{]}}}}{1},
}
\definecolor{dkgreen}{rgb}{0,0.6,0}
\definecolor{dred}{rgb}{0.545,0,0}
\definecolor{dblue}{rgb}{0,0,0.545}
\definecolor{lgrey}{rgb}{0.9,0.9,0.9}
\definecolor{gray}{rgb}{0.4,0.4,0.4}
\definecolor{darkblue}{rgb}{0.0,0.0,0.6}
\lstdefinelanguage{cpp}{
      breaklines=true,               
      postbreak=\raisebox{0ex}[0ex][0ex]{\ensuremath{\color{gray}\hookrightarrow\space}},
      deletekeywords={...},          
      escapeinside={\%*}{*)},                  
      language=C++,                
      keywordstyle=\color{purple},  
      morekeywords={string,float}, 
      identifierstyle=\color{black},
      stringstyle=\color{blue},      
      showspaces=false,               
      showstringspaces=false,        
      showtabs=false,                
      tabsize=5,                     
    }

\usepackage[english]{babel}
\usepackage[breaklinks=true]{hyperref}
\definecolor{darkpastelblue}{HTML}{0279AF}
\definecolor{linkcolor}{HTML}{0279AF}

\hypersetup{colorlinks,
      linkcolor=darkpastelblue,
      citecolor=darkpastelblue,
      urlcolor=darkpastelblue,
      filecolor=darkpastelblue}
\usepackage{algorithm}
\usepackage{algorithmicx}
\usepackage[noend]{algpseudocode}
\usepackage{color,soul}
\usepackage{ bbold }
\usepackage{multirow}
\usepackage{multicol}
\usepackage{caption}
\usepackage{tabularx,booktabs,xltabular}
\usepackage{graphicx}
\usepackage{tabu}
\usepackage{array}
\usepackage{siunitx}
\usepackage{subcaption}
\usepackage{fontawesome}
\usepackage{stackengine}
\usepackage{svg}
\usepackage{mathtools}
\usepackage{xspace}
\usepackage{wrapfig}
\usepackage{makecell}
\usepackage{placeins}
\usepackage{float}
\usepackage{pifont}
\usepackage{svg}
\usepackage[most]{tcolorbox}
\usepackage{array}
\usepackage{dsfont}
\usepackage{enumitem}
\usepackage[utf8]{inputenc}
\usepackage{textcomp}
\usepackage{amsthm}
\usepackage{amssymb,amsmath}
\usepackage{syntax}
\usepackage{semantic}

\newcolumntype{x}[2]{S[table-format=#1.#2,table-auto-round]}

\usepackage{tikz}

\usetikzlibrary{arrows}
\usetikzlibrary{automata}
\usetikzlibrary{calc}
\usetikzlibrary{backgrounds}
\usetikzlibrary{decorations.markings}
\usetikzlibrary{decorations.pathmorphing}
\usetikzlibrary{decorations.pathreplacing}
\usetikzlibrary{fit}
\usetikzlibrary{patterns}
\usetikzlibrary{positioning}
\usetikzlibrary{matrix}
\usetikzlibrary{shadows}
\usetikzlibrary{shapes}
\usetikzlibrary{shapes.geometric}
\usetikzlibrary{arrows.meta}
\usetikzlibrary{shadows.blur}

\definecolor{blue}{HTML}{347bc6}
\definecolor{green-underline}{HTML}{2de12c}
\definecolor{yellow-underline}{HTML}{ffd700}

\tikzstyle{block} = [
    minimum width=1cm,
    minimum height=0.5cm,
    align=center,
    fill=gray!30,
    rounded corners=5pt,
]

\tikzstyle{arrow} = [
	-{Latex[length=1.4mm, width=1.4mm]},
	draw=blue,
]

\tikzstyle{dashedline} = [
    draw=blue,
    dashed
]

\tikzstyle{line} = [
    draw=blue
]

\definecolor{lightblue}{RGB}{173,216,230} 

\usepackage{pgfplots}
\usepackage[capitalize,noabbrev]{cleveref}
\AtBeginDocument{
\crefname{appendix}{Appendix}{Appendices}
}

\newtheorem{definition}{Definition}
\newtheorem{theorem}{Theorem}
\newtheorem{lemma}[theorem]{Lemma}
\newtheorem{assumption}{Assumption}
\newtheorem{corrolary}{Corrolary}

\usepackage{pifont}
\definecolor{pastelgreen}{HTML}{059C05}
\definecolor{pastelred}{HTML}{FF7373}
\newcommand{\rev}[1]{#1}
\DeclareRobustCommand{\lexfix}[1]{#1}

\usepackage{setspace}

\newcommand{\ttt}[1]{\text{\texttt{#1}}}

\newcommand{\strcat}{\ensuremath{\circ}{}}

\newcommand*\circled[1]{\tikz[baseline=(char.base)]{
            \node[shape=circle,draw,inner sep=1pt] (char) {#1};}}

\usepackage{bm}

\usepackage{scalerel}

\newcommand{\wrongtok}[1]{\colorbox{myred}{\hz{#1}}}
\newcommand{\righttok}[1]{\colorbox{mygreen}{\hz{#1}}}

\newcommand{\hz}{\vphantom{\parbox[c]{1cm}{\rule{1cm}{1.2mm}}}}
\newcommand{\appcref}[1]{\cref{#1}}
\newcommand{\Hsquare}{%
  \text{\kern2\scriptspace\fboxsep=-.2pt\fbox{\rule{0pt}{1ex}\rule{1ex}{0pt}}\kern2\scriptspace}%
}

\newcommand{\ctwof}{C2F$^{+\varepsilon}$}
\def\emptyset{\ensuremath{\varnothing}}
\def\epsilon{\ensuremath{\varepsilon}}
\def\mask{\ensuremath{\bot}}
\def\hole{\ensuremath{\Hsquare}}

\newcommand{\indnt}{\hspace*{\algorithmicindent}}

\def\presuper#1#2%
  {\mathop{}%
   \mathopen{\vphantom{#2}}^{#1}%
   \kern-\scriptspace%
   #2}
\def\surround#1#2#3%
  {\mathop{}%
   \mathopen{\vphantom{#2}}^{#1}%
   \kern-3\scriptspace%
   #2%
   \kern4\scriptspace%
   \mathopen{\vphantom{#2}}^{#3}%
   }

\newcommand{\triple}[3]{\surround{#1}{\vec{#2}}{#3}}

\usepackage{stmaryrd}
\newcommand{\dfato}[2]{#1 \shortrightarrow #2}

\newcommand{\ltr}{\textsc{pre}}
\newcommand{\fim}{\textsc{fim}}
\newcommand{\mri}{\textsc{mri}}
\newcommand{\mril}{\textsc{mri-l}}
\newcommand{\dlm}{\textsc{dlm}}

\newcommand{\smiles}{\textsc{smiles}}
\newcommand{\json}{\textsc{json}}
\newcommand{\jsonD}{\textsc{json-nous}}
\newcommand{\gsmk}{\textsc{gsm8k-symbolic}}
\newcommand{\humaneval}{\textsc{C++}}
\newcommand{\mrione}{\textsc{1-mri}}
\newcommand{\mritwo}{\textsc{2-mri}}
\newcommand{\mrithree}{\textsc{3-mri}}
\newcommand{\mrionel}{\textsc{1-mri-l}}
\newcommand{\mritwol}{\textsc{2-mri-l}}
\newcommand{\mrithreel}{\textsc{3-mri-l}}

\newcommand{\starcodertwo}{\textsc{StarCoder2 7B}}
\newcommand{\codegemma}{\textsc{CodeGemma 7B}}
\newcommand{\deepseekcoder}{\textsc{DeepSeek Coder}}
\newcommand{\deepseekcodertwob}{\deepseekcoder{} \textsc{1.3B}}
\newcommand{\deepseekctwob}{\textsc{DeepSeek C. 1.3B}}

\newcommand{\deepseekcsevenb}{\textsc{DeepSeek C. 6.7B}}
\newcommand{\deepseekcoderthirtyb}{\textsc{DeepSeek Coder 33B}}
\newcommand{\deepseekcthirtyb}{\textsc{DeepSeek C. 33B}}

\newcommand{\dream}{\textsc{Dream 7B}}
\newcommand{\dreamcoder}{\textsc{DreamCoder 7B}}
\newcommand{\dreamc}{\textsc{DreamC. 7B}}
\newcommand{\llada}{\textsc{LLaDA 8B}}
\newcommand{\diffucoder}{\textsc{DiffuCoder 7B}}
\newcommand{\diffuc}{\textsc{DiffuC. 7B}}

\newcommand{\gemini}{\textsc{Gemini-2.5-Pro}}

\newcommand{\vanilla}{\ensuremath{\text{Van.}}}
\newcommand{\constrained}{\ensuremath{\text{Con.}^{-}}}
\newcommand{\constrplus}{\ensuremath{\text{Con.}}}
\newcommand{\grammarprompting}{\ensuremath{\text{G.P.}}}
\newcommand{\dingo}{\textsc{dingo}}

\newcommand{\lexeme}[1]{\texttt{<#1>}}

\newcommand{\maxresample}{100}

\algrenewcommand\alglinenumber[1]{\scriptsize #1\hspace{1mm}}
\newcounter{algline}[algorithm]

\makeatletter
\newcommand{\algline}[1]{%
  \setcounter{algline}{\value{ALG@line}}%
  \refstepcounter{algline}%
  \label{#1}}
\renewcommand{\ALG@step}{%
  \refstepcounter{ALG@line}
  \addtocounter{ALG@rem}{1}%
  \ifthenelse{\equal{\arabic{ALG@rem}}{\ALG@numberfreq}}%
    {\setcounter{ALG@rem}{0}\alglinenumber{\arabic{ALG@line}}}%
    {}%
}
\makeatother

\newcommand{\Hquad}{\hspace{0.5em}}

\usepackage[capitalize]{cleveref}

\crefformat{section}{\S#2#1#3}

\crefrangeformat{section}{\S#3#1#4\crefrangeconjunction\S#5#2#6}

\crefmultiformat{section}{\S#2#1#3}{\crefpairconjunction\S#2#1#3}{\crefmiddleconjunction\S#2#1#3}{\creflastconjunction\S#2#1#3}

\newcommand{\crefrangeconjunction}{--}

\crefname{listing}{Lst.}{listings}
\crefname{line}{Line}{Lines}
\crefname{algline}{Line}{Lines}
\crefalias{ALG@line}{line}
\crefname{appendix}{App.}{App.}
\crefname{lemma}{Lemma}{Lemmata}

\newcommand{\app}[1]{%
	\ifbool{includeappendix}{\cref{#1}}{the appendix}%
}
\newcommand{\App}[1]{%
	\ifbool{includeappendix}{\cref{#1}}{The appendix}%
}

\makeatletter
\AddToHook{cmd/appendix/before}{\def\cref@section@alias{appendix}}
\AddToHook{cmd/appendix/before}{\def\cref@subsection@alias{appendix}}
\AddToHook{cmd/appendix/before}{\def\cref@subsubsection@alias{appendix}}
\makeatother

\title{Constrained Decoding of Diffusion LLMs\\ with Context-Free Grammars}

\author{Niels Mündler, Jasper Dekoninck, Martin Vechev\\
Department of Computer Science\\
ETH Zurich, Switzerland\\
\texttt{\{niels.muendler,jasper.dekoninck,martin.vechev\}@inf.ethz.ch}\\
}

\begin{document}

\maketitle

\begin{center}
  \vspace{-10mm}
\raisebox{-0.2\height}{\includegraphics[height=1em]{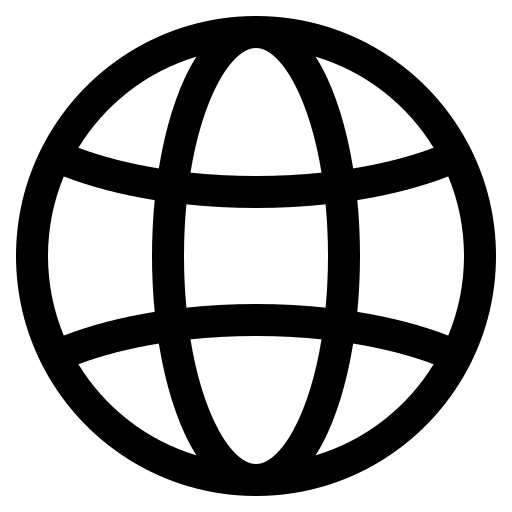}} \url{https://constrained-diffusion.ai}\vspace{0.5mm}\\
\raisebox{-0.2\height}{\includegraphics[height=1em]{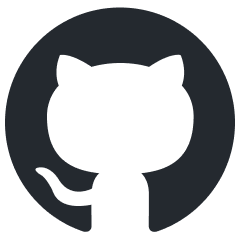}} \url{https://github.com/eth-sri/constrained-diffusion}\\
\vspace{3mm}
\end{center}

\begin{abstract}
Large language models (LLMs) have shown promising performance across diverse domains. Many practical applications of LLMs, such as code completion and structured data extraction, require adherence to syntactic constraints specified by a formal language. Yet, due to their probabilistic nature, LLM output is not guaranteed to adhere to such formal languages. To address this, prior work has proposed constrained decoding to restrict LLM generation to particular formal languages. However, existing works are not applicable to the emerging paradigm of diffusion LLMs, as this requires supporting token generation in arbitrary order instead of the traditional left-to-right order. In this paper, we address this challenge and present the first constrained decoding method for diffusion models, one that can handle formal languages captured by context-free grammars.
Our method relies on solving a newly defined additive infilling problem, which asks whether a partial output with holes can be completed to a valid word in the target language. We reduce this problem to deciding whether the intersection of the target language and a particular regular language is empty, and present an efficient decision algorithm for context-free languages. Empirical results on various applications, such as C++ code infilling and structured data extraction in JSON, demonstrate that our method achieves near-perfect syntactic correctness while consistently preserving or improving functional correctness. Importantly, our efficiency optimizations ensure that the computational overhead remains practical.
\end{abstract}

\vspace{-3mm}
\section{Introduction}
\label{sec:introduction}

\vspace{-2mm}
Large language models (LLMs) have recently achieved promising performance across a wide range of tasks \citep{openai2023gpt4,deepmind2025geminipro}. Due to their capabilities in code synthesis, they achieve impressive scores on diverse code benchmarks \citep{humaneval,vero2025baxbench,swebench,livecodebench} and are integrated into developer workflows as programming copilots \citep{github-copilot-2025,tabnine2025}. Further, they are used for processing information into machine-readable formats in various domains \citep{schmidt2025exploringuselargelanguage,schillingwilhelmi2024textinsightlargelanguage,goel2023llmsaccelerateannotationmedical}.
Despite these successes, LLMs are inherently probabilistic and offer no guarantees about syntactic validity of generated output, providing an inherent limitation for LLM users.

\vspace{-2mm}
\paragraph{Constrained decoding}
A promising approach that mitigates this limitation is constrained decoding \citep{synchromesh,dominosDblp2403,ugare2024syncode,melcer2024constraineddecodingfillinthemiddlecode}. This technique leverages the formal grammar of a target language to guide the generation process, ensuring that the output remains within the language's bounds.
Constrained decoding leverages parsing and validation of the generated output in lockstep with the incremental generation process, allowing the model to avoid invalid continuations without restarting inference.
It has been widely adopted in practice, with commercial providers offering the option to restrict output to JSON or context-free grammars~\citep{openai-function-calling,anthropic-json-mode}. 

\vspace{-2mm}
\paragraph{Current limitations of constrained decoding}
Constrained decoding is usually applied to context-free grammars (CFGs), which capture the syntax of common programming languages and popular data formats, like C++ and JSON \citep{knuth1965translation,cogumbreiro2020cs420}.
In this context, they can only be applied to left-to-right prefix completion, a common LLM generation setting.
However, this setting does not capture more advanced use cases with LLMs, such as diffusion LLMs. While \citet{melcer2024constraineddecodingfillinthemiddlecode} extend constrained decoding to completions between a fixed prefix and suffix, and \citet{suresh2025dingoconstrainedinferencediffusion} constrain diffusion LLMs to regular languages, no prior work supports diffusion LLM constraining with CFGs.

\begin{figure}[t]
    \centering
    \resizebox{\textwidth}{!}{
    \begin{tikzpicture}[
    node distance=6mm and 8mm,
    base/.style={
        draw=gray,
        fill=lightgray,
        rounded corners=3pt,
        font=\sffamily,
        blur shadow={shadow xshift=0.5pt, shadow yshift=-0.5pt, shadow opacity=20}
    },
    darkbox/.style={
        draw=gray,
        fill=lightgray!20,
        rounded corners=3pt,
        minimum width=6.3cm,
        minimum height=3.5cm,
        anchor=west,
        xshift=0.3cm
    },
    codebox/.style={
        base,
        minimum height=0.5cm,
        minimum width=3cm,
        align=center
    },
    proposalbox/.style={
        base,
        minimum height=0.4cm,
        minimum width=5.7cm,
        align=center
    },
    optionbox/.style={
        base,
        minimum width=2.8cm,
        minimum height=0.4cm,
        font=\ttfamily\tiny
    },
    syntaxbox/.style={
        base,
        fill=accentblue!20,
        draw=accentblue,
        minimum width=3cm,
        font=\ttfamily\tiny,
        align=left,
        text width=2.6cm
    },
    rulebox/.style={
        base,
        minimum width=2.8cm,
        minimum height=0.4cm,
        font=\ttfamily\tiny,
        text width=2.1cm
    },
    cfgbox/.style={
        base,
        fill=accentblue!20,
        draw=accentblue,
        minimum width=0.6cm,
        font=\tiny,
        align=center,
        anchor=center,
        text width=0.5cm,
    },
    placeholder/.style={
        fill=gray2,
        rounded corners=1pt,
        inner xsep=1pt,
        inner ysep=1.5pt,
        font=\ttfamily\tiny,
        text height=1ex,
        text depth=0.25ex,
        anchor=base west
    },
    placeholder2/.style={
        fill=gray2,
        rounded corners=1pt,
        inner xsep=1pt,
        inner ysep=1.5pt,
        font=\ttfamily\tiny,
        color=white,
        text height=1ex,
        text depth=0.25ex,
        anchor=base west
    },
    resultbox/.style={
        base,
        minimum size=0.8cm,
        font=\Large\bfseries,
    },
    accept/.style={resultbox, fill=lightgreen, draw=darkgreen, text=darkgreen},
    reject/.style={resultbox, fill=lightred, draw=darkred, text=darkred},
    arrow/.style={
        ->,
        thick,
        line width=0.4mm,
        color=blue!80, 
        >={Triangle[scale=0.6]}
    }
]


\node[codebox] (code) at (0,0) {};
\node[anchor=west, font=\ttfamily\tiny, xshift=0.1cm] (text1) at (code.west) {int};
\node[placeholder2, right=0pt of text1, xshift=-0cm] (ph1) {\_\_\_\_};
\node[right=0pt of ph1, font=\ttfamily\tiny, xshift=-0.15cm] (text2) {() \{ return 2};
\node[placeholder2, right=2pt of text2, xshift=-0.1cm] (ph2) {\_\_};
\node[anchor=south west, font=\tiny, text=gray, yshift=-0.1cm] (codecomment) at (code.north west) {\emph{Input $x$}};

\node[syntaxbox, text width=2.4cm, below=of code] (cfg) {
    S \(\rightarrow\) Def ( ) \{ Lines \} \\
    Lines \(\rightarrow\) Line ; Lines | \(\varepsilon\) \\
    Def \(\rightarrow\) \lexeme{type} \lexeme{name}
};
\node[yshift=-0.1cm, font=\tiny, anchor=south west, text=gray] (cfgtext) at (cfg.north west) {\emph{Constraint Language (CFG)}};

\node[codebox, below=of cfg] (output) {};
\node[anchor=west, font=\ttfamily\tiny, xshift=0.1cm] (outputtext1) at (output.west) {int};
\node[right=0pt of outputtext1, font=\ttfamily\tiny, xshift=-0.1cm] (outputph1) {foo};
\node[right=0pt of outputph1, font=\ttfamily\tiny, xshift=-0.15cm] (outputtext2) {() \{ return 2};
\node[placeholder2,right=0pt of outputtext2, font=\ttfamily\tiny, xshift=-0cm] (outputph2) {\_\_};
\node[anchor=south west, font=\tiny, text=gray, yshift=-0.1cm] (outputcomment) at (output.north west) {\emph{Output $x'$}};

\node[darkbox] (darkbox) at ($(code.east)!0.5!(output.east)$) {};

\node[proposalbox, anchor=north, font=\tiny, yshift=-0.1cm] (proposalbox) at (darkbox.north) {\emph{Update $x' \sim M ( x )$ }};

\node[optionbox, below=of proposalbox, xshift=-1.6cm, yshift=0.3cm] (option1) {};
\node[anchor=west, font=\ttfamily\tiny, xshift=0cm] (optiontext1) at (option1.west) {int};
\node[placeholder, right=0pt of optiontext1, xshift=-0.05cm] (optionph1) {foo()};
\node[right=0pt of optionph1, font=\ttfamily\tiny, xshift=-0.10cm, yshift=-0.01cm] (optiontext2) {() \{ return 2};
\node[placeholder2, right=-0pt of optiontext2, xshift=-.05cm, yshift=0.01cm] (optionph2) {\_\_};
\node[anchor=south west, font=\tiny, text=gray, xshift=-0.1cm, yshift=-0.1cm] (optioncomment1) at (option1.north west) {\emph{Proposal}};

\node[optionbox, below=of proposalbox, xshift=1.6cm, yshift=0.3cm] (option2) {};
\node[anchor=west, font=\ttfamily\tiny, xshift=0cm] (optiontext2) at (option2.west) {int};
\node[placeholder, right=0pt of optiontext2, xshift=-0.05cm] (optionph2) {foo};
\node[right=0pt of optionph2, font=\ttfamily\tiny, xshift=-0.10cm, yshift=-0.01cm] (optiontext3) {() \{ return 2};
\node[placeholder2, right=-0pt of optiontext3, xshift=-0cm, yshift=0.01cm] (optionph2) {\_\_};
\node[anchor=south west, font=\tiny, text=gray, xshift=-0.1cm, yshift=-0.1cm] (optioncomment2) at (option2.north west) {\emph{Proposal}};

\node[rulebox, below=of option1, yshift=0.3cm] (regex1) {};
\node[anchor=west, font=\ttfamily\tiny, xshift=0cm] (regex1-p1) at (regex1.west) {int};
\node[placeholder, right=0pt of regex1-p1, xshift=-0.05cm] (optionre1) {foo()};
\node[right=1pt of optionre1, font=\ttfamily\tiny, xshift=-0.15cm, yshift=-0.01cm] (regex1-p2) {() \{ return 2};
\node[placeholder, right=0pt of regex1-p2, xshift=-0.05cm, yshift=0.01cm] (optionre2) {.*};
\node[anchor=south west, font=\tiny, text=gray, xshift=-0.1cm, yshift=-0.1cm] (regexcomment1) at (regex1.north west) {\emph{Regex}};

\node[rulebox, below=of option2, yshift=0.3cm] (regex2) {};
\node[anchor=west, font=\ttfamily\tiny, xshift=0cm] (regex2-p1) at (regex2.west) {int};
\node[placeholder, right=0pt of regex2-p1, xshift=-0.05cm] (optionre2) {foo};
\node[right=1pt of optionre2, font=\ttfamily\tiny, xshift=-0.15cm, yshift=-0.01cm] (regex2-p2) {() \{ return 2};
\node[placeholder, right=0pt of regex2-p2, xshift=-0cm, yshift=0.01cm] (optionre3) {.*};
\node[anchor=south west, font=\tiny, text=gray, xshift=-0.1cm, yshift=-0.1cm] (regexcomment2) at (regex2.north west) {\emph{Regex}};

\node[reject, below=of regex1, yshift=0cm] (decision1) {\ding{55}};
\node[accept, below=of regex2, yshift=0cm] (decision2) {\ding{51}};
\node[anchor=west, font=\tiny, text=gray, xshift=0.1cm, yshift=0.15cm] (rejectcomment1) at (decision1.east) {\emph{Rejected}};

\node[cfgbox, xshift=-0.8cm] (cfgbox) at ($(decision1.north)!0.5!(regex1.south)$) {CFG};
\node[cfgbox, xshift=0.8cm] (cfgbox2) at ($(decision2.north)!0.5!(regex2.south)$) {CFG};

\node[font=\small, xshift=0.2cm] (intersection1) at (cfgbox.east) {$\cap$};
\node[font=\small, xshift=-0.2cm] (intersection2) at (cfgbox2.west) {$\cap$};


\draw[arrow] (code.east) -- ++(0.3,0);
\draw[arrow] (cfg.east) -- ++(0.3,0);
\draw[arrow, color=darkgreen!60] (output.east) ++(0.3,0) -- (output.east);

\draw[arrow, dotted] (output.west) -- ++(-0.3,0) |- (code.west);

\draw[arrow] (proposalbox.south -| option1.north) -- (option1.north);
\draw[arrow, dotted] (proposalbox.south -| option2.north) -- (option2.north);
\draw[arrow] (option1.south) -- (regex1.north);
\draw[arrow] (option2.south) -- (regex2.north);
\draw[arrow] (regex1.south) -- (decision1.north);
\draw[arrow] (regex2.south) -- (decision2.north);

\draw[arrow, darkred] (decision1.east) -- ++(1.1,0) |- (option2.west);

\end{tikzpicture}
    }
    \caption{An overview of our approach. In each step, the input consists of a partial text $x$ with arbitrarily many infilling regions and a context-free grammar (\textcolor{blue}{CFG}) specifying formal constraints. During decoding, we sample an updated input $x'$ from $M$, obtained, e.g., by inserting a token in one of the regions in $x$. Our method then intersects the CFG with the regular language of all possible completions of $x'$. If the intersection is empty, the update is \textcolor{darkred}{rejected} and a new $x'$ is sampled. Otherwise, it is \textcolor{darkgreen}{accepted} and the decoding continues from $x'$. In the example, the invalid update inserting \texttt{"foo()"} is rejected and \texttt{"foo"} is accepted instead.}
    \label{fig:overview}
\vspace{-3mm}
\end{figure}

\vspace{-2mm}
\paragraph{This work: Constrained decoding for \mri{} and \dlm{}s}
In this work, we present a generalized method for constrained decoding of diffusion LLMs (\dlm{}), which also naturally subsumes the previously unaddressed setting of multi-region infilling (\mri{}). We first generalize the formal framework of constrained decoding to support unordered updates of a partial output with arbitrarily many infilling regions, capturing both \mri{} and \dlm{}. The decoding process is illustrated in \cref{fig:overview}. A model iteratively updates, e.g., inserting a token in a specific location. We verify that the updated output is valid by intersecting the target language's CFG with the language of all possible partial output completions. This intersection is non-empty if and only if a valid completion exists. 

A key challenge in this approach is efficiently determining the intersection's emptiness. To this end, we first show that the set of possible completions is described by a regular language, allowing us to describe the intersection using standard formal language operations.
We then drastically reduce the worst case cubic cost of checking the emptiness of the resulting intersection language using specialized methods for grammar size reduction and search optimizations, including a custom normal form and an implicit search that avoids generating the entire language.

\vspace{-2mm}
\paragraph{Experimentally confirmed consistent improvements}
Our experiments demonstrate a substantial improvement in the reliability of formal language adherence across all evaluated settings. Specifically, the algorithm guarantees valid completions in all settings, up to sampling timeouts. Additionally, it improves functional correctness by up to $7\%$. Importantly, our approach incurs no initial latency and only modest runtime overhead on tested models, with inference time less than doubling on average, enabling practical usage even for complex constraining grammars.

\vspace{-2mm}
\paragraph{Key contributions} Our three key contributions are: (i) a generalized formal constrained decoding framework for the \mri{} and \dlm{} settings, (ii) a novel constrained decoding algorithm for these settings, and (iii) an extensive evaluation of our method using state-of-the-art open-weight infilling and diffusion LLMs, demonstrating consistent improvements in syntactic and functional correctness on C++ code generation, JSON schema extraction, and chemical molecule description.

\section{Background}\label{sec:background}
We outline the necessary background relevant to this work, including generation paradigms with LLMs, constrained decoding, and the relevant properties of regular and context-free languages.

\begin{figure}[t]
    \centering
    \hfill{}
    \begin{subfigure}{.30\linewidth}
      \centering
      \resizebox{\linewidth}{!}{
%

\definecolor{blueplaceholder}{HTML}{DEE8F4}
\begin{tikzpicture}[
    font=\small\ttfamily,
    title/.style={
        font=\bfseries\sffamily\normalsize,
        anchor=east,
        xshift=-0.6cm
    },
    timestep/.style={
        font=\tiny\sffamily\bfseries,
        anchor=south,
        yshift=1.5mm
    },
    placeholder/.style={
        fill=blueplaceholder,
        rounded corners=1pt,
        inner xsep=1pt,
        inner ysep=1.5pt,
        font=\ttfamily\small,
        color=blueplaceholder,
        text height=1.5ex,
        text depth=0.25ex,
        anchor=base west
    }
]

\node (ltr) at (0,0) {};
\node (fim) at (0, -0.6) {};
\node (mri) at (0, -1.2) {};
\node (dllm) at (0, -2.1) {};
\draw[dashed] (-1.8, -1.65) -- (3, -1.65);

\node[title] at (ltr.west) {\ltr{}};
\node[title] at (fim.west) {\fim{}};
\node[title] at (mri.west) {\mri{}};
\node[title] at (dllm.west) {\dlm{}};

\node[xshift=-0.8cm, anchor=west] (ltrtext) at (ltr.east) {int main() \{ };
\node[placeholder, anchor=west] (ltext) at (ltrtext.east) {\_\_\_\_\_\_\_\_\_\_};

\node[xshift=-0.8cm, anchor=west] (fimtext) at (fim.east) {int};
\node[placeholder, anchor=west] (ftext1) at (fimtext.east) {\_\_\_\_};
\node[anchor=west, xshift=-0.15cm] (ftext2) at (ftext1.east) {() \{ return 2;\}};

\node[xshift=-0.8cm, anchor=west] (mritext) at (mri.east) {<type>};
\node[placeholder, anchor=west] (mtext1) at (mritext.east) {\_\_\_\_};
\node[anchor=west, xshift=-0.15cm] (mtext2) at (mtext1.east) {() \{};
\node[placeholder, anchor=west] (mtext3) at (mtext2.east) {\_\_\_\_\_\_};
\node[anchor=west] (mtext4) at (mtext3.east) { <int>;\}};

\node[xshift=-0.8cm, anchor=west] (dllmtext) at (dllm.east) {int};
\node[placeholder, anchor=west] (dtext1) at (dllmtext.east) {\_\_\_\_};
\node[anchor=west, xshift=-0.15cm] (dtext2) at (dtext1.east) {() \{};
\node[placeholder, anchor=west] (dtext3) at (dtext2.east) {\_\_\_};
\node[placeholder, xshift=0.08cm, anchor=west] (dtext4) at (dtext3.east) {\_\_\_};
\node[anchor=west] (dtext5) at (dtext4.east) {;\}};

\node[anchor=north, yshift=-0.04cm] (dtext7) at (dtext1.south) {\circled{2}};

\node[anchor=north, yshift=-0.04cm] (dtext9) at (dtext3.south) {\circled{1}};
\node[anchor=north, yshift=-0.04cm] (dtext7) at (dtext4.south) {\circled{3}};

\end{tikzpicture}
      }
      \subcaption{Generation paradigms}
      \label{fig:code-generation-paradigms}
    \end{subfigure}
    \hfill{}
    \begin{subfigure}{.68\linewidth}
      \centering
      \resizebox{\linewidth}{!}{
          \begin{tikzpicture}[->, >=stealth', auto, semithick, node distance=2cm]
    \node[state, initial left, initial text=] (q0) {$q_0$};
    \begin{scope}[local bounding box=D1, shift=(q0.west), xshift=25mm]
        \node[state, right of=q0] (d1-0) {};
        \node[state, right of=d1-0] (d1-1) {};
    \end{scope}
    \node[state, right of=d1-1] (q1) {$q_1$};
    \begin{scope}[local bounding box=D2, shift=(q0.west), xshift=25mm]
        \node[state, right of=q1] (d2-0) {};
        \node[state, right of=d2-0] (d2-1) {};
        \node[state, right of=d2-1] (d2-2) {};
        \node[state, right of=d2-2] (d2-3) {};
    \end{scope}
    \node[state, below of=d2-3, node distance=1.7cm] (q2) {$q_2$};
    \begin{scope}[local bounding box=D3, shift=(q0.west), xshift=25mm]
        \node[state, left of=q2] (d3-0) {};
        \node[state, left of=d3-0] (d3-1) {};
        \node[state, left of=d3-1] (d3-2) {};
        \node[state, left of=d3-2] (d3-3) {};
    \end{scope}
    \node[state, left of=d3-3, accepting] (q3) {$q_3$};
  
    \path (q0) edge [bend left]  node {$\varepsilon$} (d1-0);

    \path (d1-0) edge [bend left] node {\lexeme{type}} (d1-1);
    \path (d1-1) edge [bend left] node {$\varepsilon$} (q1);
  
    \path (q1) edge [loop above] node {$\Sigma^{*}$} (q1);
    \path (q1) edge [bend left] node {$\varepsilon$} (d2-0);

    \path (d2-0) edge [bend left] node {(} (d2-1);
    \path (d2-1) edge [bend left] node {)} (d2-2);
    \path (d2-2) edge [bend left] node {\{} (d2-3);
    \path (d2-3) edge [bend left] node {$\varepsilon$} (q2);

    \path (q2) edge [loop right] node {$\Sigma^{*}$} (q2);
    \path (q2) edge [bend left] node {$\varepsilon$} (d3-0);

    \path (d3-0) edge [bend left] node {\lexeme{int}} (d3-1);
    \path (d3-1) edge [bend left] node {;} (d3-2);
    \path (d3-2) edge [bend left] node {\}} (d3-3);
    \path (d3-3) edge [bend left] node {$\varepsilon$} (q3);

    \node[draw,dashed,fit=(D1), yshift=2mm, inner xsep=2mm, inner ysep=3.5mm] (boxD1) {};
    \node[anchor=south west] at (boxD1.north west) {$D_1$};
    \node[draw,dashed,fit=(D2), yshift=2mm, inner xsep=2mm, inner ysep=3.5mm] (boxD2) {};
    \node[anchor=south west] at (boxD2.north west) {$D_2$};
    \node[draw,dashed,fit=(D3), yshift=-2mm, inner xsep=2mm, inner ysep=3.5mm] (boxD3) {};
    \node[anchor=south west] at (boxD3.north west) {$D_3$};
               
\end{tikzpicture}
      }
      \subcaption{NFA accepting all possible completions of the \mri{} example.}
    \label{fig:constructed-dfa}
      
    \end{subfigure}
    \hfill{}
    \vspace{-3mm}
    \caption{We consider three left-to-right (\ltr{}, \fim{}, \mri{}) and one out-of-order (\dlm{}) generation paradigms (a). The NFA in (b) describes the language of all additive completions for the \mri{} task.}
    \vspace{-3mm}
  \end{figure}

\subsection{LLM Generation Paradigms}\label{sec:code-generation}

We focus on four generation settings with LLMs illustrated in \cref{fig:code-generation-paradigms}. The first three approaches are commonly used with autoregressive models and generate outputs left-to-right.
\vspace{-2mm}

\paragraph{\ltr{}, \fim{} and \mri{}} The first approach, Prefix generation (\ltr{}) completes a fixed prefix, and is commonly used for synthesizing text or code from scratch. Second, Fill-In-the-Middle (\fim{}) completes text between a given prefix and suffix, and is widely used in code completion assistants \citep{github-copilot-2025,pycharm-code-completion}. Third, Multi-Region Infilling (\mri{}) generalizes \fim{} by allowing prefix and suffix constraints as well as fixed segments in between, with the model infilling the gaps. This enables more flexible editing, useful for repository-level code modifications \citep{coeditor}.

\paragraph{Generation with \dlm{}s} 
Diffusion Language Models (\dlm{}s) \citep{dream2025,nie2025llada} iteratively insert tokens into an initially empty or partially filled sequence $(x_1, x_2, \ldots, x_n)$ where each $x_i$ is either a token from the vocabulary $V$ or a mask $\mask$. At each step, the model predicts one or more indices $k$ of a masked token, i.e., $x_k = \mask$, and a token $t \in V$ to produce the updated sequence $(x_1, \ldots, x_{k-1}, t, x_{k+1}, \ldots, x_n)$. This process continues until no masks remain. The number of predicted indices and tokens per forward pass is a hyperparameter that controls a trade-off between increased generation speed and quality \citep{nie2025llada}. In the example in \cref{fig:code-generation-paradigms}, the model would generate one index and token at a time, first producing the return keyword \circled{1}, then the function name \circled{2}, and finally the return value \circled{3}. 

\vspace{-2mm}
\paragraph{Constrained generation} 
Constrained generation restricts the model to produce outputs that conform to predefined syntactic or structural rules, ensuring syntactically valid code or adherence to structural patterns \citep{synchromesh}. Formally, the model must generate an output $w \in L$, where $L$ is a formal language defining admissible outputs for the given task.
\rev{Constrained generation is implemented by restricting the model's probability distribution, either using precomputed masks \citep{ugare2024syncode,synchromesh}, sampling and rejecting invalid tokens \citep{melcer2024constraineddecodingfillinthemiddlecode,mundler2025typeaware} or a combination of these \citep{dong2024xgrammar,dominosDblp2403}.}

\vspace{-2mm}
\subsection{Regular and Context-Free Languages}
\label{sec:background-languages}

We briefly outline the properties and notation of regular and context-free languages that are relevant to our method. We provide a more detailed introduction in \cref{app:background-languages}.

\vspace{-2mm}
\paragraph{Regular Languages} A regular language is a set of strings that can be described by a deterministic finite automaton (DFA). A DFA is defined as a tuple $(Q, \Sigma, \delta, q_0, F)$, where: (1) $Q$ is a finite set of states, (2) $\Sigma$ is a finite alphabet of symbols, (3) $\delta: Q \times \Sigma \rightarrow Q$ is a transition function that maps a state and an input symbol to the next state, (4) $q_0 \in Q$ is the initial state, and (5) $F \subseteq Q$ is the set of accepting states. The language of a DFA consists of those strings that transition the automaton from the initial to an accepting state through the transition function. Non-deterministic finite automata (NFA) additionally allow multiple next states for the same state and symbol and traversing $\varepsilon$-transitions without consuming a symbol.
An example is depicted in \cref{fig:constructed-dfa}. Every NFA is equivalent to some DFA.

\vspace{-2mm}
\paragraph{Context-Free Languages} Context-free languages (CFLs) are a superset of regular languages, including languages that enforce recursive structures, such as balanced parentheses or nested control statements. They can be described by context-free grammars (CFGs). A CFG is a tuple $(V, \Sigma, P, S)$, where: (1) $V$ is a finite set of nonterminals, (2) $\Sigma$ is a finite set of terminals (with $V \cap \Sigma = \emptyset$), (3) $P$ is a set of productions $A \rightarrow \alpha$, with $A \in V$ and $\alpha \in (V \cup \Sigma)^{\ast}$, and (4) $S \in V$ is the start symbol. The language is defined as all strings generated by the following procedure: Starting with $S$, apply a rule $A \rightarrow \alpha$ from $P$ to replace nonterminal $A$ with $\alpha$, until the result contains only terminals.

\section{Constrained Decoding for Infilling and Diffusion}
\label{sec:generalized-constraiend-decoding}

In this section, we first define the decision problem that enables \mri{} and \dlm{} generation settings, and then introduce our algorithm for efficiently deciding the problem. We then provide adapted constrained decoding algorithms for \mri{} and \dlm{}. Finally, we show how to apply the algorithm to LLMs, where additional challenges arise from the need to handle tokens instead of terminals.

\vspace{-2mm}
\subsection{The Constrained Infilling Problem}\label{mri-decision}

\begin{wrapfigure}[10]{r}{0.37\linewidth}
    \vspace{-7.5mm}
    \resizebox{.37\textwidth}{!}{\
    \begin{minipage}{.45\textwidth}
        \begin{algorithm}[H]
        \caption{Constrained decoding}
        \label{alg:constrained-decision}
        \begin{algorithmic}[1]
            \Require Input $\mathbf{x}$, model $M$, target language~$L$
            \Ensure Completed output $\mathbf{x} \in L$
            \While{true}\algline{alg:constrained-decision:line:while}
                \State $\mathbf{x}' \sim M(\mathbf{x})$\algline{alg:constrained-decision:line:sample}
                \If{$\textsc{completable}(\mathbf{x}', L)$}\algline{alg:constrained-decision:line:check}
                    \State $\mathbf{x} \leftarrow \mathbf{x}'$\algline{alg:constrained-decision:line:update}
                    \If{$\hole \notin \mathbf{x}$}\algline{alg:constrained-decision:line:complete}
                        \State \Return $\mathbf{x}$\algline{alg:constrained-decision:line:return}
                    \EndIf
                \Else
                    \State reject $\mathbf{x}'$\algline{alg:constrained-decision:line:resample}
                \EndIf
            \EndWhile
        \end{algorithmic}
        \end{algorithm}
    \end{minipage}\
    }
\end{wrapfigure}
\paragraph{Constrained decoding with infilling} 
First, let us define a partial output $\mathbf{x}$ as a sequence of strings $x_i \in \Sigma^{*}$ interleaved with infilling regions $\hole \notin \Sigma$, i.e., $\mathbf{x} = x_1 \hole x_2 \dots \hole x_n$.
In constrained decoding, illustrated in \cref{alg:constrained-decision}, we complete $\mathbf{x}$ using model $M$ and target language $L$. We iteratively sample an updated partial output $\mathbf{x}'$ from $M$ (\cref{alg:constrained-decision:line:while,alg:constrained-decision:line:sample}).
All updated outputs are derived via \emph{additive} modifications to $\mathbf{x}$, meaning they either insert a string into infilling regions, e.g., insert $b$ into $a\hole{}c$ resulting in $a\hole{}b\hole{}c$, or remove a region by merging adjacent strings, e.g., converting $a\hole{}b\hole{}c$ to $a\hole{}bc$.
We then check whether the updated output can be completed into a valid word in $L$ (\cref{alg:constrained-decision:line:check}). 
If not, we reject the update and remove it from the model distribution, preventing the loop from resampling the update (\cref{alg:constrained-decision:line:resample}).
However, if the update is completable, we replace $\mathbf{x}$ with $\mathbf{x'}$ (\cref{alg:constrained-decision:line:update}) and return the output if the update removes the last infilling region (\cref{alg:constrained-decision:line:complete,alg:constrained-decision:line:return}). This is valid since $\mathbf{x}$ is both completable and has no infilling regions, implying $\mathbf{x} \in L$.
\rev{Since completability is preserved in updates, there always exists a series of additive updates that completes $\mathbf{x}$ to be in $L$}.
\vspace{-1mm}

\paragraph{Deciding update validity}  To enable constraining additive generation, we need an incremental verifier $\textsc{completable}$ to determine whether the regions in a partial output can be filled to produce a valid output in $L$. We formalize the decision problem solved by $\textsc{completable}$ as follows:
\begin{definition}[Constrained infilling problem]
    For a language $L$, partial output $\mathbf{x} = x_1 \hole x_2 \dots \hole x_n$ with $x_i \in \Sigma^{*}$ and $\hole$ denoting infilling regions, the constrained infilling problem asks whether there exists a list of $n-1$ words $\mathbf{y} = (y_1, \dots, y_{n-1})$ such that $w = x_1 \cdot{} y_1 \cdot{} x_2 \cdot{} \dots \cdot{} y_{n-1} \cdot{} x_n$ is in $L$.
\end{definition}
Thus, with the incremental verifier deciding the constrained decision problem, we have effectively reduced constrained decoding with infilling to the constrained infilling problem.
\vspace{-2mm}

\paragraph{Applications of the constrained infilling problem} We now reduce constrained decoding for \mri{} and \dlm{} generation to the constrained infilling problem.
For \mri{}, the list of words corresponds to the list of fixed strings $x_i$, with infilling regions in between. For the \dlm{} setting, we add implicit $\varepsilon$ tokens at the beginning and end of the partially filled sequence and then merge all consecutive non-mask tokens to build $\mathbf{x}$. For example, the sequence $(a,\mask,\mask,b,c,\mask)$ becomes $\mathbf{x} = a\hole bc\hole \varepsilon$.
Note that, similar to prior work \citep{dominosDblp2403,ugare2024syncode}, we slightly overapproximate the space of possible completions in these representations by allowing infillings of arbitrary size. In practice, there might be practical limitations to the number of tokens an LLM could insert. We discuss this in more detail in \cref{sec:discussion}.

\vspace{-2mm}
\subsection{Deciding the Constrained Infilling Problem Efficiently}\label{sec:deciding-mri}
\vspace{-2mm}

\newcommand{\cx}{\ensuremath{C_\mathbf{x}}}
\paragraph{Overview} We now give a brief overview of how to solve the constrained infilling problem efficiently. The problem is determined by two separate constraints: (1) the structural constraints on the output, described by the context-free language $L$, and (2) all possible completions of the partial output $\mathbf{x}$, which form a language \cx{}. For example, $L$ could be the language of syntactically valid C++ programs, and \cx{} the language of completions of partial program $\mathbf{x}=$\texttt{int$\hole$()\{$\hole$2;\}}.
The infilling problem is answered positively if and only if the intersection $L_\cap = L \cap \cx$ is not empty, i.e., some infilling of the partial output exists to generate a valid word in $L$.
We will show that \cx{} is a regular language that we can describe with a simple DFA, and that $L_\cap$ can be described by a context-free grammar, which we can construct from $L$'s grammar and \cx{}'s DFA. The constrained infilling problem is then reduced to checking whether $L_\cap$ is empty, for which we design an efficient algorithm.  In the example, a word in the intersection language is \texttt{int main() \{return 2;\}}.


\vspace{-2mm}
\paragraph{Constructing the regular language} 
\label{sec:constructing-reg}
The language \cx{} of all possible completions of $\mathbf{x} = x_1\hole{}\dots\hole{}x_n$ contains all words that start with $x_1$, end with $x_{n}$, and contain the strings $x_i$ ($1 \leq i \leq n$) in the correct order, with arbitrary symbols in between. We prove that \cx{} is regular by constructing an NFA that accepts \cx{}. We first construct automata $D_i$, which accept exactly $x_i$. Then, we concatenate $D_i$ with an additional state $q_i$ that accepts any string in $\Sigma^{\ast}$, i.e., $\delta(q_i, \sigma) = q_i$ for all $\sigma \in \Sigma$. For the concatenation, we add an $\varepsilon$-edge from the accepting states of $D_i$ to $q_i$ and from $q_i$ to the start state of $D_{i+1}$.
A visualization for the prior example is shown in \cref{fig:constructed-dfa}.
In our algorithm, we construct this NFA for each update. We then transform it into an equivalent DFA and minimize the DFA using standard methods~\citep{hopcroft1979sensitive}, as shown in \cref{fig:running-example}\textcolor{linkcolor}{b}.

\vspace{-2mm}
\paragraph{Constructing the intersection language} We leverage the well-established facts that (a) the intersection $L_\cap$ of CFL $L$ and regular language \cx{} is a CFL, whose grammar can be constructed from $L$'s grammar $G$ and \cx{}'s DFA, and (b) that the emptiness of a CFL can be checked in time polynomial to the size of the grammar \citep{gasarch-cfgreg,hopcroft1979sensitive}. \rev{The symbols in the intersection language have the form $\triple{p}{A}{q}$ for $p,q \in \Sigma$ and $A \in V$, where each symbol intuitively represents deriving a word from $A$ that also traverses the DFA from state $p$ to $q$. The language is nonempty if we can derive a word from $\triple{q_0}{S}{q_f}$ for start symbol $S$ and initial and final state $q_0$ and $q_f$. An example of deriving a word in the intersection language is shown in \cref{fig:running-example}\textcolor{linkcolor}{c}.
The intersection grammar $G_{\cap} = (V_{\cap}, \Sigma, P_{\cap}, S_{\cap})$ will have a cubic size in nonterminals and productions, with $|V_{\cap}| \in O(|V||Q|^2)$ and $|P_{\cap}| \in O(|P||Q|^3 + |P||Q|^2|\Sigma|)$ \citep{gasarch-cfgreg,hillel1962formal}. While we can not reduce the worst case complexity of this blowup, we carefully construct the intersection language to keep its size at a minimum, and employ several heuristics to reduce the practical cost of determining its emptiness, explained next.}

\vspace{-2mm}
\paragraph{Efficient normalization} The standard intersection algorithms require $G$ to be transformed to Chomsky normal form, which only allows rules of the form $A \rightarrow BC$ or $A \rightarrow a$, where $A, B, C \in V$ and $a \in \Sigma$ \citep{hopcroft1979sensitive}. The resulting grammar may have a quadratic increase in the number of production rules \citep{lange2009cnf}. To avoid this increase, we extend the standard construction to support CFGs in \ctwof, a normal form that additionally allows productions of the form $A \rightarrow \varepsilon$ and $A \rightarrow B$. \rev{We provide an example of the normalized C++ grammar in \cref{fig:running-example}\textcolor{linkcolor}{a}.} This normal form can be obtained with only a linear increase in production rules \citep{lange2009cnf}. Our adaptations to the standard intersection algorithm and a proof of its correctness are provided in \cref{app:proof-intersection}. In \cref{app:optimizations}, we describe several further heuristics to reduce the size of the normalized CFG of $G$.
\rev{After this step, we can intersect the languages and determine the emptiness of the intersection language.}

\vspace{-2mm}
\paragraph{Avoiding nongenerating nonterminals}
\rev{The standard algorithm to determine whether language $L$ is empty determines whether the start symbol $S$ is \emph{generating}, i.e., whether there is a sequence of production applications $S \rightarrow \dots \rightarrow w$ such that $w \in \Sigma^{\ast}$. This property can be decided in time linear to the size of $L$ \citep{hopcroft1979sensitive}.
When applied to intersection language $L_{\cap{}}$, it is important to note that intersection languages contain a significant fraction of non-generating symbols \citep{nederhof2008probabilistic,hanneforth-2011-practical}}. We therefore adopt a bottom-up search, that by construction only explores generating symbols \citep{sipser1996introduction}, and adapt it to \ctwof{}. The algorithm starts with symbols that generate terminals or empty strings directly, i.e., all $A$ with productions $A \rightarrow \sigma$ and $A \rightarrow \epsilon$. It marks these symbols as generating and inserts them into a queue. Next, for each symbol $X$ in the queue, it checks whether some production has $X$ on the right-hand side (i.e., either $A \rightarrow XC$, $A \rightarrow BX$ or $A \rightarrow X$), and whether the other symbol ($B$ or $C$) in the production was previously marked as generating. If so, $A$ is marked as generating and added to the queue. As soon as the start symbol $S$ is marked, we conclude that the language is non-empty.\rev{We confirm in our experiments that this avoids exploring $98\%$ to $99.99\%$ of productions in $L_{\cap{}}$.}

\vspace{-2mm}
\paragraph{Searching through the implicit intersection language} \rev{To speed up the emptiness check, we avoid constructing the entire intersection language. Instead, we only construct the parts of the language visited during the search.}
All symbols in the intersection language have the form $\triple{p}{A}{q}$ for $p,q \in \Sigma$ and $A \in V$. All production rules in the intersection grammar are directly derived from corresponding rules in the original CFG. Specifically, all rules of the form $\triple{p}{A}{q} \rightarrow \varepsilon$ and $\triple{p}{A}{q} \rightarrow \sigma$ are based on corresponding rules $A \rightarrow \varepsilon$ and $A \rightarrow \sigma$ in the original grammar without further dependencies, allowing us to iterate over directly generating symbols without constructing the remaining grammar. Further, all other rules are of the form $\triple{p}{A}{q} \rightarrow \triple{p}{B}{r}\triple{r}{C}{q}$ and $\triple{p}{A}{q} \rightarrow \triple{p}{B}{q}$ based on original productions $A \rightarrow BC$ and $A \rightarrow B$, for all $p,q,r \in Q$. This enables enumerating all such rules for a given $\triple{p}{B}{r}$, $\triple{r}{C}{q}$ or $\triple{p}{B}{q}$ during the search. We present the corresponding pseudo-code and additional explanations in \cref{app:search-algorithm}.

\vspace{-2mm}
\paragraph{Sampling a valid completion from the intersection language} 
The algorithm presented above decides intersection emptiness. We now extend it to return a valid completion from the intersection language. To achieve this, we modify the algorithm to track production rules that were applied when marking symbols as generating. These rules describe a parse tree for some word $w$ in the intersection language. We traverse the terminals at the leaf nodes of this tree from left to right to reconstruct a valid completion in the intersection language. This completion is used after a fixed number of rejected updates from the LLM. Since the algorithm leverages the results from the prior emptiness search, it can be run at no additional cost.

\begin{figure}
    \centering
    \def\csurround#1#2#3%
  {\mathop{}%
   \mathopen{\vphantom{#2}}^{#1}%
   \kern\scriptspace%
   #2%
   \kern4\scriptspace%
   \mathopen{\vphantom{#2}}^{#3}%
   }


\newcommand{\ctriple}[3]{\overset{#1 \rightarrow #3}{#2}}
\begin{tikzpicture}[
    node distance=6mm and 8mm,
    base/.style={
        draw=gray,
        fill=lightgray,
        rounded corners=3pt,
        font=\sffamily,
        blur shadow={shadow xshift=0.5pt, shadow yshift=-0.5pt, shadow opacity=20}
    },
    darkbox/.style={
        draw=gray,
        fill=lightgray!20,
        rounded corners=3pt,
        minimum width=6.3cm,
        minimum height=3.5cm,
        anchor=west,
        xshift=0.3cm
    },
    codebox/.style={
        base,
        minimum height=0.5cm,
        minimum width=3cm,
        align=center
    },
    proposalbox/.style={
        base,
        minimum height=0.4cm,
        minimum width=5.7cm,
        align=center
    },
    optionbox/.style={
        base,
        minimum width=2.8cm,
        minimum height=0.4cm,
        font=\ttfamily\tiny
    },
    syntaxbox/.style={
        base,
        fill=accentblue!20,
        draw=accentblue,
        minimum width=3cm,
        font=\ttfamily\tiny,
        align=left,
        text width=2.6cm
    },
    rulebox/.style={
        base,
        minimum width=2.8cm,
        minimum height=0.4cm,
        font=\ttfamily\tiny,
        text width=2.1cm
    },
    cfgbox/.style={
        base,
        fill=accentblue!20,
        draw=accentblue,
        minimum width=0.6cm,
        font=\tiny,
        align=center,
        anchor=center,
        text width=0.5cm,
    },
    placeholder/.style={
        fill=gray2,
        rounded corners=1pt,
        inner xsep=1pt,
        inner ysep=1.5pt,
        font=\ttfamily\tiny,
        text height=1ex,
        text depth=0.25ex,
        anchor=base west
    },
    placeholder2/.style={
        fill=gray2,
        rounded corners=1pt,
        inner xsep=1pt,
        inner ysep=1.5pt,
        font=\ttfamily\tiny,
        color=white,
        text height=1ex,
        text depth=0.25ex,
        anchor=base west
    },
    resultbox/.style={
        base,
        minimum size=0.8cm,
        font=\Large\bfseries,
    },
    accept/.style={resultbox, fill=lightgreen, draw=darkgreen, text=darkgreen},
    reject/.style={resultbox, fill=lightred, draw=darkred, text=darkred},
    arrow/.style={
        ->,
        thick,
        line width=0.4mm,
        color=blue!80, 
        >={Triangle[scale=0.6]}
    },
    derivnode/.style={
        font=\sffamily\footnotesize,
        anchor=center
    },
    derivarrow/.style={
        ->,
        line width=0.3mm,
        color=gray!60,
        >={stealth}
    }
]


\node[syntaxbox, text width=2.5cm, inner xsep=2mm, align=left] at (0,-0.4cm) (cfg) {
    S \(\rightarrow\) Def S$_1$ \\
    S$_1$ \(\rightarrow\) T$_{(}$ S$_2$ \\
    S$_2$ \(\rightarrow\) T$_{)}$ S$_3$ \\
    S$_3$ \(\rightarrow\) T$_{\{}$ S$_4$ \\
    S$_4$ \(\rightarrow\) Lines T$_{\}}$ \\
    Lines \(\rightarrow\) Line Lines$_1$ \\
    Lines$_1$ \(\rightarrow\) T$_{;}$ Lines \\
    Lines \(\rightarrow\) \(\varepsilon\) \\
    Def \(\rightarrow\) T$_{\lexeme{type}}$ Def$_1$ \\
    Def$_1$ \(\rightarrow\) T$_{\lexeme{name}}$ \\
    \hspace{0.1mm}\\
    T$_{(}$ \(\rightarrow\) (\\
    T$_{\{}$ \(\rightarrow\) \{\\
    T$_{\}}$ \(\rightarrow\) \} \\
    T$_{;}$ \(\rightarrow\) ; \\
    T$_{\lexeme{type}}$ \(\rightarrow\) \lexeme{type} \\
    T$_{\lexeme{name}}$ \(\rightarrow\) \lexeme{name} \\
} ;
\node[yshift=0.05cm, font=\tiny, anchor=south west, text=gray] (cfgtext) at (cfg.north west) {\emph{(a) Normalized CFG} (\ctwof)};

\begin{scope}[shift={(cfg.north east)}, xshift=1.3cm, yshift=-0.25cm, scale=0.55, every node/.style={transform shape}]
    \begin{scope}[->, >=stealth', auto, semithick, node distance=12mm]
        \node[state, initial left, initial text=] (d0) {$d_0$};
        \node[state, right=of d0] (d1) {$d_1$};
        \node[state, right=of d1] (d2) {$d_2$};
        \node[state, right=of d2] (d3) {$d_3$};
        \node[state, right=of d3] (d4) {$d_4$};
        \node[state, right=of d4] (d5) {$d_5$};
        \node[state, right=of d5] (d6) {$d_6$};
        \node[state, right=of d6, accepting] (d7) {$d_7$};

        \path (d0) edge node [above] {\lexeme{type}} (d1)
            coordinate[pos=1] (d0tod1);
        \path (d1) edge [loop below] node {$\Sigma^{*}$} (d1);
        \path (d1) edge node [above] {(} (d2);
        \path (d2) edge node [above] {)} (d3);
        \path (d3) edge node [above] {\{} (d4);
        \path (d4) edge [loop below] node {$\Sigma^{*}$} (d4);
        \path (d4) edge node [above] {\lexeme{int}} (d5);
        \path (d5) edge node [above] {;} (d6);
        \path (d6) edge node [above] {\}} (d7);
    \end{scope}

\end{scope}
\node[yshift=0.1cm, xshift=-0.5cm, font=\tiny, anchor=south west, text=gray] at (d0.north west) (mindfacap){(b) \emph{Minimized DFA} of all possible completions} ;

\begin{scope}[shift={(d3.south)}, yshift=-3.3cm,scale=0.7, every node/.style={transform shape},font=\tiny]
    \matrix (S) [matrix of nodes,
                 nodes={derivnode, anchor=center, transform shape, font=\sffamily\tiny,scale=0.9},
                 column sep=0mm] at (0,0) {
        $\ctriple{d_0}{\text{Def}}{d_1}$ & $\ctriple{d_1}{\text{S}_1}{d_7}$ \\
    };
    \node[derivnode, right=27mm of S] (root) {$\ctriple{d_0}{\text{S}}{d_7}$};

    \matrix (Def) [matrix of nodes,
                   nodes={derivnode, anchor=center, transform shape, font=\sffamily\tiny,scale=0.9},
                   column sep=0mm,
                   above=6mm of S-1-1,
                   xshift=-15mm] {
        $\ctriple{d_0}{\text{T}_{\lexeme{type}}}{d_1}$ & $\ctriple{d_1}{\text{Def}_1}{d_1}$ \\
    };
    \node[derivnode, above=5mm of Def-1-2] (Def1) {$\ctriple{d_1}{\text{T}_{\lexeme{name}}}{d_1}$};
    \node[derivnode] (<name>) at ($(d1 |- Def1)+(0,6mm)$) {\lexeme{name}};
    \node[derivnode] (<type>) at ($(d0 |- <name>)+(0,-.2mm)$) {\lexeme{type}};

    \node[derivnode, above right=6mm and 1mm of S-1-2, inner sep=8pt] (S3) {$\dots$};
    \matrix (S4) [matrix of nodes,
                  nodes={derivnode, anchor=center, transform shape, font=\sffamily\tiny,scale=0.9},
                  column sep=0mm,
                  above=6mm of S3] {
        $\ctriple{d_4}{\text{Lines}}{d_6}$ & $\ctriple{d_6}{\text{T}_{\}}}{d_7}$ \\
    };
    \node[derivnode, inner sep=5pt,yshift=5mm] (Lines) at ($(d4 |- S4-1-1)+(1,5mm)$) {\dots}; 
    \node[derivnode,yshift=5mm] (<brackclose>) at ($(d6 |- S4-1-2)+(0,5mm)$) {\}};

    \draw[derivarrow] (S) -- (root);
    \draw[derivarrow] (Def) -- (S-1-1);
    \draw[derivarrow] (<type>) -- (Def-1-1);
    \draw[derivarrow] (Def1) -- (Def-1-2);
    \draw[derivarrow] (<name>) -- (Def1);
    \draw[derivarrow] (S3) -- (S-1-2);
    \draw[derivarrow] (S4) -- (S3);
    \draw[derivarrow] (Lines) -- (S4-1-1);
    \draw[derivarrow] (<brackclose>) -- (S4-1-2);
\end{scope}
\node[font=\tiny, text=gray, below=4.5cm of mindfacap.west,anchor=west] {(c) \emph{Derivation of word in $L_{\cap}$}};

\end{tikzpicture}
    \caption{\rev{Examples of \cref{fig:overview,fig:background} processed during our method. (a) The grammar is first normalized into \ctwof{}, and (b) the NFA is transformed into a minimal DFA. (c) To determine emptiness of $L_{\cap}$, the algorithm then searches the initial state $\triple{d0}{S}{d7}$ through the productions in reverse, starting from the terminals.}}
    \label{fig:running-example}
\end{figure}

\subsection{Application of Constrained Infilling to LLMs}\label{sec:lexing}
\vspace{-2mm}

We now briefly outline how to apply the algorithm from \cref{sec:deciding-mri} to LLMs, which generate arbitrary Unicode text rather than language terminals. Full details are provided in \cref{app:lexing}.

\vspace{-2mm}
\paragraph{Lexing}
For typical applications of CFGs, a string of Unicode characters $u$ is converted to terminals $x = t_1\dots{}t_k$ in a process called \emph{lexing}. First, note that every terminal $t$ corresponds to a regular language $R_t$ over Unicode characters. During lexing, $t_1$ is obtained by finding the terminal $t$ such that $R_t$ matches a prefix $p$ of $u$, i.e., $u = p \cdot s$. The lexing process then recurses on $u' = s$ to obtain the remainder of $x$, continuing until the string is empty. In principle, to apply this procedure to a sequence with infilling regions $s_1 \mask{} s_2 \mask{} \dots s_k$, we would lex each consecutive string $s_i$ to obtain $\mathbf{x} = t_1 \hole \dots \hole t_n$. \rev{For example, \ttt{int \mask{} () \{ \mask{} 2;\}} would be lexed to \ttt{\lexeme{type} \hole{} ( ) \{ \hole{} \lexeme{int} ; \}}. However, several caveats to this procedure need to be addressed.}

\vspace{-2mm}
\paragraph{Handling infilling regions} 
First, it does not accurately handle partial terminals that border infilling regions, since LLM tokens are Unicode strings that may not align with terminals. \rev{For example, the partial LLM output \texttt{\mask{}2} could correspond to both \texttt{\lexeme{int}} and \texttt{\lexeme{ident}}. The ambiguity stems from the possibility to fill the gap with, e.g., either the token \ttt{x} or \ttt{1}, resulting in \ttt{x2} or \ttt{12} respectively.}

To address this, we treat the text around an infilling region as possibly belonging to an incomplete terminal. Specifically, in the lexing process, we additionally look for terminals $t$ such that the current output $u$ is a prefix of a word in $R_t$ right before an infilling region or a suffix right after a region. Further, we include terminals $t$ that could span across one or more infilling regions by determining if prefixes and suffixes can be infilled to form a single word in $R_t$, as in the example above. We can thus generate all terminal sequences consistent with a partial output. If any such sequence can be completed to a valid program, then the partial output itself admits a valid completion. \rev{In the example above, our algorithm would yield two possible lexings, both \ttt{\hole{} \lexeme{int}} and \ttt{\hole{} \lexeme{ident}}, and intersect both with the context-free grammar.}.

\vspace{-2mm}
\paragraph{Efficiency optimizations}
The number of possible terminal sequences grows quickly with the number of regions and ambiguities. To improve efficiency, we introduce two optimizations. First, for each $x$, we directly construct a single NFA for all possible terminal sequences. This allows us to apply the intersection algorithm once rather than for each sequence. Second, we reduce ambiguity by preprocessing terminals: whenever the accepted language of terminal $t_{\leq}$ is contained within terminal $t_{\geq}$, we remove the overlap from $t_{\geq}$ and adapt the CFG to allow $t_{\leq}$ wherever $t_{\geq}$ is allowed.

\vspace{-2mm}
\paragraph{Sampling a valid completion} The sampling method from \cref{sec:deciding-mri} returns a sequence of terminals rather than Unicode characters. To sample a Unicode completion, we first concatenate the regular languages of the terminals in the sampled completion. We then construct a regular language for the current partial LLM output and intersect the two languages. Sampling a random string from this intersection yields a valid completion at the Unicode level. \rev{In the given example, we would construct regular language of terminals \ttt{$R$(\lexeme{type}) $R$(\lexeme{name}) $R$(\ttt{(}) $R$(\ttt{)}) $R$(\ttt{\{}) $R$(\lexeme{int}) $R$(\ttt{\}})} and intersect it with \ttt{int .* ()\{ .* 2;\}}, resulting in \ttt{int x()\{2;\}}.}

\vspace{-2mm}
\rev{\subsection{Soundness, Completeness, and Alignment}}

\vspace{-2mm}
In this subsection, we briefly analyze desirable properties of our algorithm. We first show that the algorithm is sound and complete with respect to the constrained grammar, and fulfills the minimal invasiveness guarantees introduced by \citet{dominosDblp2403}.

\vspace{-2mm}
\paragraph{Soundness and Completeness}
Our algorithm is sound, i.e., all generated output is valid according to the formal grammar and lexer. This requires the assumption that the lexer of the target language uses maximal munch for lexing, as is common in many programming languages \citep{park2024grammar,melcer2024constraineddecodingfillinthemiddlecode}. Moreover, our algorithm is complete, i.e., it allows sampling any token that would result in a correct output. A detailed proof of these properties is provided in \cref{app:lexing}, together with a more precise description of the employed lexing algorithm.

\vspace{-2mm}
\paragraph{Minimally invasive} 
Our algorithm is \emph{minimally invasive} \citep{dominosDblp2403}. This means that, if the model $M$ without constraints would generate a valid output $w \in L$, it will also produce the valid output when our constraining algorithm is applied. This follows from the algorithm's completeness; it does not reject any partial variants of a valid output. Therefore, in \cref{alg:constrained-decision} for the case where $M$ suggests only valid partial outputs, each $\mathbf{x'}$ would be marked as completable, and the algorithm thus would return the same output as if the check was not present.

\section{Experimental Evaluation}
\label{sec:experiments}

\vspace{-2mm}
We evaluate our method across a range of tasks and models, first in the \mri{} setting, and then in \dlm{}, demonstrating improvements in both syntactic and functional correctness.  We provide further experimental details, ablate \dlm{} diffusion steps, and provide a case study in \appcref{app:experiments}.
\subsection{Experimental Setup}
\label{sec:experiments-setup}
\paragraph{Metrics}
We compute two main metrics to evaluate the effectiveness of our method. First, we determine the percentage of syntactically correct completions (Syntax), which indicates how many of the obtained completions adhere to the specified grammar.
We also measure functional correctness (Functional) by either comparing the sample to a golden solution, or by reporting the percentage of solutions that pass all test cases, pass@1, depending on the dataset. All results are averaged over four independent runs with different seeds. We compute confidence intervals at $95\%$, \textbf{boldface} the best method, and \underline{underline} all methods over which the increase is not significant. The usual size of the confidence interval is $1\%$ to $2\%$.

\vspace{-2mm}
\paragraph{Compared methods} We run unconstrained LLM sampling, reported as Vanilla (\emph{\vanilla}) and constrained decoding with our method (\emph{\constrplus}). This includes sampling random completions when generation aborts. As an ablation, we report \emph{\constrained}, where these aborted instances are marked as syntactically and functionally invalid. \rev{To decide the cutoff for aborting generation, we run our method on a development set of \humaneval{}, \json{} and \smiles{} tasks. We observe that if a task requires more than $50$ resamples, it is functionally correct in only $0.7\%$ of cases. Thus, we significantly speed up the sampling process without losing performance by aborting after \maxresample{}. Further, all tested tasks can be solved within $256$ tokens, which we set as a maximum output size for all methods. On the instruction tuned \dlm s, we further compare to the baseline of \emph{Grammar Prompting (\grammarprompting)} \citep{wang2023grammarpromptingdomainspecificlanguage}, where the model is provided with the Grammar Rules in its prompt.}

\vspace{-2mm}
\subsection{Fill-In-the-Middle and Multi-Region-Infilling}
\label{sec:exp:fim-mri}

\vspace{-2mm}
\paragraph{Models} We compare the performance of five recent open-weight infilling models, including \starcodertwo{} \citep{starcoder2}, \codegemma{} \citep{codegemma}, and the \deepseekcoder{} Family \citep{guo2024deepseekcoder}, covering 7B parameter models from three distinct model families and model sizes from 1.3B to 33B.

\vspace{-2mm}
\paragraph{Tasks and benchmarks}
Infilling is commonly used to complete partial code \citep{bavarian2022efficienttraininglanguagemodels-fim,fried2023incodergenerativemodelcode-multiregion}.
We therefore evaluate our method on the C++ translation of the HumanEval dataset \citep{zheng2023codegeex-humaneval-x,humaneval}, of 164 diverse basic coding tasks. Similar to \citet{bavarian2022efficienttraininglanguagemodels-fim}, we construct an infilling dataset by removing random spans from the human-written reference implementation. We evaluate up to three removed spans, resulting in \mrione{}, being equivalent to \fim{}, and \mritwo{} and \mrithree{}, with two and three infilling regions respectively. 
We design a CFG for the subset of C++ syntax needed to solve the tasks in HumanEval. We report adherence to this CFG as syntactic correctness. Functional correctness is measured by computing the pass@1 score on provided test cases \citep{brown2020fewshot}. \rev{In \cref{app:experiments-add}, we also evaluate our method when removing specific lines of code rather than random spans, observing similar improvements.}

\vspace{-2mm}
\paragraph{Syntactic correctness}
As shown in \cref{tab:infilling-results:syntax}, our method increases syntactic correctness significantly across all models and numbers of infilling regions. Our method (\constrplus) recovers a syntactically valid completion in on average $95.8\%$ of instances. Remaining errors are due to timeouts. Constrained decoding without completions  (\constrained) increases syntactic correctness more for code with multiple regions. This coincides with models struggling more, achieving an absolute increase of $5.2\%$, $22.5\%$, and $31.5\%$ for \mrione{}, \mritwo{}, and \mrithree{}, respectively. These improvements are consistent across model families and sizes, ranging between $17\%$ and $21\%$ per model.

\vspace{-2mm}
\paragraph{Functional correctness}
In the lower half of \cref{tab:infilling-results:pass1}, we observe that constraining (\constrplus{}) consistently increases functional correctness, on average by $2.8\%$, and even without randomly sampling valid completions (\constrained{}), the average increase is $2.4\%$. This is expected, as syntactically incorrect completions can not be functionally correct and are effectively prevented by our method.

\vspace{-2mm}
\paragraph{Runtime overhead}
We compare the time per token between constrained and vanilla decoding. \rev{The median runtime overhead of constrained decoding is \SI{4.2}{ms}, where the overhead on the small \deepseekcodertwob{} is higher (\SI{5.8}{ms}) than on the 7B models (\SI{4.6}{ms}) and \deepseekcoderthirtyb{} (\SI{4.3}{ms}). Moreover, median overhead increases with more complex infilling, growing from \SI{3.1}{ms} on \mrione{} to \SI{7.7}{ms} on \mrithree{}.} Further details on runtime are provided in \cref{app:experiments-add}.

\vspace{-2mm}
\begin{table}
  \centering
    \renewcommand{\arraystretch}{1.0}
    \setlength{\tabcolsep}{3pt}
    \caption{Our method consistently improves the percentage of syntactically and functionally correct infillings for varying numbers of regions in \mri{} under standard decoding (\vanilla), constrained decoding (\constrained), and completing partially completed outputs (\constrplus).}
    \label{tab:infilling-results:syntax}
    \label{tab:infilling-results:pass1}
    \begin{tabular}{@{}ll @{}p{7mm}@{} ccc @{}p{7mm}@{} ccc @{}p{7mm}@{} ccc @{}}
        \toprule
       &&&   \multicolumn{3}{c}{\mrione} &&   \multicolumn{3}{c}{\mritwo}             &&  \multicolumn{3}{c}{\mrithree}             \\
\cmidrule{4-6} \cmidrule{8-10} \cmidrule{12-14}
 & Model        &&   \vanilla{} &   \constrained{} & \constrplus{} &&\vanilla{} &   \constrained{} & \constrplus{} &&  \vanilla{} &   \constrained{} & \constrplus{} \\
\midrule
\parbox[t]{3mm}{\multirow{5}{*}{\rotatebox[origin=c]{90}{Syntax}}}
& \starcodertwo     &           &        88.2 &       95.0 & \textbf{98.9}  &    &        55.4 &       77.7 & \textbf{96.3} &    &        24.5 &       57.2 & \textbf{88.3} \\
& \codegemma        &           &        92.5 &       97.2 & \textbf{100.0} &    &        61.5 &       85.6 & \textbf{99.0} &    &        29.9 &       66.4 & \textbf{96.0} \\
& \deepseekctwob    &           &        86.5 &       91.7 & \textbf{98.7}  &    &        51.5 &       72.9 & \textbf{93.1} &    &        22.7 &       47.7 & \textbf{83.0} \\
& \deepseekcsevenb  &           &        93.9 &       98.3 & \textbf{100.0} &    &        62.0 &       84.0 & \textbf{97.3} &    &        32.9 &       64.9 & \textbf{94.6} \\
& \deepseekcthirtyb &           &        93.1 &       97.6 & \textbf{100.0} &    &        66.3 &       86.5 & \textbf{97.8} &    &        36.4 &       67.8 & \textbf{93.5} \\
 \midrule
\parbox[t]{3mm}{\multirow{5}{*}{\rotatebox[origin=c]{90}{Functional}}}
& \starcodertwo     &           & 53.8             & \underline{56.1} & \textbf{56.3} &    &        20.5 & 23.7             & \textbf{24.2} &    & 7.5              & \underline{10.3} & \textbf{11.0} \\
& \codegemma        &           & 57.1             & \textbf{59.6}    & \textbf{59.6} &    &        24.8 & \underline{29.0} & \textbf{29.2} &    & 8.7              & \underline{12.6} & \textbf{12.8} \\
& \deepseekctwob    &           & \underline{46.5} & \underline{46.4} & \textbf{47.2} &    &        16.1 & 18.4             & \textbf{19.2} &    & 4.9              & 5.4              & \textbf{6.5}  \\
& \deepseekcsevenb  &           & 64.8             & \underline{67.1} & \textbf{67.3} &    &        29.8 & \underline{32.7} & \textbf{33.2} &    & \underline{11.9} & \textbf{13.5}    & \textbf{13.5} \\
& \deepseekcthirtyb &           & 69.8             & \underline{71.2} & \textbf{71.4} &    &        29.8 & \underline{34.0} & \textbf{34.3} &    & 12.6             & 14.3             & \textbf{15.4} \\
 \bottomrule
\end{tabular}
\end{table}

\subsection{Diffusion Language Models}

\vspace{-2mm}
\paragraph{Models} We evaluate our method on the instruction-tuned versions of four state-of-the-art diffusion language models, \llada{} \citep{nie2025llada}, \dream{} \citep{dream2025}, \dreamcoder{} \citep{dreamcoder2025} and \diffucoder{} \citep{gong2025diffucoder}. We run all models with 32 steps on 256 tokens and with a temperature of 0.2.

\vspace{-2mm}
\paragraph{Tasks and benchmarks}
As \dlm{}s are generic text generation models with many different applications, we design three distinct and diverse tasks:

\begin{tabular}{@{}lp{.85\textwidth}@{}}
\humaneval & Based on the dataset used in \cref{sec:exp:fim-mri}, the model should generate the entire function specified in natural language \citep{humaneval,zheng2023codegeex-humaneval-x}. \\\noalign{\vspace{0.5em}}
\json & The model should extract relevant information from natural language input, adhering to a JSON-Schema specification \citep{nousresearch-json-mode-eval}. \\\noalign{\vspace{0.5em}}
\smiles & The model should write down a chemical molecule described in natural language in the SMILES specification language \citep{smiles}.
\end{tabular}

For \smiles{} and \json{} we generate synthetic benchmarks using \gemini{} \citep{deepmind2025geminipro} with verification to ensure that the generated samples are correct and solvable, resulting in 167 and \lexfix{269} instances respectively. 
We implement the syntax of each language as a CFG and use it to enforce and evaluate the syntactic correctness of the generated output. For \humaneval{}, we measure functional correctness using pass@1 as in \cref{sec:exp:fim-mri}. For \json{} and \smiles{}, correctness is evaluated by normalizing and comparing to a golden solution. \rev{More details about the dataset generation procedure and the correctness evaluation are provided in \appcref{app:experiments-data}.}

\vspace{-2mm}
\paragraph{Syntax errors}
\rev{We observe that our method consistently increases syntactic correctness for all tasks and models, as shown in \cref{tab:diffusion-results:syntax}, in stark contrast to the mixed impact of the non-constraining baseline (\grammarprompting{}).} Without sampling valid completions (\constrained), our method increases the percentage of syntactically correct instances by $16.1\%$, $14.7\%$, and $26.0\%$ for \humaneval{}, \json{}, and \smiles{}, respectively.  We observe that many models fail to generate syntactically correct output even under constraints, with, for example, only $19.0\%$ correct \humaneval{} generations for \dreamcoder{}. However, sampling valid completions (\constrplus) recovers the failed instances, increasing to $99.2\%$. In \json{}, constrained decoding with completion achieves $100\%$ syntactic correctness.

\vspace{-2mm}
\paragraph{Functional correctness}
As shown in the lower half of \cref{tab:diffusion-results:pass1}, the consistent positive effect of constraining on functional correctness is also present for \dlm{}, with an average increase in functional correctness without completions (\constrained) of $1.9\%$, and a slight additional boost with completions (\constrplus) to $2.2\%$. Notably, \dream{} performance on \json{} increases by \lexfix{$7.0\%$}. \rev{\dream{} appears to benefit a lot from being provided the target syntax (\grammarprompting{}) in \humaneval{} and \smiles{}, where it outperforms our constraining (\constrplus{}).} In the \smiles{} setting, where models generally perform very poorly at only $1.5\%$ average correctness, syntactic constraints are not able to improve functional correctness significantly, achieving only a modest average increase of $0.2\%$.

\vspace{-2mm}
\paragraph{Runtime overhead}
We compare the runtime to complete samples in constrained decoding with the vanilla setting.  \rev{The median completion overhead is only \SI{0.1}{s}. We observe both speed-ups of up to \SI{1}{s} and slowdowns of up to \SI{7.8}{s}.} Speed-ups occur when the decoding is preemptively aborted. Further details for this experiment are provided in \cref{app:experiments-add}.

\vspace{-2mm}
\paragraph{\rev{Comparison to DINGO}}
\rev{
DINGO \citep{suresh2025dingoconstrainedinferencediffusion} is a recently proposed method for constrained decoding of regular languages in \dlm s. In \cref{app:experiments:dingo}, we show that while our method is more general, being able to enforce context-free constraints, it achieves the same syntactic and similar functional correctness. Furthermore, our method has similar runtime overhead of less than \SI{0.3}{ms} on the tested tasks, without requiring preprocessing as opposed to DINGO.
}

\begin{table}
  \centering
    \renewcommand{\arraystretch}{1.0}
    \setlength{\tabcolsep}{3pt}
    \caption{Constrained decoding (\constrained) consistently increases the percentage of syntactically correct completions for \dlm{}s over standard decoding (\vanilla). \rev{Non-constraining baselines like Grammar prompting (\grammarprompting{}) do not consistently improve syntactic or semantic performance.}}
    \label{tab:diffusion-results:syntax}
    \label{tab:diffusion-results:pass1}
    \begin{tabular}{@{}ll @{}p{5mm}@{} cc@{\hspace{0.6mm}}c@{\hspace{0.5mm}}c @{}p{5mm}@{} cc@{\hspace{0.6mm}}c@{\hspace{0.5mm}}c @{}p{5mm}@{} cc@{\hspace{0.6mm}}c@{\hspace{0.5mm}}c @{}}
        \toprule
        &&&   \multicolumn{4}{c}{\humaneval} &&   \multicolumn{4}{c}{\json}             &&  \multicolumn{4}{c}{\smiles}             \\
\cmidrule{4-7} \cmidrule{9-12} \cmidrule{14-17}
& Model        &&   \vanilla{} & \rev{\grammarprompting} &  \constrained & \constrplus && \vanilla{} & \rev{\grammarprompting} &   \constrained & \constrplus &&   \vanilla{} & \rev{\grammarprompting} &   \constrained & \constrplus \\
\midrule
\parbox[t]{3mm}{\multirow{4}{*}{\rotatebox[origin=c]{90}{Syntax}}}
& \dream    &    &     40.5 &     48.6 &      58.7 & \textbf{99.4} &    &     \lexfix{22.7} &     \lexfix{15.9} &      \lexfix{45.4} & \textbf{100.0} &    &     67.7 &     84.2 &      93.7 & \textbf{99.4}  \\
& \dreamc   &    &     11.0 &      9.1 &      19.0 & \textbf{99.2} &    &     \lexfix{73.4} &     \lexfix{73.0} &      \lexfix{86.4} & \textbf{100.0} &    &     73.1 &     82.1 &      94.9 & \textbf{100.0} \\
& \llada    &    &     13.3 &      6.7 &      36.1 & \textbf{99.7} &    &     \lexfix{77.6} &     \lexfix{75.7} &      \lexfix{88.8} & \textbf{100.0} &    &     58.2 &     67.4 &      91.3 & \textbf{100.0} \\
& \diffuc   &    &     39.2 &     33.1 &      54.7 & \textbf{99.7} &    &     64.5 &     \lexfix{45.4} &      \lexfix{76.4} & \textbf{100.0} &    &     69.3 &     57.3 &      92.2 & \textbf{99.1}  \\
\midrule
\parbox[t]{3mm}{\multirow{4}{*}{\rotatebox[origin=c]{90}{Funct.}}}
& \dream    &    &      6.6 & \textbf{11.6} & 8.8             & 9.5           &    &      7.4 & \lexfix{6.7}              & \lexfix{11.5}             & \textbf{\lexfix{14.4}}    &    & 0.6             & \textbf{2.7}    & 1.1          & 1.1          \\
& \dreamc   &    &      3.7 & 0.6           & \underline{4.9} & \textbf{5.2}  &    &     \lexfix{44.3} & \underline{\lexfix{45.6}} & \textbf{\lexfix{46.5}}    & \textbf{\lexfix{46.5}}    &    & \textbf{3.4}    & 1.2             & \textbf{3.4} & \textbf{3.4} \\
& \llada    &    &      3.8 & 2.0           & \underline{5.0} & \textbf{5.3}  &    &     43.1 & \textbf{\lexfix{50.1}}    & \underline{\lexfix{49.3}} & \underline{\lexfix{49.3}} &    & \underline{0.7} & \underline{0.6} & \textbf{1.0} & \textbf{1.0} \\
& \diffuc   &    &     12.5 & 7.6           & 13.7            & \textbf{14.8} &    &     34.3 & \lexfix{21.2}             & 38.0             & \textbf{\lexfix{38.3}}    &    & \textbf{1.1}    & \underline{0.6} & \textbf{1.1} & \textbf{1.1} \\
 \bottomrule
\end{tabular}
\end{table}

\vspace{-2mm}
\section{Related Work}
\label{sec:related-work}

\vspace{-2mm}
\paragraph{Large language models}
LLMs have gained traction for diverse tasks such as code generation \citep{jiang2024surveylargelanguagemodels} and structured output generation \citep{langchain-structured-outputs,openai-structured-outputs,anthropic-json-mode}. While the most common approach trains LLMs for \ltr{} generation, many modern code models also support \fim{} settings \citep{guo2024deepseekcoder,lozhkov2024starcoder2stackv2,codegemma}. More recently, diffusion language models have been scaled to billion-parameter sizes and demonstrate promising performance on a variety of tasks \citep{nie2025llada,gong2025diffucoder,dreamcoder2025}. Meanwhile, LLMs are prone to errors during generation. For example, they often make mistakes in niche programming languages \citep{enhancing-code-generation} and fundamentally struggle to model specific types of formal languages \citep{transformers-formal-languages,dyck}.

\vspace{-2mm}
\paragraph{Constrained decoding}
\rev{Constraining LLM generation to context-free languages has been explored extensively in prior work \citep{dominosDblp2403,synchromesh,kellner2023lmql,willard2023efficientguidedgenerationlarge}. Most prior works apply these techniques to the \ltr{} setting \citep{synchromesh,dominosDblp2403,ugare2024syncode,dong2024xgrammar,sun2025earleydriven,park2025flexible}, with some extensions to \fim{} and context-sensitive features \citep{melcer2024constraineddecodingfillinthemiddlecode,mundler2025typeaware}.}
\citet{suresh2025dingoconstrainedinferencediffusion} constrain \dlm{}s specifically, but only to regular languages.
To our knowledge, constrained decoding with CFGs has not yet been applied to the \mri{} or \dlm{} paradigms.
Additionally, unlike prior work that employs masking \citep{ugare2024syncode,synchromesh,kellner2023lmql} or masking-rejection hybrids \citep{dong2024xgrammar,dominosDblp2403}, our rejection sampling approach incurs no additional latency before starting language inference, significantly reducing friction of switching to a different CFG.

\paragraph{Leveraging language intersections}
Two similar works leverage the intersection of CFLs and regular languages.
First, \citet{fazekas-subsequence-matching-2024} discuss subsequence matching, which asks whether $w$ is a subsequence of any word in language $L$. This is a special case of our decision problem, with $\mathbf{x} = \epsilon\hole{}w_1\hole{}\dots\hole{}w_{|w|}\hole{}\epsilon$, and can also be solved by using the emptiness check for intersection languages. Their work is not applicable to our setting, as it only handles this special case, does not consider practical performance, and does not consider how to handle lexing.

Second, \citet{nederhof2008probabilistic} use intersections of weighted CFGs and DFAs for parsing natural language words, using the intersection language as a succinct representation of admissible parses of lexeme sequences. To reduce the size of these intersections, they also filter non-generating symbols during the intersection construction.
\section{Conclusion}
\label{sec:conclusion}

We presented the first constrained decoding method for diffusion models that is able to handle context-free languages such as C++ and JSON. We showed how to reduce the problem of valid completion to an infilling decision problem solvable using formal language techniques. Our optimized algorithm demonstrates a consistent and significant increase in syntactic and functional correctness on a variety of benchmarks and models,  while still ensuring efficiency at inference time.

\clearpage
\section*{Reproducibility Statement}
We describe our implementation in detail in \cref{sec:experiments,app:experiments}, including details such as hyperparameters and the used compute hardware. Further, all of our experiments were run with fixed seeds and disabled optimizations that would introduce nondeterminism. To ensure complete reproducibility of our results, we publicly release the code implementation of our method, as well as datasets, models, and code used for the evaluation at \url{https://github.com/eth-sri/constrained-diffusion}. We also include the content of this released code as an anonymized artifact for the double-blind review.

\section*{Acknowledgements}
This work has been done as part of the grant SAFEAI (Certified Safe, Fair and Robust Artificial Intelligence). The work has received funding from the Swiss State Secretariat for Education, Research and Innovation (SERI), contract no. MB22.00088.


\bibliography{references}
\bibliographystyle{plainnat}

\clearpage
\appendix

\section{Extended Background on Formal Languages}
\label{app:extended-background}
\label{app:background-languages}

\emph{Formal languages} allow to unambiguously specify valid or invalid strings, usually for ensuring machine-readability, i.e., in the case of JSON schemas, or when specifying the syntactic rules of programming languages. Formal languages are, in their most general form, defined as a set of strings over an alphabet $\Sigma$. For instance, over the alphabet $\Sigma = \{a, b\}$, one can define the formal language $\{ \varepsilon, b, aa, bb, aabb, aaaabbb, \ldots \}$ of strings consisting of any number of a's followed by b's. In this section, we provide a short explanation of two key classes of formal languages: regular and context-free languages.

\subsection{Regular Languages}
\begin{figure}[t]
    \centering
        \begin{subfigure}[t]{0.48\textwidth}
            \centering
            \resizebox{\linewidth}{!}{
                \begin{tikzpicture}[->, >=stealth', auto, semithick, node distance=3cm]
    \node[state, accepting, initial left, initial text=] (q0) {$q_0$};
    \node[state, right of=q0, accepting] (q1) {$q_1$};
    \node[state, right of=q1] (q2) {$q_2$};
  
    \path (q0) edge [loop above] node {a} (q0)
               edge [bend left]  node {b} (q1);
  
    \path (q1) edge [loop above] node {b} (q1);
    \path (q2) edge [loop above] node {\{a, b\}} (q2);

    \path (q1) edge [bend left] node {a} (q2);
               
\end{tikzpicture}
 }
            \caption{A DFA where $q_0$ is the start state, $\{q_0, q_1, q_2\}$ are the states, and $q_0$ and $q_1$ are the accepting states. The arrows represent the transition function $\delta$.}
        \label{fig:dfa}
    \end{subfigure}\hfill%
    \begin{subfigure}[t]{0.48\textwidth}
        \centering
        \vspace{-2.22cm}
        \begin{minipage}{\linewidth}
        \centering
            \begin{align*}
 V &= \{S, B\} \quad \text{(Nonterminals)} \\
                \Sigma &= \{a, b\} \quad \text{(Terminals)} \\
 S &\rightarrow aS \mid bB \mid \varepsilon \\
 B &\rightarrow bB | \varepsilon
            \end{align*}
        \end{minipage}
        \caption{A CFG with start symbol $S$, terminal alphabet $\Sigma = \{a, b\}$, and nonterminals $V = \{S, B\}$. The production rules are the last two lines.}
        \label{fig:cfg}
    \end{subfigure}
    \caption{Two representations of a formal language: a DFA (\cref{fig:dfa}) and a CFG (\cref{fig:cfg}). Both accept strings that start with a's and end with b's.}
    \label{fig:background}
\end{figure}

Regular languages are commonly encountered when describing string patterns with regular expressions. For example, the language of a's followed by b's is described by the regular expression \texttt{a*b*}, where the star denotes zero or more repetitions. A regular language can alternatively be described through a Deterministic Finite Automaton (DFA) that accepts the language \citep{hopcroft1979sensitive}. A DFA is a state machine that processes an input string symbol by symbol, transitioning between states based on a deterministic transition function. Thus, a string gets processed by the DFA by starting in the initial state and following the transitions associated with the current input symbol until the end of the string is reached. A string is accepted if the DFA ends in an accepting state after processing the entire string. Formally, a DFA is defined as a tuple $(Q, \Sigma, \delta, q_0, F)$, where: (1) $Q$ is the finite set of states, (2) $\Sigma$ is the finite alphabet of symbols, (3) $\delta: Q \times \Sigma \rightarrow Q$ is the transition function that maps a state and an input symbol to the next state, (4) $q_0 \in Q$ is the initial state, and (5) $F \subseteq Q$ is the set of accepting states. \cref{fig:dfa} depicts the DFA recognizing the previously introduced language of strings with arbitrarily many a's followed by b's.
Note that in this example, the transition function $\delta$ is defined for every state and symbol combination.
Per convention, omitted transitions implicitly transfer to a state like $q_2$, from which no accepting state can be reached.

In DFAs, the next transition is thus uniquely determined for each combination of state and input symbol. In contrast, nondeterministic finite automata (NFAs) allow multiple transitions for a state and input symbol combination, making it nondeterministic. One often additionally adds the option to transition between states without consuming any input symbols, through so-called $\varepsilon$-transitions. This added flexibility allows for a more concise depiction and simplifies construction, which is why we use them throughout this work. NFAs accept a word if \emph{any} possible transition according to the input symbols leads to an accepting state. Every NFA (including $\varepsilon$-transitions) can be converted to an equivalent DFA using a standard algorithm \citep{hopcroft1979sensitive}. The NFAs constructed for partial LLM outputs in our method are usually converted into a DFA of around the same number of states, even though the worst-case equivalent DFA can have exponentially many states.

\subsection{Context Free Languages}

Context-Free Languages (CFLs) extend regular languages by enabling the expression of recursively nested structures, such as balanced parentheses or properly nested control statements in code. They are described using Context-Free Grammars (CFGs), which consist of production rules that specify how strings in the language can be generated \citep{hopcroft1979sensitive}. For most programming languages, the syntactic rules of the language can be adequately captured by a CFG.

CFGs operate with two types of symbols: terminals, which are the actual characters of the language, and nonterminals, which are used to define the language patterns. A CFG is a formal grammar that consists of a finite set of production rules that describe how strings in the language can be generated.  Formally, a CFG is a tuple $(V, \Sigma, P, S)$, where: (1) $V$ is a finite set of nonterminal symbols, (2) $\Sigma$ is a finite set of terminal symbols (with $V \cap \Sigma = \emptyset$), (3) $P$ is a set of production rules of the form $A \rightarrow \alpha$, with $A \in V$ and $\alpha \in (V \cup \Sigma)^{\ast}$, and (4) $S \in V$ is the start symbol. To generate a string, one starts with $S$ and applies rules from $P$ until the resulting string contains only terminal symbols. This process defines all valid strings in the language. \cref{fig:cfg} shows a CFG that generates strings over $\{a, b\}$ starting with arbitrarily many a's followed by b's, demonstrating that the same language recognized by a DFA can also be described by a CFG. To generate the string \texttt{aabb}, one could apply the following sequence of production rules: $S \rightarrow aS \rightarrow aaS \rightarrow aabB \rightarrow aabbB \rightarrow aabb$.

CFGs are often specified in normal forms, which restrict the grammar to certain types of production rules. The benefit of the resulting language is that it reduces edge cases to handle in productions and simplifies proofs about properties of the language. The most common normal form is the Chomsky normal form, where each production rule is of the form $A \rightarrow BC$ or $A \rightarrow a$, with $A, B, C \in V$ and $a \in \Sigma$. Many other normal forms exist, such as C2F, which is based on the Chomsky normal form but additionally allows so-called unit production rules of the form $A \rightarrow B$ \citep{lange2009cnf}. Languages in Chomsky normal form and C2F can not produce the empty word, as they lack productions that generate the empty string $\varepsilon$ \citep{hopcroft1979sensitive}. We therefore introduce \ctwof, which additionally allows production rules of the form $A \rightarrow \varepsilon$.

\section{Details on Efficient Intersection Language Searches}

In this section, we first detail generic optimizations to reduce the size of context-free grammars, then provide a detailed proof of the correctness of our intersection language construction,  and finally provide some more details on the search algorithm we employ to decide emptiness.

\subsection{Construction of the Intersection Language for CFGs in \ctwof}
\label{app:proof-intersection}

We now provide the full constructive proof that the intersection of a CFL and a regular language is a CFL, since it is rarely written out in the literature. We have further adapted it for grammars in \ctwof{}. It forms the basis of \cref{alg:generating}, the core algorithm of our method.

\begin{lemma}
The intersection language $L \cap R$ between a context-free language $L$ and the regular language $R$ is context-free.
\label{lemma:intersection-cfg}
\end{lemma}
\begin{proof}
 We give a constructive proof by explicitly building a CFG that generates $L \cap R$. We provide the details omitted in the proof given by \citet{gasarch-cfgreg} and extend it to allow grammars in \ctwof.
  
 Let $L_\text{CFL}$ be generated by a CFG $G = (V, \Sigma, P, S)$, and let $L_{R}$ be accepted by a DFA $(Q, \Sigma, \delta, q_0, F)$. We first convert $G$ to \ctwof. Then, we construct a new CFG $G_{\cap}$ whose language is exactly $L_{\cap} = L_\text{CFL} \cap L_{R}$.

The idea is to simulate the CFG $G$ and DFA $(Q, \Sigma, \delta, q_0, F)$ in parallel. Specifically, we define the nonterminals of $G_{\cap}$ to be of the form $\triple{p}{A}{q}$, where $A \in V$ is a nonterminal of $G$, and $p, q \in Q$ are states of the DFA. We then create production rules in such a way that if there exists a sequence of productions such that $\triple{p}{A}{q} \rightarrow \dots \rightarrow w$, then there exists a sequence of productions in $G$ such that $A \rightarrow \dots \rightarrow w$ and $w$ takes the DFA from state $p$ to state $q$. We then add a start symbol $S_{\cap}$ and productions $S_{\cap} \rightarrow \triple{q_0}{S}{f}$ for all $f \in F$ to ensure that $L_{\cap}$ contains exactly the words that can be derived from the start symbol $S$ of $G$ and that also take the DFA from the start state $q_0$ to an accepting state $f \in F$, i.e., all words that are generated by the grammar and all words that are accepted by the DFA. 

 The productions of $G_{\cap}$ are defined as follows (adapting \citep{gasarch-cfgreg}, additional rules in \textcolor{pastelgreen}{green}):

\begin{enumerate}
  \item For each production $A \rightarrow \sigma$, for all $p, q \in Q$ where $\delta(p, \sigma) = q$, we add $\triple{p}{A}{q} \rightarrow \sigma$\label{item:app-construct-ruleSigma}
  \item \textcolor{pastelgreen}{For each production $A \rightarrow \varepsilon$, for all $p \in Q$, add $\triple{p}{A}{p} \rightarrow \varepsilon$}\label{item:app-construct-ruleEpsilon}
  \item For each production $A \rightarrow B C$, and for all $p, q, r \in Q$, we add $\triple{p}{A}{r} \rightarrow \triple{p}{B}{q} \triple{q}{C}{r}$\label{item:app-construct-ruleAB}
  \item \textcolor{pastelgreen}{For each production $A \rightarrow B$, for all $p,q \in Q$, add $\triple{p}{A}{q} \rightarrow \triple{p}{B}{q}$}\label{item:app-construct-ruleA}
\end{enumerate}

The intuition behind the additional rules is that if the automaton is in some state $q$, we can "switch the current symbol" ($A \rightarrow B$) or "produce an empty string" ($A \rightarrow \varepsilon$) without affecting the state. These productions cover the two additional allowed productions in \ctwof grammars, which are not present in CNF grammars.

Finally, we add a new start symbol $S'$ with productions $S' \rightarrow \triple{q_0}{S}{f}$ for all $f \in F$.

We show that the language generated by the constructed CFG $L_{\cap}$ is equivalent to the intersection language of the CFL $L_\text{CFL}$ and regular language $L_R$, i.e., $L_{\cap} = L_\text{CFL} \cap L_R$. To do so, we first need some additional notations:

\begin{itemize}
  \item For any $p, q \in Q$ and $A \in V$, $L(\triple{p}{A}{q})$ denotes the language generated by the nonterminal $\triple{p}{A}{q}$ in the constructed CFG, i.e., the set of all words that can be derived with $\triple{p}{A}{q}$ as the start symbol. Note that $L_{\cap} = \bigcup_{f \in F} L(\triple{q_0}{S}{f})$.
  \item For any $A \in V$, $L(A)$ denotes the language generated by the nonterminal $A$ in the original CFG, i.e., the set of all words that can be derived with $A$ as the start symbol. Note that $L_\text{CFL} = L(S)$.
  \item For any $p, q \in Q$, $L(\dfato{p}{q})$ denotes the language accepted by the DFA with start state $p$ and final state $q$, i.e., the set of all words that can be accepted by the DFA starting in state $p$ and ending in state $q$. Note that $L_R = \bigcup_{f \in F} L(\dfato{q_0}{f})$.
\end{itemize}

We will show that for any $p, q \in Q$ and $A \in V$
\begin{equation*}
 L(\triple{p}{A}{q}) = L(A) \cap L(\dfato{p}{q}).
\end{equation*}
This immediately implies that $L_{\cap} = L_\text{CFL} \cap L_R$, as 
\begin{equation*}
 L_{\cap} = \bigcup_{f \in F} L(\triple{q_0}{S}{f}) = \bigcup_{f \in F} (L(S) \cap L(\dfato{q_0}{f})) = L(S) \cap \bigcup_{f \in F} L(\dfato{q_0}{f}) = L_\text{CFL} \cap L_R.
\end{equation*}

We prove both inclusions separately.

\begin{itemize}
  \item [($\subseteq$)] We show that for any $p, q \in Q$ and $A \in V$, $L(\triple{p}{A}{q}) \subseteq L(A) \cap L(\dfato{p}{q})$. Let the generation path of $w \in L(\triple{p}{A}{q})$ be defined as the sequence of productions that were used to derive $w$ from $\triple{p}{A}{q}$. Denote 
  \begin{equation*}
 L_{n}(\triple{p}{A}{q}) = \{ w \in L(\triple{p}{A}{q}) \mid \text{the generation path of } w \text{ has length  at most } n \}.
  \end{equation*}
 We show the inclusion by induction over the length of the generation path.
  \begin{itemize}
    \item[$\bf{n=1}$.] We show that $L_{1}(\triple{p}{A}{q}) \subseteq L(A) \cap L(\dfato{p}{q})$. Since $w$ is a word, the only possible productions that can be used to derive $w$ from $\triple{p}{A}{q}$ are either rule \labelcref{item:app-construct-ruleSigma} or rule \labelcref{item:app-construct-ruleEpsilon}. 
    
 In the first case, we know $w = \sigma$, $A \rightarrow \sigma$ is a production of the original CFG $G$, and $\delta(p, \sigma) = q$. Hence, $w \in L(A)$ and $w \in L(\dfato{p}{q})$. In the second case, we have $w = \varepsilon$, $A \rightarrow \varepsilon$ is a production of $G$, and $p=q$. Hence, $w \in L(A)$ and $w \in L(\dfato{p}{q})$.
    \item[$\bf{n>1}$.] Suppose that for all $p, q \in Q$ and $A \in V$, $
 L_{n-1}(\triple{p}{A}{q}) \subseteq L(A) \cap L(\dfato{p}{q})$. Let $w \in L_{n}(\triple{p}{A}{q})$ be a word with a generation path of length $n > 1$. Then the first production rule applied to $w$ cannot be rules \labelcref{item:app-construct-ruleEpsilon,item:app-construct-ruleSigma}, as these would yield a generation path of length one. Hence, the first rule applied must be either of the rules \labelcref{item:app-construct-ruleAB,item:app-construct-ruleA}. 
    
 In the former case, we know there exist two words $w_1 \in L_{n-1}(\triple{p}{B}{r}), w_2 \in L_{n-1}(\triple{r}{C}{q})$ such that $w = w_1 \strcat w_2$. By induction, we have $w_1 \in L(B) \cap L(\dfato{p}{r})$ and $w_2 \in L(C) \cap L(\dfato{r}{q})$. Since $A \rightarrow B C$ is a production of $G$, we have $w \in L(A)$ as well. Furthermore, since $w_1$ transitions the DFA from $p$ to $r$ and $w_2$ transitions from $r$ to $q$, we have $w \in L(\dfato{p}{q})$. Hence, $w \in L(A) \cap L(\dfato{p}{q})$. 
    
 In the latter case, we have $w \in L_{n-1}(\triple{p}{B}{q})$ for some nonterminal $B \in V$ such that $A \rightarrow B$ is a production of $G$. By the induction hypothesis, we have $w \in L(B) \cap L(\dfato{p}{q})$. Since $A \rightarrow B$ is a production of $G$, we have $w \in L(A)$ as well. Hence, $w \in L(A) \cap L(\dfato{p}{q})$. 
  \end{itemize}
  \item [($\supseteq$)] We show that $L(\triple{p}{A}{q}) \supseteq L(A) \cap L(\dfato{p}{q})$. Let the generation path of $w$ now be measured with respect to the original CFG, i.e., the sequence of productions that were used to derive $w$ from $A$. Denote 
  \begin{equation*}
 L_{n}(A) = \{ w \in L(A) \mid \text{the generation path of } w \text{ has length at most } n \}.
  \end{equation*}
 We once again show the inclusion by induction over the length of the generation path.
  \begin{itemize}
    \item[$\bf{n=1}$.] We show that for any $p, q \in Q$ and $A \in V$, 
    $L_{1}(A) \cap L(\dfato{p}{q}) \subseteq L(\triple{p}{A}{q})$. Since $w$ is a word, the only possible productions that can be used to derive $w$ from $A$ directly are $A \rightarrow \sigma$ or $A \rightarrow \varepsilon$.  
    
 In the former case, we have $w = \sigma$, and since a DFA only consumes symbols one-by-one, there must be a corresponding state transition, i.e., $\delta(p, \sigma) = q$. Hence, $w \in L(\triple{p}{A}{q})$ by rule \labelcref{item:app-construct-ruleSigma}. 
    
 In the latter case, $w = \varepsilon$, which immediately implies that $p=q$ since a DFA does not contain epsilon transitions. Hence, $w \in L(\triple{p}{A}{q})$ by rule \labelcref{item:app-construct-ruleEpsilon}.
    \item[$\bf{n>1}$.] Suppose that for all $p, q \in Q$ and $A \in V$, $L_{n-1}(A) \cap L(\dfato{p}{q}) \subseteq L(\triple{p}{A}{q})$. Let $w \in L_n(A) \cap L(\dfato{p}{q})$ be a word with a generation path of length $n > 1$. Then the first rule applied cannot be $A \rightarrow \sigma$ or $A \rightarrow \varepsilon$, as these would yield a generation path of length one. Hence, the first rule applied must be either $A \rightarrow B C$ or $A \rightarrow B$ for some nonterminals $B, C \in V$.

 In the former case, we know there exist two words $w_1 \in L_{n-1}(B), w_2 \in L_{n-1}(C)$ such that $w = w_1 \strcat w_2$ and $A \rightarrow B C$ is a production in the original CFG. Since $w \in L(\dfato{p}{q})$, we also know that consuming $w$ transitions the DFA from state $p$ to $q$. We also know that, starting in $q$, after consuming $w_1$, the DFA will arrive at some intermediate state $r$. Clearly therefore $w_1 \in L(\dfato{p}{r})$. Moreover, since $w = w_1 \strcat w_2$ and $w \in L(\dfato{p}{q})$, also $w_2 \in L(\dfato{r}{q})$. By induction, we then have $w_1 \in L_{n-1}(B) \cap L(\dfato{p}{r}) \subseteq L(\triple{p}{B}{r})$ and similarly $w_2 \in L(\triple{r}{C}{q})$. We know that production $\triple{p}{A}{q} \rightarrow \triple{p}{B}{r}\triple{r}{C}{q}$ is in the intersection language, due to rule \labelcref{item:app-construct-ruleAB} quantifying over all states in $Q$. Hence, $w \in L(\triple{p}{A}{q})$.

 In the latter case, we have $w \in L_{n-1}(B)$ and $w \in L(\dfato{p}{q})$. By the induction hypothesis, we have $w \in L(\triple{p}{B}{q})$. Since $A \rightarrow B$ is a production of the original CFG, we have $w \in L(\triple{p}{A}{q})$ as well by rule \labelcref{item:app-construct-ruleA}. 
  \end{itemize}
\end{itemize}
This completes the proof of the lemma.

\end{proof}

\subsection{Grammar Size Optimizations}
\label{app:optimizations}

The size of the grammar used for the intersection generation is of high importance to the overall runtime, as the number of productions in the intersection grammar scales cubically with the number of productions in the original grammar. While the size of the intersection grammar also depends on the size of the intersected DFA, generic and efficient methods to minimize DFAs exist. Meanwhile minimization of CFGs is undecidable \citep{hopcroft1979sensitive}.

We therefore apply several heuristics to reduce the grammar size:
\begin{itemize}
  \item Inlinable terminal elimination: Inline the productions of nonterminals that are only used in a single production. In particular, when $B$ is only used in a single production $A \rightarrow \alpha B \beta$, with $B \rightarrow \gamma$, remove $B$ and its production and inline it into the production of $A$ to create $A \rightarrow \alpha \gamma \beta$.
  \item Shared 2-gram elimination: For the most frequent $BC$ such that there are several rules of the form $A \rightarrow \alpha BC \beta$, (with $\alpha, \beta$ non-empty) introduce $A' \rightarrow BC$ and rewrite $A \rightarrow \alpha A' \beta$. Repeat until no more such $BC$ with more than one occurrence can be found. 
  \item Left factoring: We eliminate shared prefixes using left factoring \citep{alfred2007compilers-left-factoring}. Specifically, if two productions of the same nonterminal $A \rightarrow \alpha \beta$ and $A \rightarrow \alpha \beta'$ share the prefix $\alpha$, we can introduce a new symbol $A'$ and replace the productions to eliminate the duplication, concretely introducing $A \rightarrow \alpha A'$ and $A' \rightarrow \beta$, $A' \rightarrow \beta'$.
\end{itemize}

After applying these heuristics, we convert the resulting CFG to \ctwof using a standard algorithm, consisting of several transformation steps, such as terminal elimination and binarization \citep{lange2009cnf}. In between each step, we detect and eliminate potentially constructed non-generating symbols.

\subsection{Details on the Search Algorithm}\label{app:search-algorithm}

We explain in detail how the search algorithm for generating nonterminals in the intersection language works. The corresponding pseudo-code is presented in \cref{alg:generating} and based on the algorithm presented by \citet{SolvingCFGEmptinessLinear}. We leverage the construction rules of the intersection language to conduct the search on the \emph{implicit} intersection grammar, i.e., we only build the parts of the grammar that we need to explore.
Nonterminals in the intersection language have the form $\triple{p}{A}{q}$ for $p,q \in \Sigma$ and $A \in V$.
All production rules of the form $\triple{p}{A}{q} \rightarrow \varepsilon$ and $\triple{p}{A}{q} \rightarrow \sigma$ are based on the corresponding productions $A \rightarrow \sigma$ (Construction \labelcref{item:app-construct-ruleSigma}) and $A \rightarrow \varepsilon$ (Construction \labelcref{item:app-construct-ruleEpsilon}) in the original grammar. We leverage this insight to perform the initialization of the search, which iterates over all production rules of this format, at the beginning of the algorithm in \crefrange{alg:generating:line-sigma}{alg:generating:line-epsilon-end}. Further, all other productions are of the form $\triple{p}{A}{q} \rightarrow \triple{p}{B}{q}\triple{q}{C}{r}$ and $\triple{p}{A}{q} \rightarrow \triple{p}{B}{q}$, as constructed by Constructions \labelcref{item:app-construct-ruleA,item:app-construct-ruleAB}. Importantly, all rules for all combinations of states $p,q,r$ exist. This allows us to enumerate all such rules for a given symbol $B$ or $C$ on the fly, as done in \crefrange{alg:generating:line-xc}{alg:generating:line-x-end}, without expending unnecessary execution time. For example, in \cref{alg:generating:line-xc}, we iterate over all production rules in which the nonterminal $\triple{y}{X}{z}$ occurs. The two states of the DFAs already fixate two of the three states quantified over in Construction \labelcref{item:app-construct-ruleAB}. Hence, given a production $A \rightarrow XC$ in the original grammar, which uses the nonterminal $X$ and additional nonterminal $C$, we need to iterate over a single additional state variable $q$ to evaluate all corresponding constructed productions $\triple{y}{A}{q} \rightarrow \triple{y}{X}{z}\triple{z}{C}{q}$.

\begin{figure}[p]
\begin{minipage}{\linewidth}
\vspace{-6mm}
\begin{algorithm}[H]
\caption{Deciding intersection emptiness of a CFG and DFA. The CFG is in CNF. $G.\textsc{add}(x)$ inserts $x$ into $G$ and returns true if $x$ was not in $G$ previously.}
\label{alg:generating}
\begin{algorithmic}[1]
\Require CFG $C$, DFA $R = (Q, \Sigma, \delta, q_0, F)$
\Ensure $L(C) \cap L(R) = \emptyset$

\State $G \gets \emptyset$ 
\State \textbf{for} all productions $A \rightarrow \sigma$ \textbf{do} \algline{alg:generating:line-sigma} \Comment{Mark terminal and epsilon productions.}
\State \quad $G \gets G \cup \{\triple{p}{A}{q} \mid \delta(p, \sigma) = q\}$
\State \textbf{for} all productions $A \rightarrow \varepsilon$ \textbf{do}
\State \quad $G \gets G \cup \{\triple{p}{A}{p} \mid p \in Q\}$ \algline{alg:generating:line-epsilon-end}
\State $s \gets G.\textsc{copy}()$ 
\State \textbf{while} $s \neq \emptyset$ \textbf{do} \Comment{Explore all remaining productions}
\State \quad $\triple{y}{X}{z} \gets s.\textsc{pop}()$
\State \quad \textbf{for} all productions $A \rightarrow XC$, all $q \in Q$ \textbf{do} \algline{alg:generating:line-xc}
\State \quad \quad \textbf{if} $\triple{z}{C}{q} \in G$ \textbf{and} $G.\textsc{add}(\triple{y}{A}{q})$ \textbf{then}
\State \quad \quad \quad $s.\textsc{add}(\triple{y}{A}{q})$
\State \quad \textbf{for} all productions $A \rightarrow BX$, all $q \in Q$ \textbf{do}
\State \quad \quad \textbf{if} $\triple{q}{B}{y} \in G$ \textbf{and} $G.\textsc{add}(\triple{q}{A}{z})$ \textbf{then}
\State \quad \quad \quad $s.\textsc{add}(\triple{q}{A}{z})$
\State \quad \textbf{for} all productions $A \rightarrow X$ \textbf{do}
\State \quad \quad \textbf{if} $G.\textsc{add}(\triple{y}{A}{z})$ \textbf{then}
\State \quad \quad \quad $s.\textsc{add}(\triple{y}{A}{z})$\algline{alg:generating:line-x-end}
\State \textbf{return} $G \cap \{\triple{q_0}{S}{f} \mid f \in F\} = \emptyset$\Comment{Whether any start symbol of $L(C) \cap L(R)$ is generating}
\end{algorithmic}
\end{algorithm}
\end{minipage}
\end{figure}

\section{Lexing with LLM Tokens}
\label{app:lexing}

The approach described in \cref{sec:deciding-mri} operates directly on the formal language alphabet $\Sigma$. LLMs produce Unicode text that can be misaligned with $\Sigma$. In this section, we describe in more detail how to handle the resulting discrepancies.

\subsection{Lexemes and LLM Tokens}\label{app:lexing:intro}
\paragraph{Discrepancies between alphabet and LLM tokens} 
For practical purposes, the alphabet $\Sigma$ of the formal language usually consists of \emph{lexemes}. These represent language components abstractly, i.e., for programming languages, they could be identifiers, literals, operators, and other syntactic elements of the language, such as \texttt{if} and \texttt{else}.
Before parsing a Unicode string, it thus first needs to be converted into a string of lexemes. This process is called \emph{lexing}.

The code generation paradigms \mri{} and \dlm{} generate code on a Unicode level and thus require lexing before our method can be applied. In addition to the normal lexing process, our approach needs to handle the partial nature of the LLM outputs, taking into account potential partial lexemes and consequently several possible lexing sequences for the same character-level output. In the remainder of this section, we first explain how to convert the partial LLM output to a set of possible lexeme sequences, and then how to apply the constrained infilling algorithm to these lexeme sequences.

\paragraph{Lexemes and lexing} Each lexeme is associated with a regular language $R$ where $\Sigma_R$ is the set of Unicode characters. For example, the \lexeme{number} lexeme is associated with regular expression \texttt{\textbackslash{}d+}, and the \lexeme{identifier} lexeme with \texttt{[a-zA-Z_]\textbackslash{}w*}. Lexing is the process of converting a Unicode-level string into a sequence of lexemes, i.e., a sequence of strings that match the regular expressions of the lexemes. We call such a sequence of lexemes a \emph{lexeme sequence}. The lexing algorithm extracts these sequences by iteratively matching the maximum match for all lexemes that match a nonempty string at the beginning of the currently remaining output. Whitespace between lexemes is commonly stripped. For example, the character-level string \texttt{"1234 hello12"} would be lexed into the lexeme sequence \texttt{(\lexeme{number},\lexeme{identifier})}.

\subsection{Converting Partial Outputs to Lexemes and DFAs}\label{app:lexing:method}
\paragraph{Lexing partial outputs} For a partial output $\mathbf{x}$ with infilling regions, we extract the represented lexeme sequences for each chunk of continuous text. For instance, the output \texttt{"x = 1234$\hole$hello12"} would be split into the chunks \texttt{"x = 1234" and "hello12"}, which would be lexed into the two lexeme sequences \texttt{(\lexeme{identifier}, \lexeme{=}, \lexeme{number})} and \texttt{(\lexeme{identifier})}. Note that the resulting list of lexeme sequences is a list of words in $\Sigma$ that can be directly used to construct the regular language for the infilling problem as described in \cref{sec:deciding-mri}, for example here forming the infilling problem \ttt{\lexeme{identifier}\lexeme{=}\lexeme{number}\hole{}\lexeme{identifier}}.

\paragraph{Handling lexemes spanning infilling regions}
However, infilling regions complicate the lexing process. Concretely, we need to handle strings that match lexemes partially on the border of infilling regions.

Concretely, strings before an infilling region may end with a string that matches a prefix of some lexeme. For example the output \texttt{"$\hole$123"} could be lexed as \texttt{(\lexeme{number})}. However, the region could be filled with token \texttt{"a"}, resulting in the overall lexing \texttt{(\lexeme{identifier})}. Similarly, strings may match suffixes of lexemes after infilling regions.

Additionally, lexemes may span over an entire infilling region. For example, for the output \texttt{"123$\hole$789"}, a trivial possible lexing is \texttt{(\lexeme{number}, \lexeme{number})}. However, it is also possible to insert a token \texttt{"456"} into the region, such that the lexing of the final character-level text is just a single lexeme sequence \texttt{(\lexeme{number})}. \rev{This also holds for any number of infilled gaps, e.g., \texttt{"123$\hole$4$\hole$5$\hole$6$\hole$789"}. In particular, for any chunk $\alpha \beta$ ending with a prefix $\beta$ of a lexeme \lexeme{a}, consecutive chunks $\gamma_i$ that are prefix of a suffix of \lexeme{a} and a final chunk $\eta \zeta$ starting with a suffix $\eta$ of \lexeme{a}, then a valid corresponding lexeme sequence for the entire chunk sequence could be~$(\text{lex}(\alpha), \lexeme{a}, \text{lex}(\zeta))$.}

\rev{
We also need to ensure the prefixes and suffixes of the lexeme are compatible. For instance, for fixed-width lexemes such as \texttt{\lexeme{while}}, we can not insert a token into the sequence \texttt{"whil$\hole$hile"} to obtain a sequence with only a single lexeme, even though both \texttt{"whil"} and \texttt{"hile"} are true prefixes and suffixes of the lexeme \texttt{while}. We resolve this by determining the intersection of the concrete partially generated output with the lexeme's regular language. Concretely, we construct partial character-level output $\beta \hole{} \gamma_1 \hole{} \dots{} \hole{} \gamma_n \hole{} \eta$ and compute the intersection with the regular language of \lexeme{a} using standard algorithms for the intersection and emptiness of regular languages \citep{hopcroft1979sensitive}. If the language is not empty, \texttt{(\lexeme{a})} is a valid lexing of the entire sequence.} This effectively generalizes a similar solution to the one proposed by \citet{melcer2024constraineddecodingfillinthemiddlecode}, in which they explicitly store the reached states within each prefix and suffix and ensure their reachability.

\begin{figure}
  \begin{minipage}{\linewidth}
      \vspace{-6mm}
  \begin{algorithm}[H]
      \caption{Extracting lexings of a chunk within a partial output.}
      \label{alg:lexing}
      \begin{algorithmic}[1]
      \Require Input string $w$, Terminals $T$
      \Ensure Set $\{(x_i, s_i, p_i)\}_{0 \leq i \leq n}$ of $n$ possible lexeme sequences $x_i$ and optional partial matches to the first ($s_i$) or last ($p_i$) lexeme
      \vspace{2mm}
      \State $S \gets \{(\varepsilon, w, \text{None}, \text{None})\}$\algline{alg:lexing:line-init}\algline{alg:lexing:line1}
      \State \textbf{for} $t \in T$ \textbf{do} \algline{alg:lexing:line-loop-tokens}\algline{alg:lexing:line2}
 \Comment{Determine if the string starts with a suffix of any terminal}
      \State \indnt \textbf{if} $\textsc{suffix}(\textsc{prefix}(t)).\textsc{match}(w) = |w|$ \textbf{then}
 \Comment{If the suffix prefix spans the entire word.} \algline{alg:lexing:line-intermediate-check}\algline{alg:lexing:line3}
      \State \indnt \indnt $S.\textsc{add}(t, \varepsilon, w, w)$\algline{alg:lexing:line4}
      \State \indnt $l \gets \textsc{suffix}(t).\textsc{match}(w)$ \algline{alg:lexing:line-suffix-check}\algline{alg:lexing:line5}
      \State \indnt \textbf{if} $l > 0$ \textbf{then} \Comment{If the suffix matches a non-zero prefix of $w$}\algline{alg:lexing:line6}
      \State \indnt \indnt $S.\textsc{add}(t, w_{>l}, w_{\leq{}l}, \text{None})$\algline{alg:lexing:line-loop-tokens-end}\algline{alg:lexing:line7}
      \State \textbf{while} $S \neq \varnothing$ \textbf{do}\algline{alg:lexing:line-loop}\algline{alg:lexing:line8}

      \State \indnt $(x, w, s, p) \gets S.\textsc{pop}()$ \algline{alg:lexing:line-pop}\algline{alg:lexing:line9}
      \State \indnt \textbf{if} $w = \epsilon$ \textbf{then yield} $(x, s, p)$\algline{alg:lexing:line-yield}\algline{alg:lexing:line10}
      \State \indnt \textbf{for} $t \in T$ \textbf{do}\algline{alg:lexing:line-extend-loop}\algline{alg:lexing:line11}
      \State \indnt \indnt \textbf{if} $\textsc{prefix}(t).\textsc{match}(w) = |w|$ \textbf{then}\algline{alg:lexing:line-prefix-check} \Comment{If the prefix spans the entire remaining word.}\algline{alg:lexing:line12}
      \State \indnt \indnt \indnt $S.\textsc{add}(x \strcat t, \varepsilon, s, w)$\algline{alg:lexing:line13}
      \State \indnt \indnt $l \gets t.\textsc{match}(w)$ \algline{alg:lexing:line-full-check}\algline{alg:lexing:line14}
      \State \indnt \indnt \textbf{if} $l > 0$ \textbf{then} \Comment{If the suffix matches a non-zero prefix of $w$}\algline{alg:lexing:line15}
      \State \indnt \indnt \indnt $S.\textsc{add}(x \strcat t, w_{>l}, s, \text{None})$ \algline{alg:lexing:add-l-in-loop}\algline{alg:lexing:line16}
      \end{algorithmic}
  \end{algorithm}
  \end{minipage}
\end{figure}

\newcommand{\nomarkcomment}[1]{\hfill #1}%
\begin{figure}
  \begin{minipage}{\linewidth}
    \vspace{-6mm}
    \begin{algorithm}[H]
      \caption{\rev{Extracting lexings of an output with infilling regions.}}
      \label{alg:fulllexing}
      \begin{algorithmic}[1]
      \Require Input string $w_1 \hole w_2 \hole \dots \hole w_n$, Terminals $T$
      \Ensure A list of partial lexeme sequences $x$ that match the input string.
      \vspace{2mm}
      \Statex \hspace{-4mm}$\triangleright{}$ Step 1: Collect all possible lexings of partial outputs\vspace{0.5em}
      \State $L \gets \{\}$
      \For{$i$ in $1$ to $n$}
        \State $L.\textsc{add}(\textsc{lex\_partial}(w_i, T))$
        \Comment{Apply \cref{alg:lexing}} \algline{alg:fullexing:for-start}
      \EndFor
      \Statex \vspace{0.5em}\hspace{-4mm}$\triangleright{}$ Step 2: Determine all possible subsequences of lexings\vspace{0.5em}
      \State $C \gets \textsc{cross\_product}(L)$
      \For{$[(x_i, s_i, p_i)]_{i=1}^n$ in $C$} \algline{alg:fullexing:for-start}
      \Statex \vspace{0.5em}\hspace{-4mm}\quad{}$\triangleright{}$ Step 2.1: Determine all possible merges of subsequences\vspace{0.5em}
        \State \textbf{if} not $p_0 = s_n = \text{None}$ \textbf{continue} \Comment{Reject unmatched front or back.} \algline{alg:fulllexing:line-init}
        \State $M \gets \{\}$
        \For{i in $1$ to $n$} \Comment{Check for mergeable subsequences.}
          \State \textbf{if} $s_i = \text{None}$ \textbf{continue} \algline{alg:fulllexing:first-break}
          \For{j in $i+1$ to $n$}
            \State \textbf{if} $p_j = \text{None}$ \textbf{break} \algline{alg:fulllexing:second-break} \Comment{Are the same lexemes matching partially?.}
            \State \textbf{if} $x_i^{(-1)} \neq x_j^{(0)}$ \textbf{break}
            \State \textbf{if} $p_j = w_j$ and $s_i = w_i$  \Comment{Case distinction for all scenarios.} \algline{alg:fulllexing:if-first}
              \State \indnt $o \gets \hole{} s_i \hole{} w_{i+1} \hole{} \dots \hole{} p_j \hole{}$
            \State \textbf{else if} $p_j = w_j$ and $s_i \neq w_i$
              \State \indnt $o \gets \textcolor{white}{\hole} s_i \hole{} w_{i+1} \hole{} \dots \hole{} p_j \hole{}$
            \State \textbf{else if} $p_j \neq w_j$ and $s_i = w_i$
              \State \indnt $o \gets \hole{} s_i \hole{} w_{i+1} \hole{} \dots \hole{} p_j$
            \State \textbf{else}
              \State \indnt $o \gets \textcolor{white}{\hole} s_i \hole{} w_{i+1} \hole{} \dots \hole{} p_j$ \algline{alg:fulllexing:last-if}
            \State \textbf{if} $x_i^{(-1)} \cap o \neq \varnothing$ \algline{alg:fulllexing:non-empty} \Comment{If lexeme can match the partial output}
              \State \indnt $M.\textsc{add}(i, j)$ \nomarkcomment{we can merge from $i$ to $j$.}
            \State \textbf{if} $p_j \neq w_j$ \textbf{break} \Comment{No further merges possible with $s_i$.} \algline{alg:fulllexing:third-break}
            \EndFor
            \EndFor
      \Statex \vspace{0.5em}\hspace{-4mm}\quad{}$\triangleright{}$ Step 2.2: Determine all possible combinations of merges\vspace{0.5em}
      \State $N \gets \{[\ ]\}$ \algline{alg:fulllexing:start-merge} \Comment{No merge is a possible sequence}
      \For{$(i, j)$ in $M$}
        \State $N.\textsc{add}([(i, j)])$
        \For{$[(i'_1, j'_1), \dots, (i'_k, j'_k)]$ in $N$}
          \State \textbf{if} $[i, j] \cap ([i'_1, j'_1] \cup \dots \cup [i'_k, j'_k]) = \emptyset$ \Comment{Non-overlapping merges.}
            \State \indnt $N.\textsc{add}([(i'_1, j'_1), \dots, (i'_k, j'_k), (i, j)])$
        \EndFor
      \EndFor
      \For{$[(i_1, j_1), \dots, (i_k, j_k)]$ in $N$}
      \State \textbf{yield} $x_1 \hole \dots \hole x_{i_1 - 1} \hole x_{i_1} x_{j_1}^{1:} \hole x_{j_1 + 1} \hole \dots \hole x_{i_k - 1} \hole x_{i_k} x_{j_k}^{1:} \hole x_{j_k + 1} \hole \dots \hole x_n$
      \EndFor
      \EndFor \algline{alg:fulllexing:end-merge}
      \end{algorithmic}
    \end{algorithm}
  \end{minipage}
\end{figure}

\paragraph{Lexing algorithm}
We use some helper operations on character-level DFAs for the lexing algorithm. For DFA $D$, we define the function $\textsc{match}$,  which returns $l$, the number of characters in the string that the suffix language automaton matches maximally. For example, $\texttt{\textbackslash{}d+}.\textsc{match}(\texttt{123}) = 3$ and $\texttt{\textbackslash{}d+}.\textsc{match}(\texttt{1hello}) = 1$.
The function $\textsc{prefix}(D)$ returns the true prefix language of $D$, where a \emph{true} prefix is a prefix that can be completed to a full match by appending at least one more character.
For example, \texttt{123} is a true prefix for \texttt{\textbackslash{}d+} but not for \texttt{\textbackslash{}d\textbackslash{}d\textbackslash{}d}. \texttt{12} is a true prefix for both regular expressions.
The function $\textsc{suffix}(D)$ analogously returns the true suffix language of $D$.
Further, we denote as $w_{\leq{}i}$ the string formed by the first $i$ characters of $w$ and $w_{>i}$ the string formed by all characters after the first $i$ characters in $w$.

The lexing algorithm applied to each chunk of continuous text in $\mathbf{x}$ is described in \cref{alg:lexing}. The main mode of operation is to keep track in $S$ of possible lexings and remainders to be processed, starting with the empty lexing and the entire string to be processed in \cref{alg:lexing:line-init}. The method then iterates over all these lexings in \cref{alg:lexing:line-loop}, returns them if the remainder is empty (\cref{alg:lexing:line-yield}) or extends them if a non-empty remainder remains (\cref{alg:lexing:line-extend-loop}). Crucially, \crefrange{alg:lexing:line-loop-tokens}{alg:lexing:line-loop-tokens-end} check whether the text starts with the suffix of any lexeme. Additionally, \cref{alg:lexing:line-prefix-check} checks whether the remainder of the current text is the prefix to some lexeme.

\rev{\paragraph{Applying the constrained infilling algorithm} \cref{alg:fulllexing} describes how to apply the lexing algorithm to a partial output with infilling regions. First, we apply the lexing algorithm to each continuous chunk of text in \cref{alg:fulllexing}, resulting in a list of sets of possible lexeme sequences for each chunk. Next, we take the cross product of these sets to obtain all possible combinations of lexeme sequences for the entire output. Each combination consists of a list of lexeme sequences, along with potential partial matches to the first and last lexemes in each chunk. 
Further steps in the algorithm enable the merging of lexemes that span infilling regions. We first find all possible indices $(i, j)$ for which such merges are possible by checking the compatibility of the partial matches and the regular language of the lexeme, as described previously. We then construct all possible combinations of non-overlapping merges and yield the resulting lexeme-level partial outputs. }

\textbf{Determining non-emptiness} The prior algorithm returns a set of possible lexeme-level partial outputs for the given character-level partial output. If any of the resulting sequences results in a intersection language $L_{\cap}$, then the current character-level partial output is valid, and we can continue generation. If no lexeme sequence results in a non-empty intersection, then we need to reject the current output. Thus, we have to apply the infilling algorithm to each of the word lists derived from the lexing process. In practice, we may derive a large number of lexeme sequences, as different possibilities from text chunks get combined and result in a combinatorial explosion. To further optimize the lexing process, we add two additional optimizations, which we describe in the following paragraphs.

\paragraph{Optimizing subset lexemes} We avoid a combinatorial explosion of possible lexeme sequences by automatically removing lexemes where the accepted language is a subset of the accepted language of another lexeme. For example, in SMILES, the string \texttt{"5"} could be interpreted as \lexeme{digit} or as \lexeme{fifteen}, which is a special lexeme only allowing numbers from 1 to 15.
We resolve this by automatically detecting lexemes $\alpha$ that accept a subset of valid strings of another lexeme $\beta$, and a) remove the subset $\alpha$ from the accepted language of lexeme $\beta$, and b) allow the lexeme $\alpha$ at any position in the grammar where either the subset token or the full token is allowed, in particular we substitute terminal $\beta$ with $\alpha\Hquad|\Hquad\beta$. This optimization reduces the number of extracted sets of possible lexeme sequences for each continuous chunk of text.

We further manually reduce the number of lexemes that overlap and lexemes that are prefixes or suffixes of other lexemes, such as \lexeme{++} and \lexeme{+}, to further optimize performance.

\begin{figure}[t]
    \centering
    \resizebox{.9\linewidth}{!}{
        \begin{tikzpicture}[->, >=stealth', auto, semithick, node distance= .5cm and 1cm]
    \node[state, initial left, initial text=] (q0) {$q_0$};
    \begin{scope}[local bounding box=D1, shift=(q0.west), xshift=25mm]
        \node[state, right=of q0] (d1-0) {$D_{1}^0$};
        \node[state, right=of d1-0] (d1-1) {$D_{1}^1$};
    \end{scope}
    \node[state, right=of d1-1] (q1) {$q_1$};
    \begin{scope}[local bounding box=D2, shift=(q0.west), xshift=25mm]
        \node[state, above right=of q1] (d2-0) {$D_{2}^0$};
        \node[state, right=1.8cm of d2-0] (d2-1) {$D_{2}^1$};
        \node[state, below right=of q1] (d3-0) {$D_{3}^0$};
        \node[state, right=1.8cm of d3-0] (d3-1) {$D_{3}^1$};
    \end{scope}
    \node[state, below right=of d2-1, accepting] (q3) {$q_3$};
  
    \path (q0) edge [bend left]  node {$\varepsilon$} (d1-0);

    \path (d1-0) edge [bend left] node {int} (d1-1);
    \path (d1-1) edge [bend left] node {$\varepsilon$} (q1);
  
    \path (q1) edge [loop above] node {$\Sigma^{*}$} (q1);
    \path (q1) edge [bend left] node {$\varepsilon$} (d2-0);
    \path (q1) edge [bend right] node {$\varepsilon$} (d3-0);

    \path (d2-0) edge [bend left] node {int} (d2-1);
    \path (d2-1) edge [bend left] node {$\varepsilon$} (q3);

    \path (d3-0) edge [bend right] node {identifier} (d3-1);
    \path (d3-1) edge [bend right] node {$\varepsilon$} (q3);

    \node[draw,dashed,fit=(D1), yshift=2mm, inner xsep=2mm, inner ysep=3.5mm] (boxD1) {};
    \node[anchor=south west] at (boxD1.north west) {$D_1$};
    \node[draw,dashed,fit=(D2), yshift=2mm, inner xsep=2mm, inner ysep=3.5mm] (boxD2) {};
    \node[anchor=south west] at (boxD2.north west) {$D_2$};
               
\end{tikzpicture}
 }
    \caption{A union automaton in the second half of the DFA for output \texttt{"123 $\hole$ 789"}, accounts for the possibility to lex the second half as either \texttt{\lexeme{int}} or \texttt{\lexeme{identifier}}. The resulting automaton accepts both valid lexeme sequences \texttt{\lexeme{int}\lexeme{int}} and \texttt{\lexeme{int}\lexeme{identifier}}.}
    \vspace{-3mm}
  \label{fig:constructed-dfa-union}
\end{figure}
\begin{figure}[t]
    \centering
    \resizebox{.9\linewidth}{!}{
        \begin{tikzpicture}[->, >=stealth', auto, semithick, node distance=2cm]
    \node[state, initial left, initial text=] (q0) {$q_0$};
    \begin{scope}[local bounding box=D1, shift=(q0.west), xshift=25mm]
        \node[state, right of=q0] (d1-0) {$D_{1}^0$};
        \node[state, right of=d1-0] (d1-1) {$D_{1}^1$};
    \end{scope}
    \node[state, right of=d1-1] (q1) {$q_1$};
    \begin{scope}[local bounding box=D2, shift=(q0.west), xshift=25mm]
        \node[state, right of=q1] (d2-0) {$D_{2}^0$};
        \node[state, right of=d2-0] (d2-1) {$D_{2}^1$};
    \end{scope}
    \node[state, right of=d2-1, accepting] (q3) {$q_3$};
  
    \path (q0) edge [bend left]  node {$\varepsilon$} (d1-0);

    \path (d1-0) edge [bend left] node {int} (d1-1);
    \path (d1-1) edge [bend left] node {$\varepsilon$} (q1);
  
    \path (q1) edge [loop above] node {$\Sigma^{*}$} (q1);
    \path (q1) edge [bend left] node {$\varepsilon$} (d2-0);

    \path (d2-0) edge [bend left] node {int} (d2-1);
    \path (d2-1) edge [bend left] node {$\varepsilon$} (q3);

    \path (d1-0) edge [bend right, color=blue] node {int} (d2-1);

    \node[draw,dashed,fit=(D1), yshift=2mm, inner xsep=2mm, inner ysep=3.5mm] (boxD1) {};
    \node[anchor=south west] at (boxD1.north west) {$D_1$};
    \node[draw,dashed,fit=(D2), yshift=2mm, inner xsep=2mm, inner ysep=3.5mm] (boxD2) {};
    \node[anchor=south west] at (boxD2.north west) {$D_2$};
               
\end{tikzpicture}
 }
    \vspace{-3mm}
    \caption{A skip connection, highlighted in \textcolor{blue}{blue}, in the DFA for output \texttt{"123 $\hole$ 789"}, accounts for the possibility to lex the input as a single \texttt{\lexeme{int}}. The resulting automaton accepts both valid lexeme sequences for a single int and two ints with intermediate tokens. This construction can be combined with \cref{fig:constructed-dfa-union}.}
    \vspace{-3mm}
  \label{fig:constructed-dfa-skip}
\end{figure}

\paragraph{Combining lexeme sequences to a single NFA} To avoid explicitly enumerating all possible combinations of lexeme sequences of a string, we directly derive a single, larger NFA that accepts all possible combinations of lexeme sequences at the same time. This NFA is structurally similar to the NFAs of each lexeme sequence, but adds alternative paths for mergeable lexemes.

If a text chunk has two or more admissible lexings, we replace the constructed $D_i$ with an NFA that accepts the union of admissible lexings. For example, the output \texttt{"123$\hole$789"} must also admit recognizing the second chunk as the suffix of an identifier. Thus, we obtain the two sequence sets \{(\texttt{\lexeme{int}})\} and \{(\texttt{\lexeme{int}}), (\texttt{\lexeme{identifier}})\}. By generating a single NFA that accepts both sequences (\texttt{\lexeme{int}}) and (\texttt{\lexeme{identifier}}), we can construct a single NFA by applying the concatenation construction to the standard NFA for the first lexeme sequence and the unionized NFA for the second sequence. The resulting NFA is presented in \cref{fig:constructed-dfa-union}.

Another example is depicted in \cref{fig:constructed-dfa-skip}. Here, for the previously shown output \texttt{"123$\hole$789"}, the first chunk ends with a prefix of the lexeme \lexeme{int}, and the second chunk starts with a suffix of the same lexeme. In addition to the standard construction for the possible extracted list (\lexeme{int}, \lexeme{int}), we add an \lexeme{int}-edge from the second-to-last state of $D_0$ to the second state of $D_{1}$, resulting in an alternative path that accepts the list (\lexeme{int}).
These paths are constructed by maintaining a list of suffixes of the previous $D_i$ when constructing $D_{i+1}$, and adding the edge if a suffix matches a prefix of the lexing of $D_{i+1}$.

In contrast to the combinatorial explosion observed when considering all possible combinations of consecutive parsed lexeme sequences, this NFA grows only linearly in the number of sequences. We also observe that the generated corresponding DFA has a similar number of states, confirming that this avoids expensive combination enumeration.

\subsection{\rev{Soundness and Completeness of Lexing Approach}}\label{app:lexing:soundness-completeness}
In this section, we show soundness and completeness of our lexing approach under reasonable assumptions. Throughout, \emph{lex} denotes the lexing function that maps a Unicode string to a lexeme sequence. Each terminal $t$ has a regular language $R_t \subseteq \Sigma^{*}$ over the Unicode alphabet $\Sigma$. For a lexeme $x$, we will denote $R(x)$ as the regular language of the terminal associated with $x$. Further, for a sequence of lexemes $x_1 x_2 \dots x_n$, we denote the regular language described by the sequence as $R(x_1 x_2 \dots x_n) = R(x_1) \circ{} R(x_2) \circ{} \dots \circ{} R(x_n)$, where $\circ$ denotes concatenation of languages. 
We denote substrings of $w_j w_{j+1} \dots{} w_k{}$ of Unicode strings $w = w_1 w_2 \dots{} w_n, k \leq n$ with $w_{j:k}$, where each $w_i$ is a single unicode character.
Finally, we say that a lexeme $x_i$ consumes a substring $w_{j:k}$ if $x_i$ is the part of $x$ that exactly matches the characters $w_{j:k}$.

We assume that the language is compatible with a maximum munch lexer (\cref{assumption:lexer}) and its terminals are not substrings of one another (\cref{assumption:invertibility}).

\begin{assumption}[Maximum Munch]\label{assumption:lexer}
  The lexer matches lexemes by greedily matching the longest possible lexeme at each position. Concretely, let $w = w_1 w_2 \ldots w_n$ be a Unicode string, where each $w_i$ is a single character, and suppose \emph{lex}$(w) = x_1 x_2 \ldots x_m$. Then:
  \begin{enumerate}
    \item There exist indices $k_0, k_1, \ldots, k_m$ with $0 = k_0 < k_1 < \ldots < k_m = n$ such that for all $1 \leq i \leq m$, the substring $w_{k_{i-1}+1:k_i} \in R(x_i)$,
    \item For each $k_i$, there is no $k > k_i$ and terminal $t$ such that the substring $w_{k_{i-1}+1:k} \in R_t$,
    \item \lexfix{For every lexeme sequence $y_1 \dots y_r$ in the target language $L$ and every Unicode string $v \in R(y_1 \dots y_r)$, we have \emph{lex}$(v) = y_1 \dots y_r$,}
    \item \lexfix{If $x_1 x_2 \ldots x_m$ is in the target language $L$, then for every $1 \leq i \leq m$ and every $h$ with $k_{i-1}+1 < h \leq k_i$, there is no $k > k_i$ such that the substring $w_{h:k}$ is a true suffix of a string in $R(x_i)$.}
  \end{enumerate}
  
\end{assumption}

Intuitively, \cref{assumption:lexer} states that the lexer is greedy: at each position it chooses the terminal that extends the match as far as possible before emitting a lexeme and moving on. \lexfix{The third and fourth property express two conditions on this maximum munch lexer to be compatible with the underlying language: they require that this greedy lexing yields the same word after writing out a word of the grammar and lexing it again \citep{chassot2025ziplex,nagy2026chopchop} and that the position at which a lexeme ends must not depend on how the lexeme begins.}

\begin{assumption}[Orthogonal Terminals]\label{assumption:invertibility}
  Let $t_1$ and $t_2$ be two terminals with regular languages $R_1$ and $R_2$, and $t_1 \neq t_2$. Then no suffix of any string in $R_1$ is a prefix of any string in $R_2$.
\end{assumption}

This assumption can be enforced for any given CFG and set of terminals by preprocessing the terminals: whenever two terminals share a prefix or suffix in this way, we factor out the intersection into a new terminal and adjust the grammar accordingly. These assumptions together imply a useful property: 

\begin{corrolary}[Unique Lexing]\label{corrolary:unique-lexing} Let $w$ be a Unicode string, and $x_1 \dots x_m$ be a lexeme sequence such that $w \in R(x_1 \dots x_m)$. Suppose further that, whenever $w_{i:j}$ is consumed by $x_k$ in the sequence, there is no $l > j$ such that $w_{i:l} \in R(x_k)$. Then, $\text{lex}(w) = x_1 \dots x_m$.
\end{corrolary} 

\begin{proof} 
Let $\text{lex}(w) = y_1 \dots y_r$. We show that $x_1$ and $y_1$ are equal and consume the same number of characters from $w$. Recursively applying this argument then shows that $r = m$ and $y_i = x_i$ for all $1 \leq i \leq m$.

Let $x_1$ consume $w_{1:k_1}$ and $y_1$ consume $w_{1:k_2}$. Suppose, for the sake of contradiction, that either $k_1 \neq k_2$ or $x_1 \neq y_1$. However, either $w_{1:k_1}$ is a prefix of $w_{1:k_2}$ or vice versa. By \cref{assumption:invertibility}, this implies that $x_1 = y_1$. Now, because of \cref{assumption:lexer} and the assumptions in the corollary, $k_1 = k_2$ must hold as well.
\end{proof}

We now characterize the outputs of \cref{alg:lexing} and \cref{alg:fulllexing} under these assumptions. We denote the first and last lexeme in a lexeme sequence $x$ with $x^{(1)}$ and $x^{(-1)}$, respectively.

\newcommand{\pref}[1]{Condition \ref{#1}}
\newcommand{\preftwo}[2]{Condition \ref{#1} and \ref{#2}}
\begin{lemma}\label[lemma]{lemma:lexing-correctness}
  Given a Unicode string $w$, \cref{alg:lexing} returns \emph{exactly} all tuples $(x, s, p)$ for which there exist strings $u$ and $v$ such that $u \strcat w \strcat v \in R(x)$ and
  \begin{enumerate}
    \item If $s = p = w$, then $|x| = 1$ and $x^{(1)} = x^{(-1)}$ consumes $u \strcat w \strcat v$.\label{lemma:p1}
    \item If $s = \text{None}$, then $u = \epsilon$. Otherwise, unless $s = p = w$, $x^{(1)}$ consumes $u\strcat s$. \label{lemma:p2}
    \item \lexfix{If $s \neq \text{None}$ and not $s = p = w$, then $s$ is the longest prefix of $w$ that is a true suffix of a string in $R(x^{(1)})$.} \label{lemma:p3}
    \item If $w_i \dots w_j$ is consumed by $x_k$, then there is no $l > j$ such that $w_i \dots w_l \in R(x_k)$. \label{lemma:p4}
    \item If $p = \text{None}$, then $v = \epsilon$. Otherwise, unless $s = p = w$, $x^{(-1)}$ consumes $p\strcat v$. \label{lemma:p5}
  \end{enumerate}
\end{lemma}

\begin{proof}
We prove both directions: (1) every tuple produced by the algorithm satisfies the stated conditions, and (2) every tuple satisfying the conditions is eventually produced by the algorithm.

\paragraph{No other tuples} We show that any element $(x,w',s,p) \in S$ (as maintained in \cref{alg:lexing}) satisfies the following:
\begin{itemize}
  \item $w'$ can be written as $w'=w_{j:|w|}$ for some $1 \le j \le |w|+1$, where we set $w_{|w|+1:|w|} = \epsilon$.
  \item There exist $u$ and $v$ such that $u \strcat w_{1:j-1} \strcat v \in R(x)$ and if $j \neq |w| + 1$, then $v = \epsilon$.
  \item \lexfix{The conditions of the lemma hold with respect to $w_{1:j-1}, u$ and $v$, where the maximality in \pref{lemma:p3} and \pref{lemma:p4} is with respect to all of $w$.}
\end{itemize}
Since the algorithm yields tuples only when $w'=\epsilon$ (\cref{alg:lexing:line-yield}), it follows that all produced tuples satisfy the lemma's conditions.

The initial elements inserted in \cref{alg:lexing:line-init} and \cref{alg:lexing:line4} satisfy these properties trivially. Likewise, the element inserted in \cref{alg:lexing:line7} satisfies the conditions: by the check in \cref{alg:lexing:line6}, some non-empty prefix of $w$ is a suffix of some $a \in R_t$. Choosing $u$ such that $u \strcat w_{\le l} = a$ and setting $v=\epsilon$, one can check directly that all required conditions hold\lexfix{; in particular, \pref{lemma:p3} holds since $l$ is the longest such match}.

Assume inductively that all elements already in $S$ satisfy the conditions. We show that any new element added in the loop \crefrange{alg:lexing:line8}{alg:lexing:line16} also satisfies the conditions. When \cref{alg:lexing:line9} pops $(x,w',s,p)$ from $S$, we know by the induction hypothesis that there exist $u$ and $v$ such that $u \strcat w_{1:|w| - |w'|} \strcat v \in R(x)$. The check in \cref{alg:lexing:line10} ensures $w' \neq \epsilon$, hence $v=\epsilon$ and not both $s=p=w$. 

If \cref{alg:lexing:line12} holds, then $w'$ is a prefix of some $a \in R_t$. In \cref{alg:lexing:line13} we add $(x \strcat t, \epsilon, s, w')$. Let $a = w' \strcat v$, then $u \strcat w \strcat v \in R(x \strcat t)$. All required conditions now hold:
\begin{itemize}
  \item By construction, $u \strcat w \strcat v \in R(x \strcat t)$.
  \item $j = |w|+1$ since $w'=\epsilon$.
  \item \pref{lemma:p1} does not apply.
  \item \pref{lemma:p2} \lexfix{and \pref{lemma:p3} hold} by the inductive hypothesis\lexfix{, since $s$ is unchanged}.
  \item Since $x$ satisfies \pref{lemma:p4}, so does $x \strcat t$.
  \item \pref{lemma:p5} holds since $t$ consumes $w' \strcat v$.
\end{itemize}

Next, consider the element added in \cref{alg:lexing:line16}. By \cref{alg:lexing:line14}, $w'_{\leq l}$ is the largest prefix of $w'$ that matches $t$. Therefore, the element $(x \strcat t, w'_{>l}, s, \text{None})$ added in \cref{alg:lexing:line16} satisfies all conditions:
\begin{itemize}
  \item By construction, $u \strcat w_{1:|w| - |w'|} \strcat w'_{\leq l} \in R(x \strcat t)$.
  \item We may take $j$ such that $w'_{>l} = w_{j:|w|}$.
  \item \pref{lemma:p1} does not apply.
  \item \pref{lemma:p2} \lexfix{and \pref{lemma:p3} hold} by the inductive hypothesis\lexfix{, since $s$ is unchanged}.
  \item Since $x$ satisfies \pref{lemma:p4}, so does $x \strcat t$.
  \item \pref{lemma:p5} holds since $p = \text{None}$ and $v = \epsilon$.
\end{itemize}
Thus, all added elements satisfy the conditions, completing this direction of the proof.

\paragraph{All tuples}
Now let $(x,s,p)$ satisfy the lemma's conditions. If $s = p = w$, then \lexfix{by \pref{lemma:p1},} $|x|=1$ and $x^{(1)}=x^{(-1)}$ consumes $u \strcat w \strcat v$. Thus, $w$ lies in the prefix language of the suffix language of $x$, and \cref{alg:lexing:line-intermediate-check} inserts $(x,\epsilon,w,w)$ into $S$. It is therefore yielded in \cref{alg:lexing:line10}.

Further, if $s \neq \epsilon$, then by \pref{lemma:p2}, $x^{(1)}$ consumes $u \strcat s$\lexfix{, and by \pref{lemma:p3}, $s$ is the longest such match}. Hence when $t = x^{(1)}$ in \cref{alg:lexing:line2}, \cref{alg:lexing:line-suffix-check} inserts $(x^{(1)}, w_{> |s|}, s, \text{None})$ into $S$.

Inductively, suppose $(x^{(1)} x^{(2)} \dots x^{(i-1)}, w', s, \text{None})$ has been added to $S$, where $w'$ is the remaining suffix after consuming the prefix handled by $x^{(1)} \dots x^{(i-1)}$ and $1 \leq i < |w|$. When $t=x^{(i)}$ in \cref{alg:lexing:line11}, the value of $l$ is precisely the length of the prefix of $w'$ consumed by $x^{(i)}$ \lexfix{by \pref{lemma:p4}}. Therefore \cref{alg:lexing:line-full-check} adds $(x^{(1)} x^{(2)} \dots x^{(i)}, w'_{>l}, s, \text{None})$ as required.

Now, if $p = \text{None}$, a similar argument shows that $(x, \epsilon, s, \text{None})$ is added to $S$, knowing that $(x^{(1)} x^{(2)} \dots x^{(-2)}, w', s, \text{None})$ was added previously, since $x^{(-1)}$ fully consumes the remaining suffix $w'$ by \lexfix{\pref{lemma:p5}}.

Finally, we know that $(x^{(1)} x^{(2)} \dots x^{(-2)}, p, s, \text{None})$ gets added to $S$, since the lexemes $x^{(1)}, \dots, x^{(-2)}$ consume all of $w$ except $p$. Because $p$ lies in the prefix language of $x^{(-1)}$ \lexfix{by \pref{lemma:p5}}, condition \cref{alg:lexing:line12} holds when $t = x^{(-1)}$.
Thus, \cref{alg:lexing:line-prefix-check} inserts $(x,\epsilon,s,p)$ into $S$.

This element is eventually popped and yielded in \cref{alg:lexing:line-yield}, completing the proof.
\end{proof}

\begin{lemma}\label[lemma]{lemma:full-lexing-correctness}
\lexfix{Let $w_1 \hole \dots \hole w_n$ be a Unicode string with holes.}
\begin{enumerate}
  \item \lexfix{Let $x_1 \hole x_2 \hole \dots \hole x_m$ be a lexeme sequence returned by \cref{alg:fulllexing}, and let $y_1, y_2, \dots, y_{m-1}$ be lexeme sequences such that $x_1 y_1 x_2 y_2 \dots y_{m-1} x_m \in L$. Then there exists a $w' \in R(w_1 \hole \dots \hole w_n) \cap R(x_1 y_1 x_2 y_2 \dots y_{m-1} x_m)$ such that $\text{lex}(w') = x_1 y_1 x_2 y_2 \dots y_{m-1} x_m$.} \label{lemma:full:sound}
  \item \lexfix{Let $w' \in R(w_1 \hole \dots \hole w_n)$, and let $x_1, x_2, \dots, x_m$ and $y_1, y_2, \dots, y_{m-1}$ be lexeme sequences such that $\text{lex}(w') = x_1 y_1 x_2 y_2 \dots y_{m-1} x_m \in L$ and each full string $w_i$ is consumed by a single lexeme sequence $x_j$ for some $1 \leq j \leq m$. Then \cref{alg:fulllexing} returns $x_1 \hole x_2 \hole \dots \hole x_m$.} \label{lemma:full:complete}
\end{enumerate}
\end{lemma}
\begin{proof}
  \lexfix{We show the two parts separately.}

  \paragraph{\lexfix{Returned sequences can be completed (\partref{lemma:full:sound})}}
  First, let $x_1 \hole x_2 \hole \dots \hole x_m$ be any lexeme sequence returned by the algorithm. By construction, for each chunk $w_i$, there exists a tuple $(x_i', s_i, p_i)$ returned by \cref{alg:lexing} such that $x_i'$ is the lexeme sequence for chunk $w_i$. By Lemma \ref{lemma:lexing-correctness}, there exist strings $u_i$ and $v_i$ such that $u_i w_i v_i \in R(x_i')$. Because of the yield (\cref{alg:fulllexing:end-merge}), we know that $x_i$ is either equal to $x_j'$ for some $j$, or is formed by merging $x_j'$ and $x_k'$ for some $j < k$, where the latter can only happen if the intersection as given in \crefrange{alg:fulllexing:if-first}{alg:fulllexing:last-if} is non-empty. Looking closely at this intersection, it can be easily determined that it implies the existence of strings $s_{j}, \dots s_{k-1}$ such that $u_j w_j s_j w_{j+1} s_{j+1} \dots s_{k-1} w_k v_k \in R(x_i)$. We set $w_i' = u_j w_j v_j$ for chunks that were not merged and to $w_i' = u_j w_j s_j w_{j+1} s_{j+1} \dots s_{k-1} w_k v_k$ for merged chunks. \cref{alg:fulllexing:line-init} implies that $u_1 = v_n = \epsilon$, implying that the concatenation $w_1' \hole w_2' \hole \dots \hole w_m' \in R(w_1 \hole \dots \hole w_n)$. \lexfix{Now, let $y_1, y_2, \dots, y_{m-1}$ be lexeme sequences such that $x_1 y_1 x_2 y_2 \dots y_{m-1} x_m \in L$, and choose arbitrary strings $b_i \in R(y_i)$. Since $w_i'$ is exactly consumed by $x_i$, the string $w' = w_1' b_1 w_2' b_2 \dots b_{m-1} w_m'$ lies in $R(w_1 \hole \dots \hole w_n) \cap R(x_1 y_1 x_2 y_2 \dots y_{m-1} x_m)$. By the third condition of \cref{assumption:lexer}, we obtain $\text{lex}(w') = x_1 y_1 x_2 y_2 \dots y_{m-1} x_m$.}

  \paragraph{\lexfix{Lexings of completions are returned (\partref{lemma:full:complete})}}
  \lexfix{Now, let $w'$ and the lexeme sequences $x_1, \dots, x_m$ and $y_1, \dots, y_{m-1}$ be as in \partref{lemma:full:complete}.} We show that the algorithm will eventually return $x_1 \hole x_2 \hole \dots \hole x_m$. Once again, we first define $w_i'$ as $w_j \hole \dots \hole w_k$ if $x_i$ is the lexeme sequence that consumes chunks $w_j, w_{j+1}, \dots, w_k$. Now, the characterization presented in \cref{lemma:lexing-correctness}\lexfix{, where \pref{lemma:p3} holds by the fourth condition of \cref{assumption:lexer},} and the simple observation that \cref{alg:fulllexing} uses a cross-product (\cref{alg:fullexing:for-start}) and returns lexeme sequences of all possible mergings (\crefrange{alg:fulllexing:start-merge}{alg:fulllexing:end-merge}) directly implies that $x_1 \hole x_2 \hole \dots \hole x_m$ will eventually be returned by the algorithm.
\end{proof}

This characterization allows us to prove soundness and completeness of our lexing approach. Both of these proofs are one-liners that directly follow from \cref{lemma:full-lexing-correctness}\lexfix{: soundness follows from its first part and completeness from its second part}. Importantly, completeness implies that if a proposed token by an LLM leads to an updated output that can be lexed to a valid lexeme sequence, we will accept the token. Further, soundness implies that if we accept a proposed token by the LLM, there exists a completion of the current output that can be lexed to a valid lexeme sequence.

\begin{theorem}[Soundness]
  Given a partial Unicode string $\mathbf{w} = w_1 \hole w_2 \hole \dots \hole w_n$ with infilling regions, if \cref{alg:fulllexing} returns a lexeme sequence that can be completed to a valid word in the language, then there exists a completion $\mathbf{w'}$ of $\mathbf{w}$ such that $\text{lex}(\mathbf{w'})$ is a valid word in the language.
\end{theorem}

\begin{theorem}[Completeness]
  Given a partial Unicode string $\mathbf{w}$ with infilling regions, if there exists a completion $\mathbf{w'}$ of $\mathbf{w}$ such that $\text{lex}(\mathbf{w'})$ is a valid word in the language, then \cref{alg:fulllexing} will return a lexeme sequence that can be completed to a valid word in the language.
\end{theorem}

\section{Experimental Details, Ablations and Case Study}
\label{app:experiments}

In this section, we provide additional details about the implementation, hyperparameters, datasets, runtime overhead, an ablation on the number of diffusion steps, and a case study.

\subsection{Implementation}
\label{app:experiments-impl}

\paragraph{Overview}
Our implementation is written in around 7000 lines of Python and 5500 lines of Rust.
The main logic, concerning LLM sampling and CFG and DFA construction, is written in Python, with the more computationally expensive formal language operations, such as \cref{alg:generating,alg:lexing}, implemented in Rust, compiled as Python bindings. Several low-level formal language operation implementations are inspired by the educational Python implementations by \citet{pyformlang}.

\paragraph{Grammars}
Our C++ grammar covers a comprehensive but not complete subset of C++, with all features used in the canonical solutions of the test set implemented, but advanced features like template functions and user-defined classes are not supported. Moreover, we disallow the insertion of multi-line comments inside function bodies, as this allows the model to generate arbitrary and broken code that is syntactically valid as long as it is finally wrapped in the multi-line comment delimiters. We further restrict models in the \mri{} setting to not generate additional function signatures and bodies to prevent the generation of additional main functions or test cases. 

We preprocess all model outputs by marking word boundaries with special $\langle$ and $\rangle$ tokens that do not appear in the original text and are never generated by the model\footnote{In particular, we use the bytes \texttt{\textbackslash{}x02} and \ttt{\textbackslash{}x03}}. For example, the string \texttt{int main()} is converted to \texttt{$\langle{}$int$\rangle{}$ $\langle{}$main$\rangle{}()$}. This enables us to check for such word boundaries inside the grammar, i.e., being able to distinguish whether white-space was present between symbols even after it is stripped in the lexing process.

The JSON schema grammars are obtained dynamically based on the JSON Schema for each task. We recursively build up the grammar based on the provided specification. \lexfix{The resulting grammars slightly over-approximate the schemas: in objects that allow additional properties, a declared key may occur again as an additional property with an arbitrary value, properties declared in \texttt{oneOf} branches are permitted even if \texttt{additionalProperties} is false, conditional subschemas (\texttt{if}/\texttt{then}/\texttt{else}) and bounds on non-integer numbers (\texttt{minimum}, \texttt{maximum}) are not enforced, and strings may contain invalid escape sequences and unescaped control characters.}
For SMILES, we implement the specification described by \citet{Apodaca2020-smiles}, which is a more precise and efficient variant of prior specifications \citep{smiles,OpenSMILES2025}.

\lexfix{The C++ grammar does not satisfy the fourth condition of \cref{assumption:lexer} for strings and character literals. 
The reason is that preceding backslashes change whether a quote is at the end or inside the literal, and whether this backslash is inside the literal or not depends on where it begins. We provide an example affected program in \cref{fig:cpp-escape-caveat}. Leaving the infilling region empty completes the program validly, but the partial output is rejected: for the string literal that begins before the region, \cref{alg:lexing} only considers the longest possible end \texttt{", "}, which assumes that the first quote is escaped. We argue that these cases are rare enough to not affect the results significantly. A simple fix is considering all suffixes rather than only the longest possible one.}

\begin{figure}[t]
  \centering
  \begin{minipage}{0.75\linewidth}
\begin{lstlisting}[language=cpp]
#include<vector>
using namespace std;
int f(){ vector<string> v = {"a%*$\hole$*)", "b"}; return 0; }
\end{lstlisting}
  \end{minipage}
  \caption{\lexfix{A partial C++ output with a single infilling region ($\hole$) directly before the closing quote of \texttt{"a"}. Leaving the region empty yields a valid program, but the partial output is rejected.}}
  \label{fig:cpp-escape-caveat}
\end{figure}

\rev{
\paragraph{Grammar Prompting}
For the grammar prompting ablation, we append the grammar in a human readable formatting into the prompt, i.e., without normalization and using descriptive symbol names. For \json{}, the ad-hoc generated terminal names are replaced by a string representation of the respective regular expression, symbol names are S$1$ to S$n$. An example for \smiles{} is shown in \cref{prompt:grammar-prompting}
}

\subsection{Models and Hyperparameters}
\label{app:experiments-models}

All methods were run four times, with seeds $0$ to $4$, and we report the averaged results in all tables. We report the maximum among \vanilla{}, \constrained{}, and \constrplus{} decoding with \textbf{boldface}. We \underline{underline} all results where the confidence interval of the improvement over the given method is not positive at $95\%$.
We limit the amount of generated tokens to 256\lexfix{. In \mri{}, we time out if the generation does not complete after 300 seconds with constraints and after 60 seconds without constraints. Without constraints, no model reaches this limit except \deepseekcoderthirtyb{}, which reaches it in $5.3\%$ and $21.0\%$ of the generations for \mritwo{} and \mrithree{}, respectively. These generations exceed roughly 265 tokens and are rarely correct for the other models; based on their results, we estimate that a longer timeout would increase its unconstrained syntactic and functional correctness by at most $0.8\%$ and $0.2\%$, respectively, not changing any of our findings. In \dlm{}, we time out all generations after 60 seconds.}
We run model inference on NVIDIA RTX A6000 GPUs.

\paragraph{Sampling algorithms and temperature} All \mri{} models were sampled with temperature 1 and greedy decoding.
The diffusion models are sampled with a temperature of 0.2.
To pick a token from the diffusion models distribution, we use the \texttt{entropy} algorithm for the \dream{} based models, \dream{}, \dreamcoder{}, and \diffucoder{}, and \texttt{low confidence} for the \llada{} model, as recommended by the model developers.

\paragraph{Diffusion steps}
Diffusion language models can be run with a varying number of diffusion steps, determining how many tokens are sampled from a single model inference \citep{dream2025,nie2025llada}. Lower numbers of steps imply more tokens being sampled from each inferred distribution, which in turn is updated less frequently. One of the key benefits of diffusion language models is to exploit this ability, resulting in overall faster decoding. At the same time, higher numbers of steps are usually associated with increased accuracy on the requested task, as the model can adapt its distribution more frequently to newly inserted tokens. 
When not explicitly stated otherwise, the diffusion models are run with $32$ diffusion steps.
Our choice of step size 32 represents a trade-off between speed and accuracy.

In each diffusion step, model inference is run once on the current state of the partially filled context window. Afterwards, $\frac{n}{k}$ tokens are sampled from the distribution according to the respective algorithms (\texttt{low confidence} or \texttt{entropy}) and replace mask tokens in the context window. While unconstrained decoding allows sampling all $\frac{n}{k}$ tokens in parallel, during constrained decoding, we iteratively sample single token-index pairs from this distribution, with rejections leading to masking out the rejected token-index pair and resampling. When a token is accepted, we remove the token's index from the distribution. After $\frac{n}{k}$ tokens have been accepted, we run model inference again.

\subsection{Datasets}
\label{app:experiments-data}

\paragraph{C++}
We leverage the C++ translation of HumanEval in the HumanEval-X dataset \citep{zheng2023codegeex-humaneval-x}. It contains 164 instances of simple programming problems and canonical solutions written by humans. For the \mri{} tasks, we remove between 1 and 3 randomly sized spans of 5 to 100 characters from these canonical solutions, generating one \mri{} task per instance in the original dataset. If we end up with insufficient remaining characters after removing the required number of spans, we resample sizes and positions up to 3 times, aborting if we do not find a valid removal. Additionally, we remove 5 human-written solutions that are not valid according to our implementation of the syntax, i.e., because they contain multi-line comments or additional helper functions. This results in three \mri{} datasets of 159, 156, and 143 samples in \mrione{}, \mritwo{} and \mrithree{} respectively. An example prompt for an instance from the dataset is presented in \cref{prompt:cpp-mri}.
For \dlm{}, we use all 164 tasks of the original dataset and extract the comment before the function as an instruction for the model. An example prompt is shown in \cref{prompt:cpp-dllm}.

We check the functional correctness in both settings by checking whether all test cases in the dataset pass with the model-generated solution.

\paragraph{JSON Schema}
We extend the JSON-Schema dataset by \citet{nousresearch-json-mode-eval}. Concretely, the dataset originally contains a unique schema per task. We clean the schemas by disallowing properties other than specified on the top level and repairing instances that accidentally do not require any fields. We then extend the dataset by sampling \gemini{} for 10 inputs and completions for each schema. We filter these samples in \lexfix{four} ways to ensure high quality. 

First, we filter the resulting extracted outputs for syntactic validity according to the schema and discard invalid generations. Second, we require \gemini{} to be able to solve the task, i.e., the model must generate a valid JSON object that passes the schema validation if it is only given the input and the schema. Third, we perform fuzzy matching to deduplicate the resulting samples. \lexfix{Finally, we validate each reference output against its schema with a JSON Schema validator and discard instances that fail.} This process results in \lexfix{269} instances. The prompts used for generation and verification are shown in \cref{prompt:json-schema-generation} and \cref{prompt:json-schema-verification}. An example prompt for this task is shown in \cref{prompt:json-dllm}.

We evaluate functional correctness on this dataset by checking for exact equality between a normalized JSON dump of the golden solution and the model-generated solution. 

\paragraph{SMILES}
To create a benchmark for SMILES, we query \gemini{} to generate pairs of descriptions of molecules and their SMILES notation. Again, we perform three filtering steps to ensure high quality. First, we verify that the generated molecule is valid using the Rdkit library \citep{greg-landrum-2025-16439048}. Second, we ensure the model can generate the correct SMILES string for the molecule if it is only given the description. Third, we filter out duplicates using fuzzy matching. This results in 167 pairs of descriptions and SMILES strings. Prompts for this generation procedure are shown in \cref{prompt:smiles-generation} and \cref{prompt:smiles-verification}. An example prompt for this task is shown in \cref{prompt:smiles-dllm}.

To check the functional correctness of the model-generated molecule, we parse it using Rdkit and check the equivalence to the molecule generated by \gemini{} in canonical representation.

\subsection{Additional Experiments}
\label{app:experiments-add}

\paragraph{Runtime overhead}
For all experiments in \cref{sec:experiments}, we measure the runtime of our constraining method and unconstrained decoding. We present a detailed comparison in \cref{tab:runtime-fim,tab:runtime-dlm}. We further measure the average number of rejections per sample.

\begin{figure}
  \center
  \includegraphics[width=\linewidth]{figures/jsonschema\_time\_vs\_grammar\_size.pdf}
  \caption{The median runtime overhead of \constrplus{} grouped by size of the grammars in \json{}, with a 2-degree polynomial fitted against the individual points. A clear increase can be seen, although it ranges only between $0$ and $3$ seconds.}
  \label{fig:json-overhead}
\end{figure}

In \mri{} we compare time per token, as constrained decoding often rejects finalizing the current output, thus making completions longer and finalization times incomparable. The median runtime overhead of constrained decoding is $125\%$, where the overhead on the small \deepseekcodertwob{} is higher ($320\%$) than on the 7B model ($100\%$) and \deepseekcoderthirtyb{} ($20\%$).
This is both due to the lower inference time of smaller models, and due to smaller models making more mistakes, with the average number of rejections increasing from $8.8$ per instance on 33B, over $9.7$ for 7B to $10.5$ in 1.3B.
Moreover, more infilling regions are more difficult, leading to more rejections, growing from $4.7$ on \mrione{} to $14.1$ in \mrithree{}. This increases the overhead from $67\%$ to $205\%$ respectively.

For \dlm{}, we compare the total runtime to finish the diffusion decoding process. The average completion overhead is only $30\%$, but varies strongly between domains. We observe both speed-ups of up to $19\%$, for \llada{} on \humaneval, where many decodings are preemptively aborted, and slowdowns of up to $190\%$, for \dreamcoder{} on the same dataset.

\rev{We further analyze the runtime overhead based on grammar size and infilling regions. 
We first analyze the runtime overhead per size of the normalized grammar, as the normalized grammar is the basis for our intersection algorithm. The normalized grammar sizes of \humaneval{} and \smiles{} are $167$ productions and $46$ productions respectively. As shown in \cref{tab:runtime-dlm}, this corresponds to a median runtime overhead of \SI{2.5}{s} and \SI{0.0}{s} respectively, indicating a strong correlation to the grammar size. For a closer analysis, we plot the median runtime overhead for each of the \json{} tasks in \cref{fig:json-overhead}, where each task has an individual grammar, and  regress a 2-degree polynomial against the data. An increase in runtime, according to the asymptotic complexity described in \cref{sec:deciding-mri} is visible, but the constant appears very small. For an analysis of the impact of infilling regions, we refer to our results of runtime overhead for increasing number of infilling regions in \mri{} \cref{tab:runtime-fim}, where more infilling regions increase runtime from \SI{3.2}{s} to \SI{8.4}{s}.}

\begin{table}
  \centering
    \renewcommand{\arraystretch}{1.0}
    \setlength{\tabcolsep}{3pt}
    \caption{Median overhead per token for different infilling settings in milliseconds and percent increase over unconstrained generation. Larger models with higher inference time experience a lower slowdown due to constraining. More infilling regions also increase constraining overhead.}
    \label{tab:runtime-fim}
    \vspace{0mm}
    \begin{tabular}{@{}l r@{}l p{0.5mm}r@{}l p{0.5mm} r@{}l@{}}
        \toprule
\#Regions        &  \multicolumn{2}{c}{\mrione}  &&  \multicolumn{2}{c}{\mritwo} &&   \multicolumn{2}{c}{\mrithree}       \\
\midrule
\codegemma        &       3.1 & $_{\uparrow 47\%}$  &        & 4.1 & $_{\uparrow 63\%}$  &          &    6.4 & $_{\uparrow 99\%}$  \\
 \starcodertwo     &       3.3 & $_{\uparrow 59\%}$  &        & 5.5 & $_{\uparrow 98\%}$  &          &    9.7 & $_{\uparrow 190\%}$ \\
 \deepseekctwob    &       3.6 & $_{\uparrow 158\%}$ &        & 5.8 & $_{\uparrow 245\%}$ &          &   11.8 & $_{\uparrow 557\%}$ \\
 \deepseekcsevenb  &       3.0 & $_{\uparrow 58\%}$  &        & 4.6 & $_{\uparrow 90\%}$  &          &    7.7 & $_{\uparrow 153\%}$ \\
 \deepseekcthirtyb &       3.1 & $_{\uparrow 13\%}$  &        & 4.3 & $_{\uparrow 19\%}$  &          &    6.5 & $_{\uparrow 28\%}$  \\
\bottomrule
\end{tabular}
\end{table}

\begin{table}[t]
  \centering
    \renewcommand{\arraystretch}{1.0}
    \setlength{\tabcolsep}{3pt}
    \caption{Median time difference per completion for different diffusion models in seconds, and the overhead over the original completion in percent. When the completion aborts pre-emptively, as no valid completion is sampled from the model, speed-ups are possible.}
    \label{tab:runtime-dlm}
    \vspace{0mm}
    \begin{tabular}{@{}l r@{}l p{0.5mm}r@{}l p{0.5mm} r@{}l@{}}
        \toprule
 Model        &   \multicolumn{2}{c}{\humaneval} &&   \multicolumn{2}{c}{\json}             &&   \multicolumn{2}{c}{\smiles}             \\
\midrule
 \dream    &                             1.1 & $_{\uparrow 36\%}$             &                              & 0.4 & $_{\uparrow 20\%}$     &                       &                 0.0 & $_{\uparrow 0\%}$ \\
 \dreamc   &                             7.8 & $_{\uparrow 190\%}$            &                              & 0.1 & $_{\uparrow 5\%}$      &                       &                 0.0 & $_{\uparrow 1\%}$ \\
 \llada    &                            -1.0 & $_{\downarrow 19\%}$            &                              & \lexfix{0.6} & $_{\uparrow 9\%}$      &                       &                 0.0 & $_{\uparrow 1\%}$ \\
 \diffuc   &                             2.2 & $_{\uparrow 74\%}$             &                              & 0.1 & $_{\uparrow 6\%}$      &                       &                 0.0 & $_{\uparrow 2\%}$ \\
\bottomrule
\end{tabular}
\end{table}

\paragraph{Ablation on diffusion steps}
We evaluate our method on common diffusion step numbers, from $16$ to $256$, where the lowest setting $16$ implies that a single inference step inserts $\frac{256}{16} = 8$ tokens at once, while the highest setting $256$ implies that every inference step inserts only a single token.

We present the results of this ablation on \dream{} in \cref{tab:ablation} and demonstrate that our method consistently improves syntactic correctness in all settings by on average $14\%$. Functional correctness on \json{} also significantly increases by $1.2\%$, while the increase in \humaneval{} is $0.7\%$ and $0.5\%$ in \smiles{}. Moreover, the runtime overhead, shown in \cref{tab:ablation-time}, decreases with the number of diffusion steps, from \lexfix{$14\% - 107\%$} down to $9\%$ or even a speed up of $3\%$.

\begin{table}
  \centering
    \renewcommand{\arraystretch}{1.0}
    \setlength{\tabcolsep}{3pt}
    \caption{Percent syntactically and functionally correct generations for \dream{} based on varying number of diffusion steps. Our method consistently increases syntactic correctness in all settings, even when model accuracy increases with step sizes.}
    \label{tab:ablation}
    \begin{tabular}{@{}lc @{}p{5mm}@{} ccc @{}p{5mm}@{} ccc @{}p{5mm}@{} ccc @{}}
        \toprule
       &&&   \multicolumn{3}{c}{\humaneval} &&   \multicolumn{3}{c}{\json}             &&  \multicolumn{3}{c}{\smiles}             \\
\cmidrule{4-6} \cmidrule{8-10} \cmidrule{12-14}
 & \#Steps        &&   \vanilla{} &   \constrained{} & \constrplus{} &&\vanilla{} &   \constrained{} & \constrplus{} &&  \vanilla{} &   \constrained{} & \constrplus{} \\
\midrule
\parbox[t]{3mm}{\multirow{5}{*}{\rotatebox[origin=c]{90}{Syntax}}}
&        16 &    &         8.1 &       20.3 & \textbf{99.2}  &    &         7.3 &       \lexfix{24.7} & \textbf{100.0} &    &        41.1 &       80.5 & \textbf{99.7}  \\
&        32 &    &        40.5 &       58.7 & \textbf{99.4}  &    &        \lexfix{22.7} &       \lexfix{45.4} & \textbf{100.0} &    &        67.5 &       93.7 & \textbf{99.4}  \\
&        64 &    &        60.1 &       74.7 & \textbf{99.8}  &    &        \lexfix{67.8} &       \lexfix{73.6} & \textbf{100.0} &    &        79.2 &       94.9 & \textbf{100.0} \\
&       128 &    &        81.1 &       90.7 & \textbf{100.0} &    &        \lexfix{90.1} &       94.0 & \textbf{100.0} &    &        80.1 &       95.8 & \textbf{100.0} \\
&       256 &    &        98.2 &       98.2 & \textbf{100.0} &    &        95.2 &       \lexfix{98.1} & \textbf{100.0} &    &        80.7 &       93.4 & \textbf{100.0} \\
 \midrule
\parbox[t]{3mm}{\multirow{5}{*}{\rotatebox[origin=c]{90}{Functional}}}
&        16 &    & 1.4           & 2.7             & \textbf{4.9}  &    & 1.5           & 2.3           & \textbf{3.4}  &    & \underline{0.6} & \textbf{1.1} & \textbf{1.1} \\
&        32 &    & 6.6           & \underline{8.8} & \textbf{9.5}  &    & 7.4           & \lexfix{11.5}          & \textbf{\lexfix{14.4}} &    & \underline{0.6} & \textbf{1.1} & \textbf{1.1} \\
&        64 &    & 21.0          & 21.8            & \textbf{22.4} &    & \lexfix{42.3}          & \lexfix{42.6}          & \textbf{\lexfix{43.3}} &    & 2.4             & \textbf{3.0} & \textbf{3.0} \\
&       128 &    & \textbf{24.5} & 23.9            & 24.1          &    & \lexfix{50.5}          & \textbf{\lexfix{51.2}} & \textbf{\lexfix{51.2}} &    & 3.1             & \textbf{4.0} & \textbf{4.0} \\
&       256 &    & 34.1          & 34.1            & \textbf{34.8} &    & \textbf{\lexfix{54.6}} & \textbf{\lexfix{54.6}} & \textbf{\lexfix{54.6}} &    & \underline{4.9} & \textbf{5.2} & \textbf{5.2} \\
 \bottomrule
\end{tabular}
\end{table}

\begin{table}
  \centering
    \renewcommand{\arraystretch}{1.0}
    \setlength{\tabcolsep}{3pt}
    \caption{Time difference per completion for different step sizes on \dream{} diffusion, in seconds, and the percentual overhead over the original completion. For larger numbers of diffusion steps, overhead reduces from \lexfix{$14\% - 107\%$} down to $9\%$ or even a speedup of  $1\%$.}
    \label{tab:ablation-time}
    \vspace{0mm}
    \begin{tabular}{@{}c r@{}l p{0.5mm}r@{}l p{0.5mm} r@{}l@{}}
        \toprule
 \#Steps        &   \multicolumn{2}{c}{\humaneval} &&   \multicolumn{2}{c}{\json}             &&   \multicolumn{2}{c}{\smiles}             \\
\midrule
        16 &                             1.7 & $_{\uparrow 107\%}$            &                              &  \lexfix{2.0} & \lexfix{$_{\uparrow 106\%}$}    &                       &                 0.0 & $_{\uparrow 14\%}$ \\
        32 &                             1.1 & $_{\uparrow 36\%}$             &                              &  0.4 & $_{\uparrow 20\%}$     &                       &                 0.0 & $_{-}$  \\
        64 &                             0.6 & $_{\uparrow 18\%}$             &                              &  0.1 & $_{\uparrow 4\%}$      &                       &                -0.2 & $_{\downarrow 3\%}$ \\
       128 &                             0.4 & $_{\uparrow 10\%}$             &                              &  0.2 & $_{\uparrow 4\%}$      &                       &                -0.4 & $_{\downarrow 3\%}$ \\
       256 &                             0.8 & $_{\uparrow 9\%}$              &                              & -0.1 & $_{\downarrow 1\%}$     &                       &                 0.2 & $_{\uparrow 1\%}$  \\

\bottomrule
\end{tabular}
\end{table}

\paragraph{Removing entire lines in \mri{}}
\rev{
The \mri{} setting is based on the tasks proposed by \citet{bavarian2022efficienttraininglanguagemodels-fim}. In this work, two settings are suggested: removing random spans and removing entire lines of code. In \cref{sec:experiments}, we presented the results for \mri{} when removing random spans. Here, we explore the alternative of removing lines, which we call \mril{}. Crucially, lines are often semantically coherent and self-contained, thus posing a different challenge.
Equivalenty to \mri{}, we introduce \mrionel{}, \mritwol{} and \mrithreel{}, where we remove between one and three lines from the code. If lines are adjacent, we merge them into a single span to complete.
}

\rev{
We present the results in \cref{tab:mri-lines}. We observe that this task appears more difficult for the LLMs, contrasting the observations by \citet{bavarian2022efficienttraininglanguagemodels-fim}. Concretely, model performance for the \deepseekcoder{} family drops to near $1\%$. Meanwhile, matching the results observed in \cref{sec:experiments} for random spans, in all but one setting our constrained decoding improves model syntactic and semantic correctness, often significantly.
}

\begin{table}
  \centering
    \renewcommand{\arraystretch}{1.0}
    \setlength{\tabcolsep}{3pt}
    \caption{\rev{When infilling between 1 and 3 missing lines, our method consistently improves syntactic and functional correctness. Shown below the results of \mril{} under standard decoding (\vanilla), constrained decoding (\constrained), and completing partially completed outputs (\constrplus).}}
    \label{tab:mri-lines}
    \begin{tabular}{@{}ll @{}p{7mm}@{} ccc @{}p{7mm}@{} ccc @{}p{7mm}@{} ccc @{}}
        \toprule
       &&&   \multicolumn{3}{c}{\mrionel} &&   \multicolumn{3}{c}{\mritwol}             &&  \multicolumn{3}{c}{\mrithreel}             \\
\cmidrule{4-6} \cmidrule{8-10} \cmidrule{12-14}
 & Model        &&   \vanilla{} &   \constrained{} & \constrplus{} &&\vanilla{} &   \constrained{} & \constrplus{} &&  \vanilla{} &   \constrained{} & \constrplus{} \\
\midrule
\parbox[t]{3mm}{\multirow{5}{*}{\rotatebox[origin=c]{90}{Syntax}}}
& \starcodertwo     &           & \underline{84.0} &       95.4 & \textbf{98.3} &    & \underline{78.5} &       91.3 & \textbf{96.5} &    &        63.5 &       86.8 & \textbf{95.7} \\
& \codegemma        &           & \underline{94.8} &       95.7 & \textbf{98.8} &    & 83.8             &       93.9 & \textbf{97.3} &    &        70.1 &       91.3 & \textbf{96.9} \\
& \deepseekctwob    &           & 62.3             &       69.1 & \textbf{87.2} &    & 30.8             &       44.5 & \textbf{75.8} &    &        23.8 &       39.4 & \textbf{76.7} \\
& \deepseekcsevenb  &           & 64.0             &       74.5 & \textbf{95.7} &    & 29.6             &       52.6 & \textbf{93.3} &    &        24.1 &       49.8 & \textbf{89.6} \\
& \deepseekcthirtyb &           & 65.2             &       71.2 & \textbf{92.8} &    & 30.0             &       45.0 & \textbf{87.3} &    &        22.9 &       39.1 & \textbf{84.4} \\
 \midrule
\parbox[t]{3mm}{\multirow{5}{*}{\rotatebox[origin=c]{90}{Functional}}}
& \starcodertwo     &           & \underline{69.5} & \underline{71.8} & \textbf{72.0} &    & 48.6             & \textbf{49.8}   & \textbf{49.8} &    & 28.5            & \textbf{31.9}    & \textbf{31.9} \\
& \codegemma        &           & \textbf{82.2}    & 79.4             & 79.4          &    & \underline{53.4} & \textbf{55.8}   & \textbf{55.8} &    & 34.5            & \underline{40.5} & \textbf{40.6} \\
& \deepseekctwob    &           & 0.9              & 5.2              & \textbf{8.7}  &    & 1.1              & \underline{2.0} & \textbf{2.6}  &    & 0.5             & \underline{1.5}  & \textbf{1.8}  \\
& \deepseekcsevenb  &           & 0.6              & \underline{8.8}  & \textbf{13.7} &    & 1.2              & \underline{2.7} & \textbf{3.7}  &    & \underline{0.0} & \underline{0.8}  & \textbf{1.1}  \\
& \deepseekcthirtyb &           & 0.6              & \underline{5.9}  & \textbf{10.1} &    & 1.4              & \underline{3.4} & \textbf{4.0}  &    & 0.0             & \underline{1.4}  & \textbf{2.1}  \\
 \bottomrule
\end{tabular}
\end{table}

\subsection{\rev{Comparison to DINGO}}\label{app:experiments:dingo}

In this section we compare the performance of our method to the method DINGO proposed by \citet{suresh2025dingoconstrainedinferencediffusion}. This method is specifically designed for diffusion language models, but can only enforce regular constraints. As such, it is less general than our method, which can enforce context-free constraints. Since the method is not openly accessible, we attempt matching their generation settings and datasets here and compare to the results by them.

\paragraph{Datasets}
\citet{suresh2025dingoconstrainedinferencediffusion} propose two datasets for constrained decoding.

The first dataset \jsonD{} is similar to our dataset \json{}: The model is provided with a natural language text and asked to extract data matching a provided JSON schema. The underlying dataset contains the $100$ tasks of \citet{nousresearch-json-mode-eval} on which our dataset \json{} is based. As opposed to our implementation, they count as syntactically correct any output that is generally valid JSON, and functionally correct any output that adheres to the JSON-Schema.

The second dataset, \gsmk{} \citep{mirzadeh2025gsmsymbolicunderstandinglimitationsmathematical}, provides the LLM with a symbolic representation of arithmetic tasks, based on the \textsc{gsm8k} dataset proposed by \citet{cobbe2021trainingverifierssolvemath}\footnote{It should be pointed out however, to avoid confusion, that this evaluation utilizes the dataset produced by \citet{mirzadeh2025gsmsymbolicunderstandinglimitationsmathematical} in a very different way than suggested by them.}. The LLM should generate an arithmetic expression that uses basic arithmetic operators and specific variables, and that matches the ground-truth solution. An example is provided in \cref{prompt:gsm8k}. We exclude the tasks whose golden solutions require conditionals (\texttt{x if y else z}) and interpret \texttt{int(x)} and \texttt{frac(y,z)} as \texttt{x} and \texttt{y/z} respectively.
Syntactically correct outputs must have valid arithmetic expressions in the last occurring pair of \texttt{<{}<} and \texttt{>{}>}. Functionally correct outputs provide an equivalent formula as the golden solution, based on an assessment by Z3 \citep{z3}. 

\paragraph{Grammars} Although both tasks require context-free grammars to be accurately solved, \citet{suresh2025dingoconstrainedinferencediffusion} disallow context-free parts of the JSON-schema language and hard-code a maximum nesting for arithmetic expressions to allow solving the tasks in these datasets with terms matched by regular expressions, provided in the Appendix of \citet{suresh2025dingoconstrainedinferencediffusion}. Our implementation for \jsonD{} constructs the same grammars as in our dataset \json{}. For \gsmk{}, we introduce a new grammar that allows the language model to first reason freely (allowing any characters), and constrains only the content of the last pair of \texttt{<{}<} and \texttt{>{}>} to a simple, context-free language of arithmetic expressions with addition, subtraction, multiplication, exponentiation and modulo.

\paragraph{Models and Parameters}
We compare the performance of \llada{} and \dream{} with $64$ diffusion steps, generation length of $128$, $1$ generation block and temperature $0.2$ and sample $4$ times to compute confidence intervals.

\paragraph{Results}
The syntactic and functional results are presented in \cref{tab:dingo-perf}. We observe that both our method \constrplus{} and \dingo{} correctly ensure perfect adherence to JSON-Schemas in the task \jsonD{}. For \gsmk{}, we observe that \dingo{} and \constrplus{} both enforce $100\%$ syntactic correctness, but \dingo{} slightly outperforms \constrplus{} in functional correctness. This can be due to multiple reasons: The alignment of \dingo{} is closer to a deeply grammar-aligned solution as proposed by \citet{park2024grammar} than the grammar-constrained approach that our method uses. Further, \dingo{} hardcodes the maximum nesting of arithmetic expressions, thereby biasing the model towards correct solutions.

Comparing the runtime overhead of both methods, we observe that our method and \dingo{} achieve similar runtime overhead per sample, with both impacting generation by less than $0.3$s. Notably, \dingo{} requires additional preprocessing for each grammar, which takes up to half a minute. In the case of \jsonD{}, this implies additional preprocessing for every sample, as each JSON Schema defines its own grammar. Meanwhile, our method requires no preprocessing.

\begin{table}
  \center
  \caption{\rev{On \jsonD{}, our method (\constrplus) achieves the same performance as \dingo{}. On \gsmk{}, \dingo{} slightly outperformsn \constrplus{}.}}
  \label{tab:dingo-perf}
\begin{tabular}{@{}lllccccccccccc}
\toprule
  &&&   \multicolumn{4}{c}{\jsonD} &&   \multicolumn{4}{c}{\gsmk}           \\
\cmidrule{4-7} \cmidrule{9-12}
 & Model        &&   \vanilla{} & \dingo{} &  \constrained{} & \constrplus{} &&\vanilla{} & \dingo{} &   \constrained{} & \constrplus{} \\
\midrule
\parbox[t]{3mm}{\multirow{2}{*}{\rotatebox[origin=c]{90}{Syn.}}} 
 & \llada    &    &        93.0 &     \textbf{100} &       88.5 & \textbf{100.0} &    &        72.0 &     100 &       87.0 & \textbf{100.0} \\
 & \dream    &    &        89.0 &     \textbf{100} &       91.0 & \textbf{100.0} &    &        65.0 &     100 &       69.0 & \textbf{100.0} \\
\midrule
\parbox[t]{3mm}{\multirow{2}{*}{\rotatebox[origin=c]{90}{Func.}}} 
& \llada    &    &        86.2 &     \textbf{100} &       87.5 & \textbf{100.0} &    & 24.7          &      \textbf{29} & 27.7 & 27.7 \\
& \dream    &    &        85.0 &     \textbf{100} &       90.0 & \textbf{100.0} &    & 25.0 &      \textbf{34} & 25.0 & 25.0 \\
\bottomrule
\end{tabular}
\end{table}

\begin{table}
  \center
  \caption{\rev{Pre-procesing time ($\text{Pre}_{X}$) and time difference per completion of \dingo{} and our method (\constrplus). We observe that our method has a similar runtime overhead as \dingo{} while requiring no preprocessing. Notably, preprocessing is done once per schema, which implies once per task on the \jsonD{} dataset.}}
  \label{tab:dingo-perf}
\begin{tabular}{@{}lp{0.5mm}ccccccccc}
\toprule
&&   \multicolumn{4}{c}{\jsonD} &&   \multicolumn{4}{c}{\gsmk}           \\
\cmidrule{3-6} \cmidrule{8-11}
Model        &&   $\text{Pre}_{\dingo{}}$ &  $\text{Pre}_{\constrplus{}}$ & \dingo{} & \constrplus{} && $\text{Pre}_{\dingo{}}$ &  $\text{Pre}_{\constrplus{}}$ & \dingo{} & \constrplus{} \\
\midrule
 \llada    &&                  13.2 &                       0 &                   0.1 &                    0.3 &    &        32.1 &             0 &         0.1 &          0.3 \\
 \dream    &&                  11.9 &                       0 &                   0.1 &                    0.1 &    &        37.0 &             0 &         0.1 &          0.2 \\
\bottomrule
\end{tabular}
\end{table}

\begin{table}[p]
    \caption{Three examples demonstrating the impact of constrained decoding on \dlm{} and \mri{} completion. Left are unconstrained completions (\vanilla{}) with problematic tokens highlighted in \wrongtok{red}, and right constrained completions (\constrained{}) with corrections highlighted in \righttok{green}, adapted for clarity. In (a), our method forces \dream{} to generate values of the correct type in a summary of a financial review for task \#30 in our \json{} dataset. In (b), generated by \dreamcoder{}, our method prevents closing more parentheses than are opened when generating a \smiles{} molecule for task \#166. In (c), our method forces \deepseekcsevenb{} to add parentheses around a condition in an if-statement when writing a string processing function in task \#150 of our \humaneval{} dataset.}
  \label{fig:case-study}
  \vspace{2mm}
  \lstset{
  basicstyle=\small,
  breaklines=true,
  numbers=none,
  escapechar=\%,
  basicstyle=\linespread{0.7}\ttfamily\footnotesize,
  framexleftmargin=3mm, 
  framexrightmargin=10mm,
  aboveskip=10mm,
  belowskip=10mm,
  }
  \def\arraystretch{1.3}
  \small
  \centering
  \setlength{\tabcolsep}{8pt}
    \begin{tabular}{@{}>{\centering\arraybackslash}m{1.6em}|l|l@{}}
        &\multicolumn{1}{c|}{\small Vanilla }&\multicolumn{1}{c}{\small Constrained}

        \\\hline
        {\rotatebox[origin=c]{90}{\textsc{json}}}&
    \begin{lstlisting}[language=json]
// summarize my financial review

{
  "capitalGains": %\wrongtok{"5210.5000"}%,
  "interestIncome": %\wrongtok{"1340.25"}%,
  "totalReturn": %\wrongtok{"4.5"}%
}
%\hspace{1pt}%
    \end{lstlisting} &
    \begin{lstlisting}[language=json]
// summarize my financial review

{
  "capitalGains": %\righttok{5210.5000}%,
  "interestIncome": %\righttok{1340.25}%,
  "totalReturn": %\righttok{4.5}%
}
%\hspace{1pt}%
    \end{lstlisting}
        \hspace{1pt}
        \\
        \hline
        {\vspace{2mm}\rotatebox[origin=c]{90}{\textsc{smiles}}}&
    \begin{lstlisting}
// generate an allene with axial chirality

C1=CC1=CC(C(OO)O)C(OO)O%\wrongtok{)C(OO)O}%

    \end{lstlisting} &
    \begin{lstlisting}
// generate an allene with axial chirality

C1=CC1=CC(C(OO)O)C(OO)O
    \end{lstlisting}
        \hspace{1pt}
        \\[8mm]\hline
        {\rotatebox[origin=c]{90}{\textsc{C++}}}&
    \begin{lstlisting}[language=cpp]
// separate the groups of nested parentheses

if (chr=='(')
{
    level+=1;
    current_paren+=chr;
}        
else if %*\wrongtok{chr==\textquotesingle{})\textquotesingle{}}*)
{
    %*\dots*)
    \end{lstlisting} &
    \begin{lstlisting}[language=cpp]
// separate the groups of nested parentheses

if (chr=='(')
{
    level+=1;
    current_paren+=chr;
}        
else if %*\righttok{(chr==\textquotesingle{})\textquotesingle{})}*)
{
    %*\dots*)
    \end{lstlisting}
        \hspace{1pt}
        \\[5mm]
      
  \end{tabular}

\end{table}
\subsection{Case Study}
\label{sec:case-study}

\definecolor{mycrefcolor}{HTML}{0279AF} 

\newcommand{\colorcref}[1]{\textcolor{mycrefcolor}{#1}}
\newcommand{\subcref}[2]{\cref{#1}\hyperref[#1]{\colorcref{#2}}}

For a qualitative evaluation, we manually inspect instances where unconstrained decoding fails and our constraining approach successfully corrects errors.
We showcase three such examples in \cref{fig:case-study}.

\paragraph{Preventing use of invalid types} In \subcref{fig:case-study}{a}, \dream{} generates a summary of a financial review for task \#30 in our \json{} dataset. The schema requires three values of type float. However, the model attempts to generate these values as strings. By applying our constraining method, it can be determined that the strings are misplaced, not constituting one of the required values, and can not match the intended value type. All attempts at placing such strings are thus rejected during generation. Instead, our method forces the model to generate the values without inserting quotes, resulting in a valid and correct result.

\paragraph{Preventing incorrect nesting}
In \subcref{fig:case-study}{b} \dreamcoder{} generates an invalid \smiles{} molecule for task \#166 by closing more parentheses than it opened. Since the CFG for SMILES correctly handles counting of nesting levels, attempts to generate the closing parentheses are rejected by our method. Instead, the model decides to end the generation.

\paragraph{Preventing inadequate syntax}
In \subcref{fig:case-study}{c}, \deepseekcsevenb{} uses conditions without parentheses in an if-statement when writing a string processing function in task \#150 of our \humaneval{} dataset. This confusion may stem from the dominance of Python code in training data, which does not require parentheses in if-statements. However, this is invalid according to C++ syntax. Our method can correct this mistake successfully, resulting in a correct infilling.

\begin{figure}
    \begin{subfigure}{\linewidth}
    \centering
    \vspace{0cm}
    \centering
    \begin{lstlisting}[language=cpp,escapechar=\%]
vector<string> numerical_letter_grade(vector<float> grades){
    vector<string> out={};
    for (int i=0;i<grades.size();i++)
    {
        if (grades[i]>=3.9999) out.push_back("A+");
        if (grades[i]>3.7001 and grades[i]<3.9999) out.push_back("A");
        if (grades[i]>3.3001 and grades[i]<=3.7001) out.push_back("A-");
        if (grades[i]>3.0001 and grades[i]<=3.3001) out.push_back("B+");
        if %(grades[i]>2.7001 and grades[i]<=3.0001) out.push_back("B");%
        %\colorbox{mygreen}{if (grades[i]>2.3001 and grades[i]<=2.7001) out.push_back("B-");}%
        %\colorbox{mygreen}{if (grades[i]>2.0001 and grades[i]<=2.3001) out.push_back("C+");}%
        %\colorbox{mygreen}{if (grades[i]>1.7001 and grades[i]<=2.0001) out.push_back("C");}%
        %\colorbox{mygreen}{if (grades[i]>1.3001 and grades[i]<=1.7001) out.push_back("C-");}%
        %\colorbox{mygreen}{if (grades[i]>1.0001 and grades[i]<=1.3001) out.push_back("D+");}%
        %\colorbox{mygreen}{if (grades[i]>0.7001 and grades[i]<=1.0001) out.push_back("D");}%
        %\colorbox{mygreen}{if}% %$\hole$% i]<=3.0001) out.push_back("B");
        if (grades[i]>2.3001 and grades[i]<=2.7001) out.push_back("B-");
        if (grades[i]>2.0001 and grades[i]<=2.3001) out.push_back("C+");
        %\dots%
    \end{lstlisting}

\vspace{-4mm}
    \subcaption{\starcodertwo{} exceeds the token limit in task \#81 in \mrione{}.}
    \label{fig:non-terminating-code}
    \end{subfigure}
    \begin{subfigure}{\linewidth}
        
    \centering
    \vspace{.5cm}
    \centering
    \begin{tabular}{ccc}
        \toprule
        Vanilla & Constrained$^{-}$ & Constrained \\
        \midrule
        \ttt{C6CCCC1)} & \ttt{C6CCCC$\mask$))} & \ttt{C6CCCC(c(c)(c))} \\
        \bottomrule
    \end{tabular}
    \subcaption{\llada{} leaves a single $\mask$ for completion in task \#153 in \smiles{}.}
    \label{fig:non-sampled-code}
    
    \end{subfigure}
    \caption{Syntax errors may remain when the model has fewer tokens left to complete than would be required to fulfil the syntactic constraints. This can happen both in \mri{} (a), when the model exceeds the maximum number of generated tokens and in \dlm{} (b), when the model has few mask tokens $\mask$ remaining.}
\end{figure}

\section{Discussion}
\label{sec:discussion}


\paragraph{Remaining syntax errors}
While our method achieves substantial improvements in syntactic correctness, using only \constrained{} still leaves a considerable gap until guaranteeing correctness. We attribute most of this gap to the overapproximation of allowing an arbitrary number of tokens to fill regions in the partial output, as done in prior work \citep{dominosDblp2403,ugare2024syncode}. In practice, the LLM is typically limited, i.e., in \fim{} and \mri{} it can only generate up to the user-defined maximum number of tokens, and in \dlm{} it can only generate one token per mask~$\mask$. Examples of this issue occurring are presented in \cref{fig:non-terminating-code} for \mri{}, where the model exceeds the token limit of 256 tokens as it generates large amounts of unnecessary code, and in \cref{fig:non-sampled-code}, where the \dlm{} model needs to open several molecule branches in a single remaining token.

One approach to resolve this issue would be to accurately model the remaining number of tokens in our regular language construction. However, we observe in experiments that this significantly increases the size of the regular language, as it consequently needs to keep track of the number of inserted tokens. This drastically increases the size of the intersection language, rendering our current implementation impractical. 

Another approach would be to train the model to signal requiring additional tokens. In \mri{}, this naturally occurs when the model does not generate an end-of-string token. For \dlm{}, a special token could be added that is replaced with two mask tokens after sampling, increasing the size of the affected infilling region. Concurrent work by \citet{Dreamon2025} reports that such capabilities appear to generally improve model performance for code infilling.

Our chosen approach to mitigate the issue is to automatically fill in the output based on the formal language constraints (\constrplus). However, this solution cannot rely on the model's probability distribution to steer generation. Determining the most effective way to handle this limitation is an important topic for future work.

\paragraph{Leveraging incremental parsing} 
While we take several steps to improve the efficiency of our method, it can still require a significant amount of time to determine the emptiness of the intersection language after each generated token. Future work may leverage the fact that the CFG for the intersection is fixed and the DFA is only updated using small modifications. This may lead to an approach for incrementally computing emptiness checks by reusing the results of the previous intersection computation. Other approaches to leverage the incremental nature of the parsing, similar to the approaches of \citet{melcer2024constraineddecodingfillinthemiddlecode,ugare2024syncode}, and \citet{mundler2025typeaware} would likely also be able to decrease the worst case and practical overhead of the constraining method.

\paragraph{Context-sensitive language features}
While our method is designed for context-free languages, an interesting future direction would be extensions to handle more powerful language classes, such as context-sensitive languages.
Similar to \citet{melcer2024constraineddecodingfillinthemiddlecode} and \citet{ugare2024syncode}, simple context-sensitive syntactic features can likely be handled by preprocessing through adequate lexers. Beyond syntactic features, prior work suggested leveraging more semantic insights, such as type systems \citep{mundler2025typeaware}, for constructing more powerful constraint systems. Type checkers with typed holes \citep{typed-holes-DBLP:journals/pacmpl/OmarVCH19} could be leveraged to achieve such systems.

\section{Prompts}
\label{app:prompt}

In this section, we detail all prompts used for the respective models and tasks.

\paragraph{\mri{}}
Since models used for \fim{} and \mri{} tasks are not instruction-fine-tuned, we provide the model with the raw code, including only a comment above the function to guide the model for completions. We use the standard templating suggested for each model to format the prompt for \fim{} and \mri{} completion.
If \mri{} is not supported explicitly, we emulate it by inserting \texttt{<TODO>} into the remaining infilling regions and repeatedly prompting the model for FIM completion on the first infilling region. In order to prevent the models from generating main methods and tests in the \mri{} setting, we add a main method at the end of the context that is marked with a TODO comment. An example prompt for the \mritwo{} setting is provided in \cref{prompt:cpp-mri}.

\paragraph{\dlm{}}
The models used for \dlm{} tasks are instruction-fine-tuned, allowing us to specify the completion intent in natural language. We provide a general description of the task in the system prompt and the specific task content as the first user prompt. The assistant response is prefilled with the start of a code fence and, in the case of \humaneval{}, with the necessary header declarations and function signature to ensure results can be extracted and tests can be executed correctly.
The prompts for \humaneval{}, \json{} and \smiles{} tasks are presented in \cref{prompt:cpp-dllm}, \labelcref{prompt:json-dllm} and \labelcref{prompt:smiles-dllm} respectively.

\paragraph{Benchmark generation prompts}
As outlined in \cref{app:experiments}, we generate the \json{} and \smiles{} dataset synthetically by prompting \gemini{}. We provide the used prompts for generation and validation of the generated samples in \cref{prompt:json-schema-generation,prompt:json-schema-verification,prompt:smiles-generation,prompt:smiles-verification}.

\paragraph{Miscellaneous Prompts}
The prompt for the \gsmk{} experiment for \dingo{} is shown in \cref{prompt:gsm8k}. An example prompt for the grammar prompting (\grammarprompting) is shown in \cref{prompt:grammar-prompting}.

\clearpage

\begin{figure}
\begin{lstlisting}[keywordstyle=\ttfamily,commentstyle=\ttfamily]
region 0
    %/*
    From a given vector of integers, generate a vector of rolling maximum element found
    until given moment in the sequence.
    >>> rolling_max({1, 2, 3, 2, 3, 4, 2})
    {1, 2, 3, 3, 3, 4, 4}
    */
    #include<stdio.h>
    #include<vector>
    using namespace std;
    vector<int> rolling_max(vector<int> numbers){%
        ??vector<int> out;??

region 1
        ??for (int i=0;i<numbers.size??

region 2
        ??return out;
}??

int main(){
    // TODO
}
\end{lstlisting}
\caption{Example prompt for the \mritwo{} task \#1. The intial comment and function signature in \textcolor{blue}{blue} are derived from the dataset prompt, and the remaining code snippets in \textcolor{pastelgreen}{green} are the remainders of the canonical solution with two randomly removed spans. We append a stub main function to prevent the model from attempting to generate a main function of its own.}
\label{prompt:cpp-mri}
\end{figure}

\begin{figure}
\begin{lstlisting}[keywordstyle=\ttfamily,commentstyle=\ttfamily]
system
    You are an expert in C++ programming. Solve the given problem by writing solution
     code in C++.
    When answering, insert the solution code in a ```cpp...``` block. Do neither include
     test cases not a main function.

user
    %Check if in given vector of numbers, are any two numbers closer to each other than
    given threshold.
    >>> has_close_elements({1.0, 2.0, 3.0}, 0.5)
    false
    >>> has_close_elements({1.0, 2.8, 3.0, 4.0, 5.0, 2.0}, 0.3)
    true%

assistant
    ```cpp
    ??#include<stdio.h>
    #include<vector>
    #include<math.h>
    using namespace std;
    #include<algorithm>
    #include<stdlib.h>
    bool has_close_elements(vector<float> numbers, float threshold){??
\end{lstlisting}
\caption{Example prompt for the \humaneval{} task \#1. The system prompt in black is fixed, whereas the user prompt in \textcolor{blue}{blue} is extracted from the comment preceding the function and the assistant response is prefilled with a codefence, and in \textcolor{pastelgreen}{green}, headers, and the function signature of each task.}
\label{prompt:cpp-dllm}
\end{figure}

\begin{figure}
\begin{lstlisting}[keywordstyle=\ttfamily,commentstyle=\ttfamily]

system
    You are a helpful assistant that answers in JSON. Here is the JSON schema you must
    adhere to:
    <schema>
    ??{
        "type": "object",
        "properties": {
            "name": {
                "type": "string"
            },
            "email": {
                "type": "string"
            },
            "shippingAddress": {
                "type": "string"
            }
        },
        "required": [
            "name",
            "email",
            "shippingAddress"
        ],
        "additionalProperties": false
    }??
    </schema>

user
    %We are registering 'Global Exports Ltd.' for your services. The main contact person
    is Samantha Davis, and her corporate email is s.davis@globalexports.co.uk. All ship-
    ments and correspondence should be directed to our headquarters: Global Exports Ltd.,
    12 Business Park Road, Manchester, M1 1AB, United Kingdom. We are looking forward to
    a fruitful partner ship and are particularly interested in your international ship-
    ping rates.%

assistant
    ```json
\end{lstlisting}
\caption{Example prompt for the \json{} task. The JSON schema in \textcolor{pastelgreen}{green} is task-specific as well as the the user prompt in \textcolor{blue}{blue} from which information should be extracted into the given schema. The system prompt and prefilled assistant response are fixed.}
\label{prompt:json-dllm}
\end{figure}

\begin{figure}
\begin{lstlisting}[keywordstyle=\ttfamily,commentstyle=\ttfamily]

system
    You are a specialized AI assistant that generates SMILES (Simplified Molecular Input
    Line Entry System) strings from chemical descriptions. You will be given a textual
    description of a chemical compound or a related task. Your goal is to produce the
    most accurate and valid SMILES string representing that description.

    Your Task:

    Based on the provided "input" description, generate the corresponding SMILES string.

    Output Requirements:

    - Provide only the SMILES string as your output.
    - Ensure the SMILES string is syntactically valid.
    - Represent all specified chemical features accurately (atoms, bonds, rings,
        aromaticity, charge, isotopes, stereochemistry).

    Output:

    - Provide only the smiles molecule as a raw string between triple backticks (```).
    For instance:
    ```smiles
    C1=CC=CC=C1
    ```

user
    %Propan-1-amine, a primary amine with a three-carbon straight chain and the amino
    group on the first carbon.%

assistant
    ```smiles
\end{lstlisting}
\caption{Example prompt for the \smiles{} task. The user prompt in \textcolor{blue}{blue} varies per task.}
\label{prompt:smiles-dllm}
\end{figure}

\begin{figure}
    \begin{lstlisting}[keywordstyle=\ttfamily,commentstyle=\ttfamily]
    user
        Your goal is to create challenging and diverse `JSON Schema` problems. You are 
        given a JSON schema that describes a specific schema for a JSON problem.

        You should generate **{num_samples}** JSON benchmark samples based on the 
        provided schema. A benchmark sample consists of a natural language description 
        describing how the JSON schema should be filled out, along with a JSON object
        that adheres to the schema.
    
        For each sample, provide a JSON object with the following structure:
    
        ```json
        {{
            "input": "A natural language description of how the JSON schema should be 
                filled out. The input should be a natural query that a user might ask an
                LLM. The input will be given to the LLM as a prompt, along with the JSON
                schema. Based on this input, the LLM should generate a JSON object that
                adheres to the schema.",
            "output": "A JSON object that adheres to the provided schema. The output 
                should be a valid JSON object that matches the schema and reflects the 
                input description."
        }}
        ```
    
        **Guidelines for generating samples:**
    
        - **Variety**: Describe a wide range of scenarios that can be expressed using 
            the JSON schema. Ensure that the samples cover a wide range of possible 
            scenarios, and make them sound natural and plausible.
        - **Difficulty**: User queries can and should contain distracting information
            and longer backgrounds. 
        - **Realism**: Test cases should reflect plausible scenarios where the JSON 
            schema would be used.
        - **Reference**: Do not reference the JSON schema in the input description. The
            input should be a natural query that a user might ask an LLM. It should not 
            reference JSON at all.
    
        JSON Schema:
        {schema}
    
        Example Input (Do not use this in your samples):
        {input_query}
    
        Example Output (Do not use this in your samples):
        {output_query}
    \end{lstlisting}
    \caption{Prompt used to generate additional JSON Schema samples for the \json{} task using \gemini{}. Several samples were generated at the same time to increase diversity.}
    \label{prompt:json-schema-generation}
\end{figure}

\begin{figure}
    \begin{lstlisting}[keywordstyle=\ttfamily,commentstyle=\ttfamily]
    user
        You are a JSON Schema assistant. You will be given a textual description of how
        a JSON schema should be filled out. Your task is to generate a JSON object that
        adheres to the provided schema.

        Your Task:
        - Analyze the textual task.
        - Construct a JSON object that correctly implements the task based on the 
            provided schema.
        
        The JSON object should be a valid JSON object that matches the schema and 
        reflects the input description.
    
        Output:
        - Provide only the JSON object as a raw string between triple backticks 
            (```json). Ensure the JSON object satisfies the JSON schema. For instance:
        ```json
        {{
        "key": "value",
        "number": 42,
        "array": [1, 2, 3]
        }}
        ```
  
        Json Schema:
        {schema}
    
        Description:
        {input_query}
    \end{lstlisting}
    \caption{Prompt used to verify additional JSON Schema samples for the \json{} task  using \gemini{}.}
    \label{prompt:json-schema-verification}
\end{figure}

\begin{figure}
    \begin{lstlisting}[keywordstyle=\ttfamily,commentstyle=\ttfamily]
    user
        You are a specialized AI assistant tasked with generating benchmark samples for
        SMILES (Simplified Molecular Input Line Entry System) string generation. Your 
        goal is to create diverse and accurate chemical structure descriptions and their
        corresponding SMILES strings.

        Please generate **{num_samples}** benchmark samples.
    
        The difficulty of these samples should be: **{difficulty_description}**.
        Examples of difficulty levels:
        * **Beginner**: Simple acyclic molecules, common functional groups (e.g., 
            ethanol, acetic acid, propanamine), small alkanes/alkenes/alkynes.
        * **Intermediate**: Molecules with single or multiple rings (e.g., cyclohexane, 
            pyridine, naphthalene), basic stereochemistry (R/S, E/Z using `@@`, `/`, `\`),
            common drugs or biomolecules (e.g., aspirin, glucose in its open-chain form).
        * **Advanced**: Complex polycyclic systems (e.g., steroids, bridged compounds),
            detailed stereochemistry, isotopic labeling, salts, mixtures, or reaction 
            SMILES (if the task is to represent a reaction).
    
        For each sample, provide a JSON object with the following structure:
    
        ```json
        {{
        "input": "A natural language description of a chemical compound or a task that
            uniquely defines a chemical structure representable by a SMILES string. 
            This could be an IUPAC name, a common name, a structural description, or 
            a request to modify a base structure.",
        "output": "The correct and valid SMILES string for the chemical structure 
            described in the 'input'. Correctness and validity are paramount."
        }}
        ```
    
        **Guidelines for generating samples**:
    
        - **Accuracy**: The generated SMILES string in the "output" field MUST 
            accurately represent the chemical structure described in the "input". Ensure 
            correct atom types, bond orders, connectivity, aromaticity, charges, 
            isotopes, and stereochemistry as implied by the input.
        - **Validity**: All generated SMILES strings must be syntactically valid.
        - **Clarity of Input**: The "input" description should be unambiguous and 
            provide enough information to define a specific chemical structure. Avoid 
            overly vague descriptions.
        - **Variety**: Generate a diverse set of samples covering different chemical 
            families, structural features (rings, unsaturation, heteroatoms, functional 
            groups), and complexities according to the specified difficulty.
    
        Output Format:
    
        Return a JSON list containing the {num_samples} generated JSON objects.
    \end{lstlisting}
    \caption{Prompt used to generate additional samples for the \smiles{} task using \gemini{}. Several samples were generated at the same time to increase diversity.}
    \label{prompt:smiles-generation}
\end{figure}

\begin{figure}
    \begin{lstlisting}[keywordstyle=\ttfamily,commentstyle=\ttfamily]
    user
        You are a specialized AI assistant that generates SMILES (Simplified Molecular 
        Input Line Entry System) strings from chemical descriptions. You will be given
        a textual description of a chemical compound or a related task. Your goal is 
        to produce the most accurate and valid SMILES string representing that 
        description.

        Your Task:

        Based on the provided "input" description, generate the corresponding SMILES 
        string.

        Output Requirements:

        - Provide only the SMILES string as your output.
        - Ensure the SMILES string is syntactically valid.
        - Represent all specified chemical features accurately (atoms, bonds, rings,
            aromaticity, charge, isotopes, stereochemistry).

        Output:

        - Provide only the smiles molecule as a raw string between triple backticks (```).
        For instance:
        ```smiles
        C1=CC=CC=C1
        ```

        {sample}
    \end{lstlisting}
    \caption{Prompt used to verify samples for the \smiles{} task  using \gemini{}.}
    \label{prompt:smiles-verification}
\end{figure}

\begin{figure}
    \begin{lstlisting}[keywordstyle=\ttfamily,commentstyle=\ttfamily,breaklines=true]
system
    You are an expert in solving grade school math tasks. You will be presented with a grade-school math word problem with symbolic variables and be asked to solve it.\n\nBefore answering you should reason about the problem (using the <reasoning> field in the response described below). Intermediate symbolic expressions generated during reasoning should be wrapped in << >>.\n\nOnly output the symbolic expression wrapped in << >> that answers the question. The expression must use numbers as well as the variables defined in the question. You are only allowed to use the following operations: +, -, /, //, %, *, and **.
    
    You will always respond in the format described below: 
    Let's think step by step. <reasoning> The final answer is <<symbolic expression>>
    
    There are {t} trees in the {g}. {g} workers will plant trees in the {g} today. After they are done, there will be {tf} trees. How many trees did the {g} workers plant today?
    
    Let's think step by step. Initially, there are {t} trees. After planting, there are {tf} trees. The number of trees planted is <<tf - t>>. The final answer is <<tf - t>>.
    
    If there are {c} cars in the parking lot and {nc} more cars arrive, how many cars are in the parking lot?
    
    Let`s think step by step. Initially, there are {c} cars. {nc} more cars arrive, so the total becomes <<c + nc>>. The final answer is <<c + nc>>.
    
    {p1} had {ch1} {o1} and {p2} had {ch2} {o1}. If they ate {a} {o1}, how many pieces do they have left in total?
    
    Let's think step by step. Initially, {p1} had {ch1} {o1}, and {p2} had {ch2} {o1}, making a total of <<ch1 + ch2>>. After eating {a} {o1}, the remaining total is <<ch1 + ch2 - a>>. The final answer is <<ch1 + ch2 - a>>.
    
    {p1} had {l1} {o1}. {p1} gave {g} {o1} to {p2}. How many {o1} does {p1} have left?
    
    Let's think step by step. {p1} started with {l1} {o1}. After giving {g} {o1} to {p2}, {p1} has <<l1 - g>> {o1} left. The final answer is <<l1 - g>>.
    
user
    %{name} picks {n1} {fruit}s on {d1}. Then he picks {n2} {fruit}s on {d2}. On {d3}, he picks {mult} the number of {fruit}s he did on {d1}. How many {fruit}s does {name} have?%

assistant
            
    \end{lstlisting}
    \caption{\rev{Prompt for the \gsmk{} task with example user prompt in \textcolor{blue}{blue} at the end.}}
    \label{prompt:gsm8k}
\end{figure}

\begin{figure}
    \begin{lstlisting}[keywordstyle=\ttfamily,commentstyle=\ttfamily,breaklines=true]
user:
    ...
    The answer must adhere to the following grammar: 
        
    %S -> Line
    Line -> Atom OptComboChainBranchList
    OptComboChainBranchList -> ComboChainBranchList | $\epsilon$
    ComboChainBranchList -> ComboChainBranchElement | ComboChainBranchElement ComboChainBranchList
    ComboChainBranchElement -> Chain | Branch
    Chain -> . Atom | OptBond ComboAtomRnumList
    OptBond -> Bond | $\epsilon$
    ComboAtomRnumList -> ComboAtomRnumElement | ComboAtomRnumElement ComboAtomRnumList
    ComboAtomRnumElement -> Atom | Rnum
    Bond -> - | bond
    Branch -> ( OptBondOrDotLineList ) 
    OptBondOrDotLineList -> OptBondOrDotLineElement | OptBondOrDotLineElement OptBondOrDotLineList
    OptBondOrDotLineElement -> OptBondOrDot Line
    OptBondOrDot -> Bond | . | $\epsilon$
    Atom -> organicSymbol | BracketAtom
    BracketAtom -> [ OptionalIsotope Symbol OptionalChiral OptionalHCount OptionalCharge OptionalMap ]
    OptionalIsotope -> Isotope | $\epsilon$
    OptionalChiral -> chiral | $\epsilon$
    OptionalHCount -> HCount | $\epsilon$
    OptionalCharge -> Charge | $\epsilon$
    OptionalMap -> Map | $\epsilon$
    Rnum -> digit | perc digit digit
    Isotope -> digit | digit digit | digit digit digit
    HCount -> h digit | h
    Charge -> + | + + | + fifteen | + digit | - | - - | - fifteen | - digit
    Map -> : Isotope
    Symbol -> organicSymbol | anorganicSymbol | h%
    \end{lstlisting}
    \caption{\rev{For grammar prompting (\grammarprompting), the grammar in EBNF is appended to the user prompt. The example grammar shown in \textcolor{blue}{blue} is the \smiles{} grammar.}}
    \label{prompt:grammar-prompting}
\end{figure}

\end{document}